\documentclass[10pt,twocolumn,letterpaper]{article}
\pdfoutput=1
\usepackage{cvpr}
\usepackage{times}
\usepackage{epsfig}
\usepackage{graphicx}
\usepackage{amsmath}
\usepackage{amssymb}
\usepackage{color}
\usepackage{times}
\usepackage{epsfig}
\usepackage{algorithm}
\usepackage{algorithmicx}
\usepackage{algpseudocode}
\usepackage{graphics}
\usepackage{threeparttable}
\usepackage{color}
\usepackage[normalem]{ulem}
\usepackage{multirow}
\usepackage{float}
\usepackage{amsfonts}
\usepackage{bm}
\usepackage{subfig}
\usepackage{array}
\usepackage[table]{xcolor}
\usepackage{colortbl}
\usepackage{pifont}
\usepackage{cite}
\usepackage{diagbox}
\usepackage{rotating}
\usepackage{booktabs}
\usepackage{overpic}
\usepackage{textcomp}
\usepackage{contour}
\usepackage{enumitem}
\usepackage{ulem}

\usepackage[pagebackref=true,breaklinks=true,letterpaper=true,colorlinks,bookmarks=false]{hyperref}
\cvprfinalcopy 

\def\cvprPaperID{****} 

\ifcvprfinal\fi
\begin{document}

\title{A Switched View of Retinex: Deep Self-Regularized Low-Light Image Enhancement}
\author{Zhuqing Jiang$^1$\quad  Haotian Li$^1$\quad Liangjie Liu$^1$\quad Aidong Men$^1$\quad Haiying Wang$^1$\\
	\small{$^1$} \small Beijing University of Posts and Telecommunications \\
	{\tt\small $\{$jiangzhuqing,lht1996,2018213148,menad,why$\}$@bupt.edu.cn
}
}

\maketitle
\thispagestyle{empty}

\vspace{-8pt}
\begin{abstract}
Self-regularized low-light image enhancement does not require any normal-light image in training, thereby freeing from the chains on paired or unpaired low-/normal-images. However, existing methods suffer color deviation and fail to generalize to various lighting conditions.
This paper presents a novel self-regularized method based on Retinex, which, inspired by HSV, preserves all colors (Hue, Saturation) and only integrates Retinex theory into brightness (Value). We build a reflectance estimation network by restricting the consistency of reflectances embedded in both the original and a novel random disturbed form of the brightness of the same scene. The generated reflectance, which is assumed to be irrelevant of  illumination by Retinex, is treated as enhanced brightness.
Our method is efficient as a low-light image is decoupled into two subspaces, color and brightness, for better preservation and enhancement. 
Extensive experiments demonstrate that our method outperforms multiple state-of-the-art algorithms qualitatively and quantitatively and adapts to more lighting conditions.

\end{abstract}

\section{Introduction}

Images captured under insufficient illumination often result in poor visual effect, challenging aesthetic quality as well as performance of many downstream computer vision tasks, such as object recognition and semantic segmentation. Hence, low-light image enhancement has attracted huge research interest.

\begin{figure}[t]
    \centering
	  \subfloat[Input]{ \label{show:input}
       \includegraphics[width=0.4\linewidth]{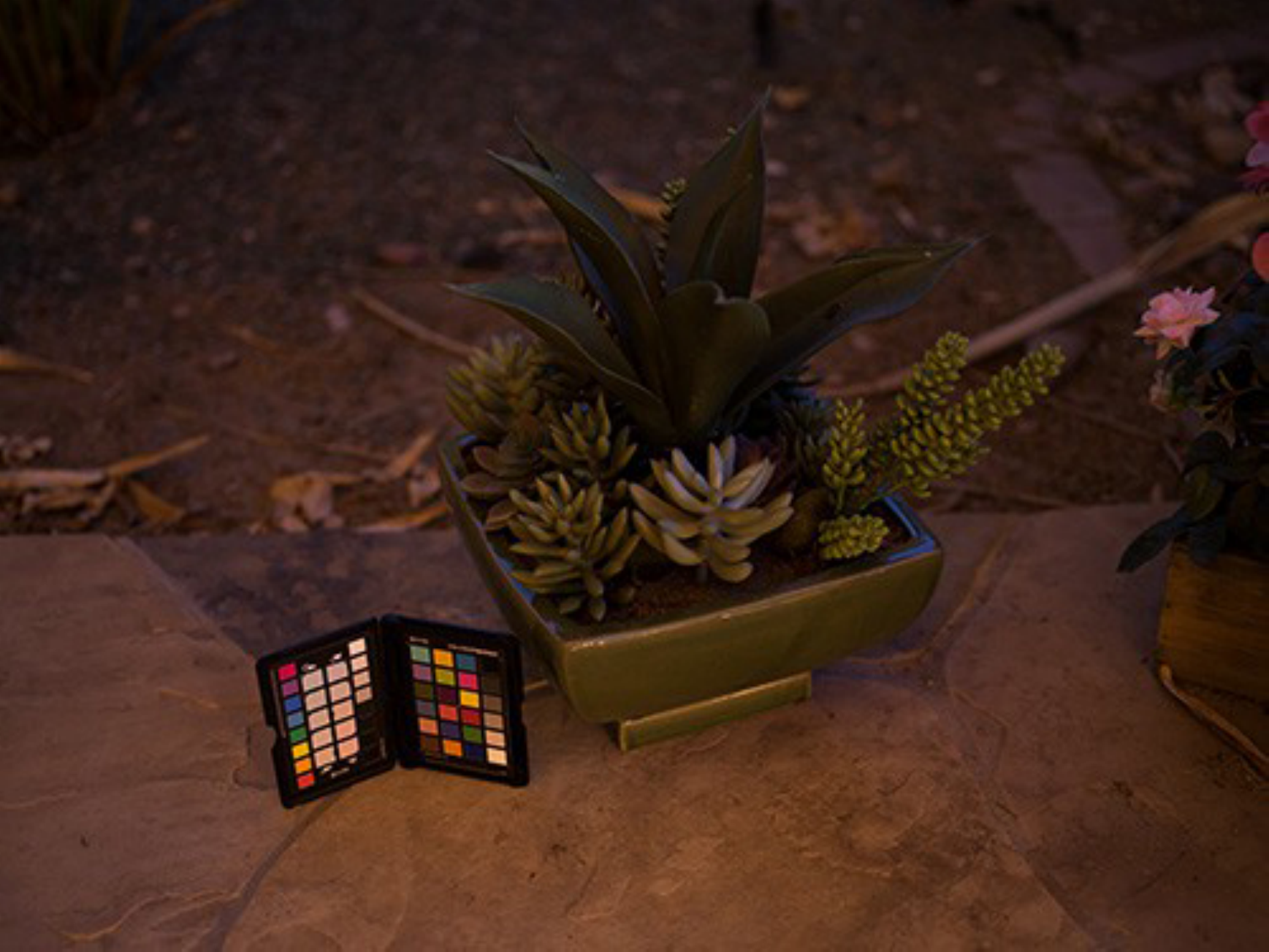}}
	  \quad
	  \subfloat[EnlightenGAN \cite{jiang2019enlightengan}]{\label{show:enlighten}
        \includegraphics[width=0.4\linewidth]{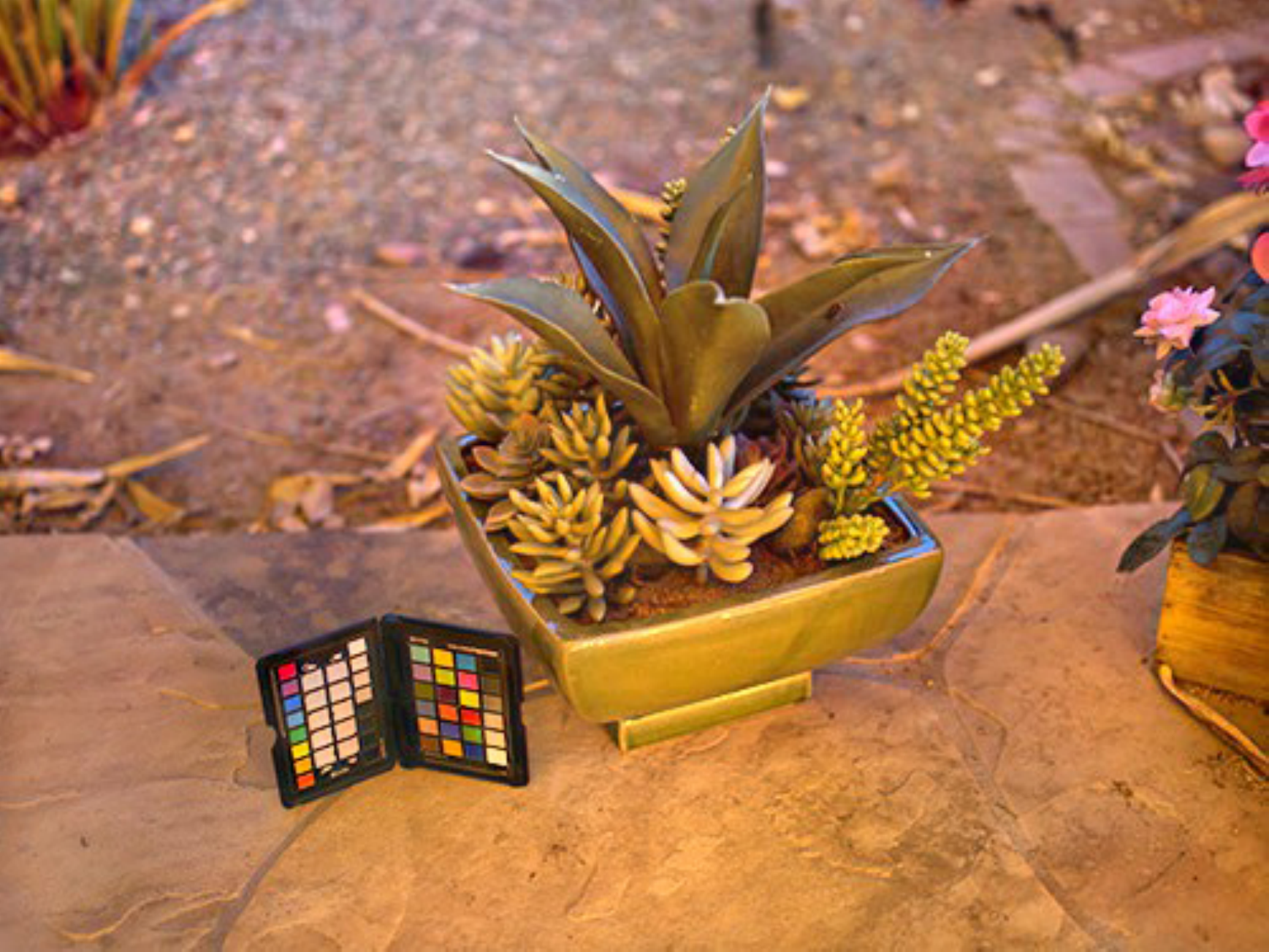}}\\
	  \subfloat[Zero-DCE \cite{guo2020zero}]{\label{show:zerodce}
        \includegraphics[width=0.4\linewidth]{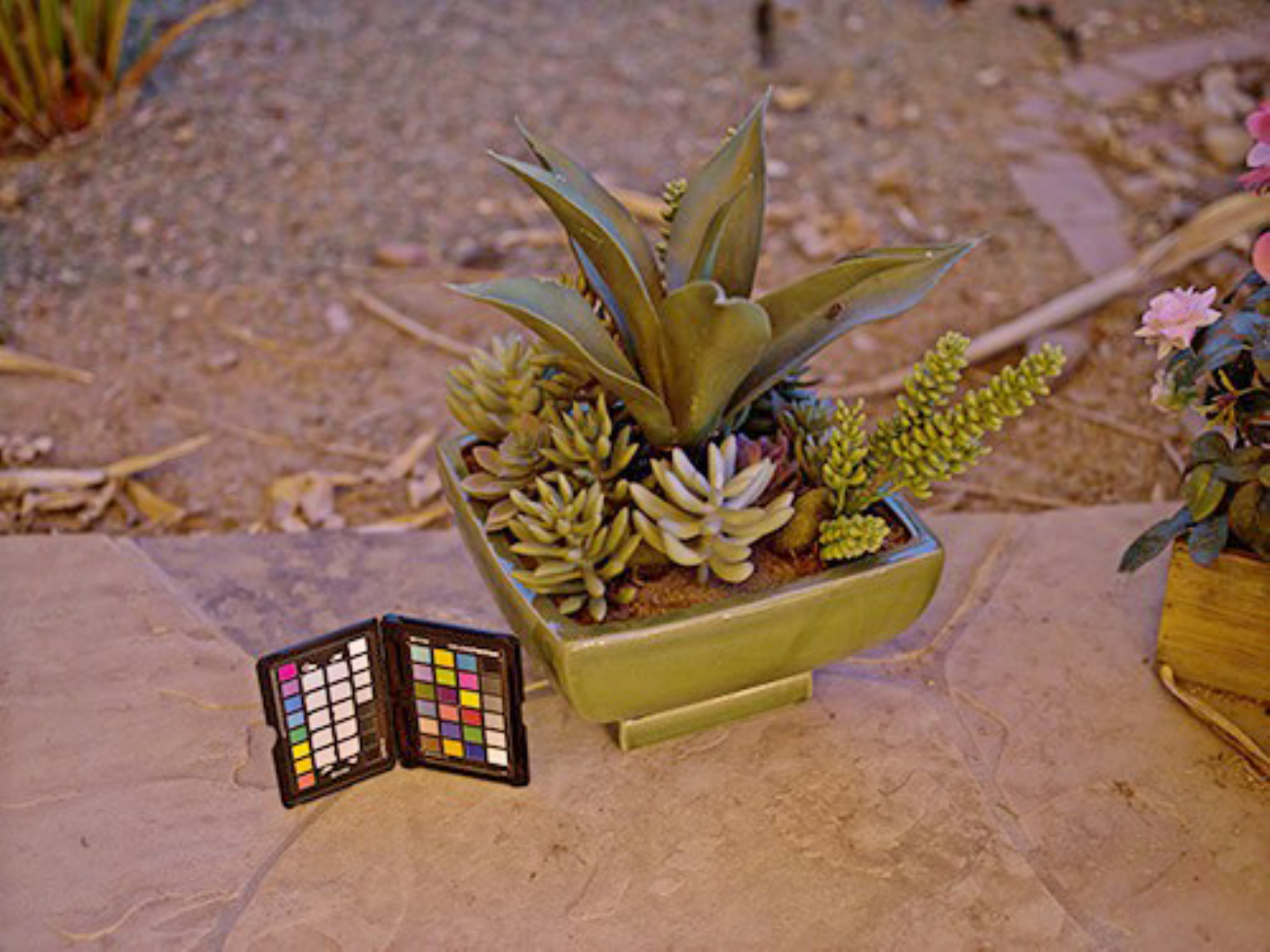}}
    \quad
	  \subfloat[Ours]{\label{show:ours}
        \includegraphics[width=0.4\linewidth]{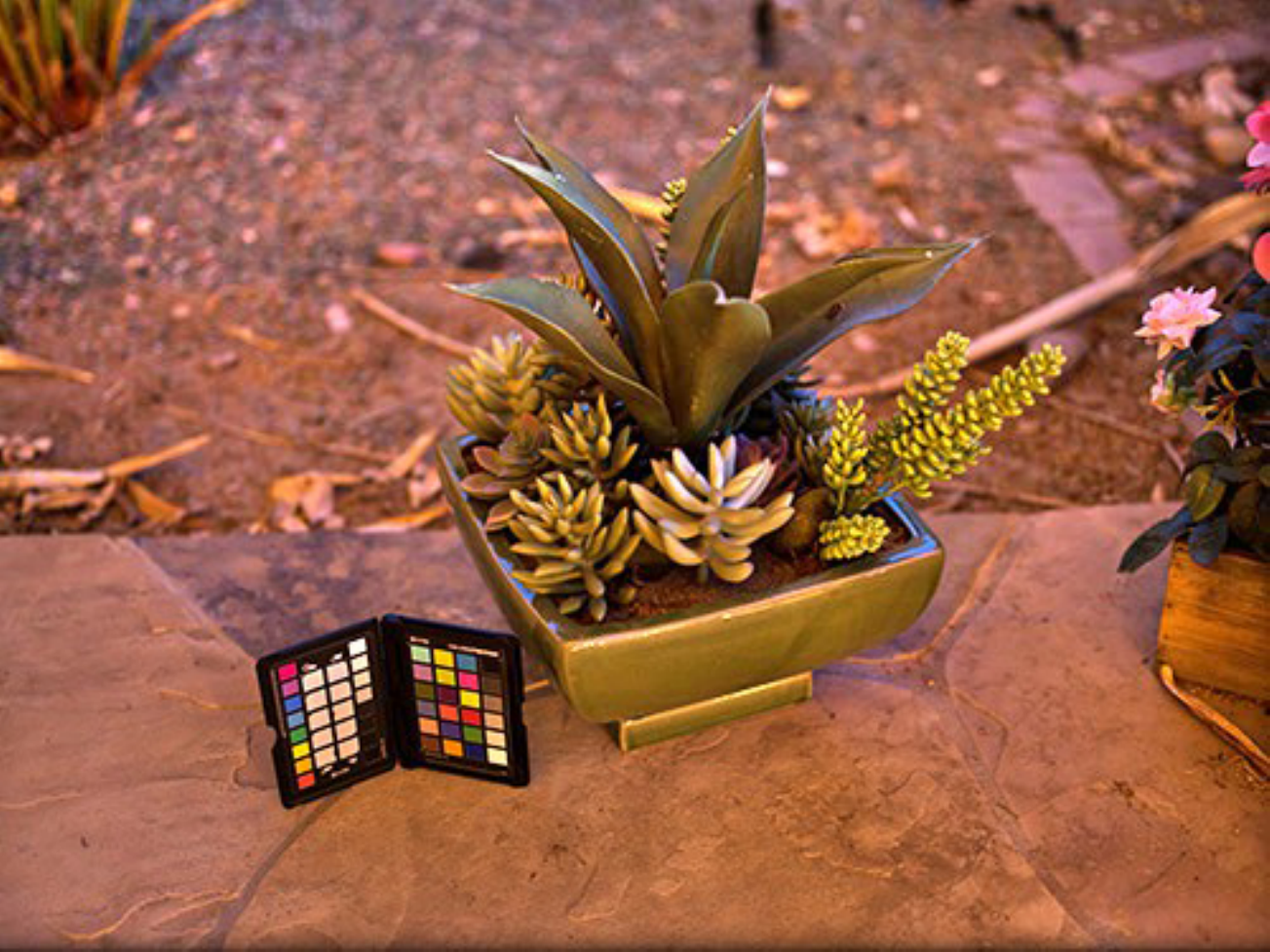}}
	  \caption{Visual performance on a typical low-light image by three methods. Our result is of favorable brightness and color. Note that in our method, all colors are derived from the input and only brightness is enhanced.}
	  \label{fig:show} 
\end{figure}

Over the past few decades, a series of algorithms have been proposed to enhance low-light images, including conventional algorithms \cite{pizer1987adaptive,abdullah2007dynamic,land1977retinex,jobson1997multiscale,jobson1997properties} and deep learning-based methods \cite{lore2017llnet,wei2018deep,jiang2019enlightengan,wang2019underexposed,zhang2019zero,zhu2020zero,guo2020zero}. In particular, self-regularized learning only requires low-light images in training, which frees from the constraints on paired or unpaired low-/normal-light datasets, and has become a trending research topic \cite{zhang2019zero,zhu2020zero,zhang2020self,guo2020zero}. Guo \etal \cite{guo2020zero} designed a deep curve estimation method, regularized by non-reference loss functions. The enhanced images are comparable to those generated based on supervised or generative adversarial network (GAN) algorithms. However, this method suffers color deviation and fails to adapt to various lighting conditions.

This paper presents a novel self-regularized low-light image enhancement method based on Retinex \cite{land1977retinex}, dedicating to reducing color deviation and adapting to more lighting conditions. This method, inspired by HSV color space, preserves all colors (Hue, Saturation) and only integrates Retinex theory into brightness (Value). Besides, since the absence of paired or unpaired low-/normal-light images, this method proposes a novel random brightness disturbance approach to obtain another abnormal brightness of the same scene. According to Retinex, reflectances embedded in both the original and the disturbed form of the brightness should be consistent. Based on this, this method designs a deep reflectance estimation network. The generated reflectance is treated as enhanced brightness, which has been presented in some existing methods \cite{guo2016lime,fu2016weighted,wang2019underexposed,zhang2020self}. 
Finally, an enhanced image is generated by regrouping the original colors and the enhanced brightness. Our method is efficient as a low-light image is decoupled into two subspaces, color and brightness, for better preservation and enhancement. A typical result is shown in Fig.\ref{fig:show}. Our result has a satisfactory performance in terms of colors and brightness.

Our contributions are summarized as follows:

(1) A novel self-regularized method is proposed for low-light image enhancement, which, inspired by HSV color space, preserves all colors (Hue, Saturation) and only integrates Retinex into brightness (Value).

(2) We design a deep reflectance estimation network, regularized by the consistency of reflectances embedded in both the original and a novel random disturbed form of the brightness of the same scene.

(3) Extensive experiments demonstrate the effectiveness of our method and its superiority over state-of-the-art methods.

The rest of this paper is organized as follows. 
In section \ref{Methodology}, the proposed method is described in detail.
In section \ref{Experiments}, we present extensive experiments to evaluate the performance of our method.
Section \ref{Conclusion} is the conclusion.

\begin{figure*}[t]
	\centering
	\centerline{\includegraphics[width=\linewidth]{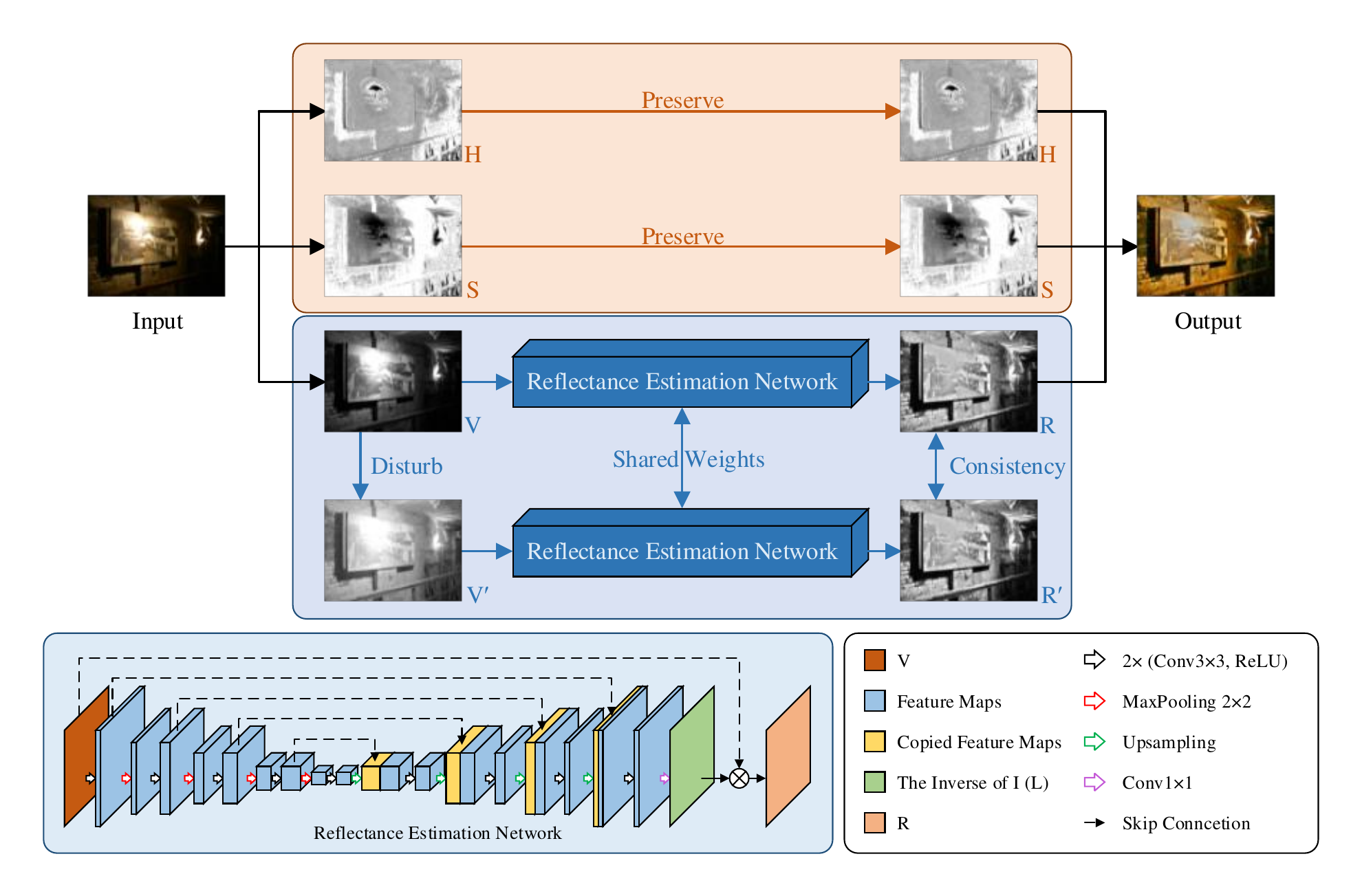}}
	\caption{The framework of our method. A low-light image is described by colors (Hue, Saturation) and brightness (Value) in HSV color space. The colors are preserved, and only the brightness is enhanced by the reflectance estimation network. An enhanced image is generated by regrouping the original colors and the enhanced brightness. The details of the reflectance estimation network are shown at the bottom.}
	\label{fig:pipeline}
\end{figure*}

\section{Methodology} 
\label{Methodology}

We present our method in Fig.\ref{fig:pipeline}. A low-light image is normalized to [0,1] and described by colors (Hue, Saturation) and brightness (Value) in HSV color space.
We design a random brightness disturbance approach to obtain another abnormal brightness of the same scene ($\mathbf{V^{'}}$ in Fig.\ref{fig:pipeline}). The original and the disturbed form of brightness share the same reflectance estimation network. We treat the generated reflectance as enhanced brightness. Finally, an enhanced image is generated by regrouping the original colors and the enhanced brightness in the direction of channel, converting to RGB color space, and regulating to [0,255]. We next detail the advantages of HSV color space for low-light image enhancement in Sec.\ref{HSV}, the novel random brightness disturbance in Sec.\ref{Random_disturbance}, and the reflectance estimation network in Sec.\ref{network}. 

\subsection{HSV Color Space for Low-Light Image Enhancement}
\label{HSV}
We propose to describe a low-light image in HSV color space, where it is defined by three channels, \ie, hue (H), saturation (S), and value (V). The first two are color descriptors, and the last one is brightness descriptor. The advantages of HSV color space are summarized as follows: 

(1) The H, S, and V channels are orthogonal to each other, which decouples brightness and color. Based on this, we design a method to keep all the colors and only enhance the brightness, which helps to faithfully restore the colors hidden in the darkness and reduces the complexity of the brightness enhancement model. Specifically in self-regularized learning, since the absence of ground-truth, convolution neural networks (CNN) fail to learn a color mapping from a low-light image to its corresponding normal-light image, highlighting the advantages of the HSV color space.

(2) The V channel corresponds to the maximum channel of the low-light image, \ie, the value of each pixel is the maximum of the three values of the corresponding pixel in R, G, and B channels. As it states in \cite{zhang2020self}, the maximum channel has the maximum entropy, which provides the maximum amount of information to enhance the low-light image.

\begin{figure} [t]
    \centering
	  \subfloat[Low-Light Image]{\label{input}
       \includegraphics[width=0.3\linewidth]{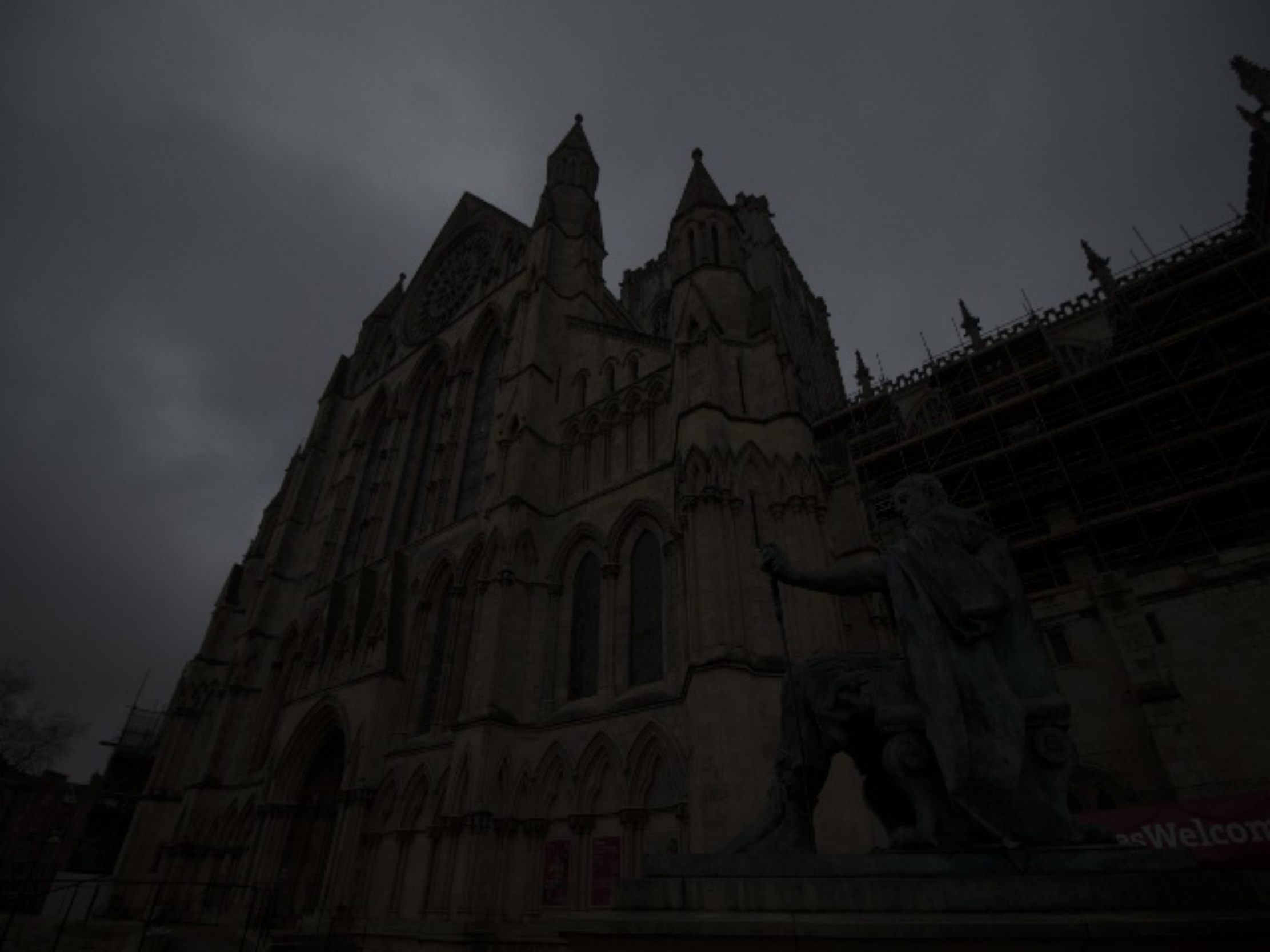}}
    \quad
	  \subfloat[Normal-Light Image]{\label{gt}
        \includegraphics[width=0.3\linewidth]{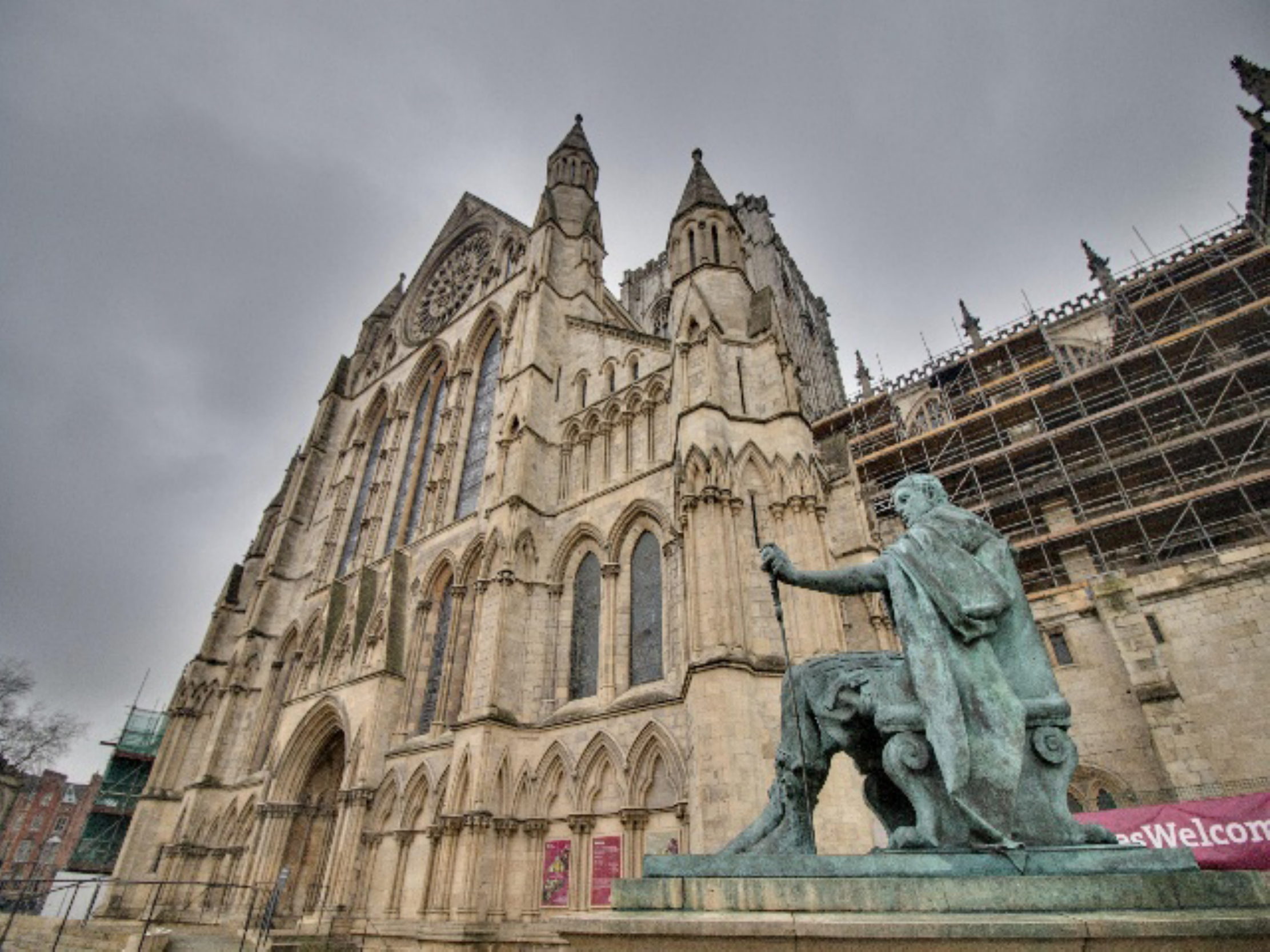}}
    \quad
	\subfloat[Regrouped Image]{\label{regrouped}
        \includegraphics[width=0.3\linewidth]{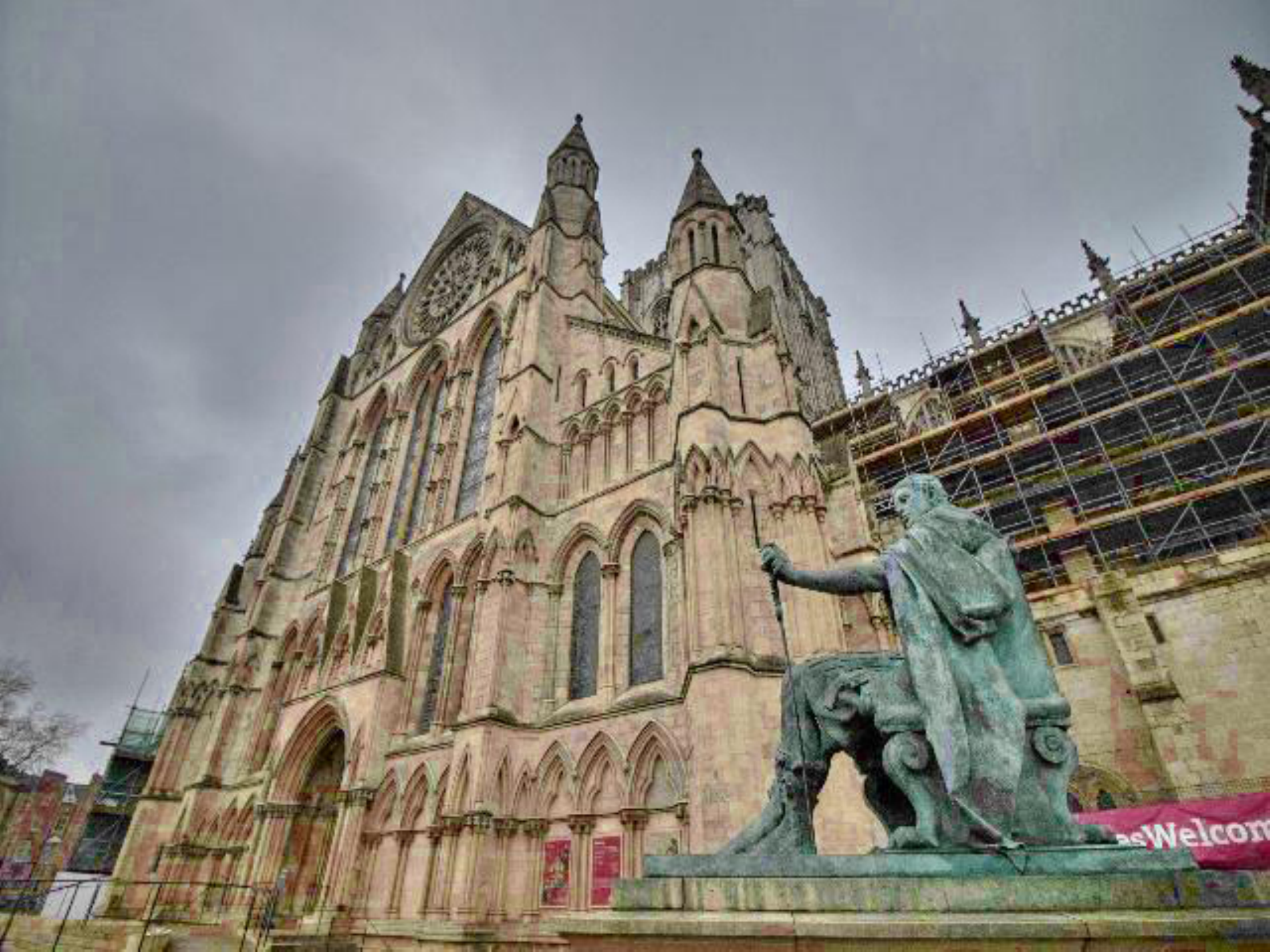}}
\caption{The visual quality result of the regrouped image generated by the H and S channels of low-light image and the V channel of its corresponding normal-light image.}
	\label{fig:hsv}
\end{figure}

Furthermore, we perform a simple experiment to illustrate the advantage of HSV color space. We randomly select a pair of low-/normal-light images from the Part2 of SICE dataset \cite{cai2018learning}, converting them into HSV color space. A regrouped image is generated by concatenating the H and S channels of the low-light image and the V channel of its corresponding normal-light image in the direction of channel, and converted back to RGB color space. As it shows in Fig.\ref{fig:hsv}, the regrouped image has comparable colors with the normal-light image. Therefore, for low-light image enhancement based on HSV color space, only the brightness needs to be enhanced, and the colors are completely derived from the low-light image.

\subsection{Random Brightness Disturbance}
\label{Random_disturbance}

To adapt to various lighting conditions, we propose to treat reflectance embedded in the brightness as enhanced brightness, since the Retinex theory \cite{land1977retinex} assumes that images of the same scene captured in different light conditions share the same reflectance. For the absence of ground truth, we design a random disturbance approach on the brightness (V channel), which achieves by a power function with a random exponent, to obtain another abnormal brightness of the same scene. This disturbed form should share the same reflectance as the original form.

Besides, the advantages of this nonlinear function are as follows: (1) The ranges of input and output of power function is [0,1] to avoid information loss by overflow truncation. (2) The power function is monotonically increasing to keep the consistency of gradient direction between the original and the disturbed form of brightness. (3) A random exponent is able to increase the diversity of disturbance.

Formally, this random brightness disturbance can be expressed as:
\begin{equation}
\label{equ:disturbance}
\mathbf{V^{'}(x)} =\mathbf{V(x)}^{\gamma},
\end{equation}
where $\mathbf{V(x)}$ denotes pixel values of original V channel, $\mathbf{V^{'}(x)}$ is the disturbed V channel, $\gamma$ is a random value. If the average value of $\mathbf{V}$ is lower than 0.5, the range of $\gamma$ is [0,1], otherwise [1,5].

Furthermore, it is worth noticing that this approach is not equivalent to data augmentation. The goal of our disturbance is to restrict the consistency of reflectance, instead of expanding training dataset. Even so, we do a relevant experiment in Sec.\ref{comp}.

\subsection{Reflectance Estimation Network}
\label{network}

We build a deep reflectance estimation network based on a Retinex variant, instead of learning a decomposition network to obtain reflectance and illumination as previous methods, reducing information loss. The original brightness $\mathbf{V}$ and the disturbed brightness $\mathbf{V^{'}}$ share the same network.
Besides, we define several loss functions to self-regularize the network. Next, we describe the Retinex variant, network architecture and loss functions of our network.

\subsubsection{Retinex Variant}
Retinex theory \cite{land1977retinex} aims to decompose an image into reflectance and illumination. The reflectance should be consistent under various lighting conditions of the same scene, while the illumination is assumed to be smooth. Specifically in our task, $\mathbf{V}$ and $\mathbf{V^{'}}$ can be decomposed by 
\begin{equation}
\label{equ:ori_ret_1}
\mathbf{V} =\mathbf{R} \circ \mathbf{I}
\end{equation}
and
\begin{equation}
\label{equ:ori_ret_2}
 \mathbf{V^{'}} =\mathbf{R^{'}} \circ \mathbf{I^{'}},
\end{equation}
where $\mathbf{R}$ and $\mathbf{I}$ are the reflectance and illumination of  $\mathbf{V}$, while $\mathbf{R^{'}}$ and $\mathbf{I^{'}}$ are the reflectance and illumination of  $\mathbf{V^{'}}$, and $\circ$ denotes element-wise product.

As we mentioned in Sec.\ref{Random_disturbance}, we treat the reflectance as well-exposed brightness. Therefore, we consider $\mathbf{I}$ as an intermediate variable for calculating $\mathbf{R}$. Formally, the Equ.\ref{equ:ori_ret_1} and Equ.\ref{equ:ori_ret_2} can be rewritten as 
\begin{equation}
\label{equ:var_ret_1}
\mathbf{R} =\mathbf{V} \circ \mathbf{L}
\end{equation}
and
\begin{equation}
\label{equ:var_ret_2}
 \mathbf{R^{'}} =\mathbf{V^{'}} \circ \mathbf{L^{'}},
\end{equation}
where $\mathbf{L}$ and $\mathbf{L^{'}}$ are the inverse of $\mathbf{I}$ and $\mathbf{I^{'}}$ \ie, $\mathbf{L} = 1/\mathbf{I}$, $\mathbf{L^{'}} = 1/\mathbf{I^{'}}$. 

The essential advantage of this Retinex variant is that it discards reconstruction loss used in most Retinex-based methods, which calculates the difference between $\mathbf{V}$ and $\mathbf{R} \circ \mathbf{I}$. Therefore, this can avoid information loss during the reconstruction process. Besides, it is worth noting that we directly generate the inverse of illumination in the network, instead of acquiring the illumination first and then calculating its inverse \cite{wang2019underexposed}.

\subsubsection{Network Architecture}

The reflectance estimation network takes the brightness as input and outputs the reflectance embedded in it. Note that $\mathbf{V}$ and $\mathbf{V^{'}}$ share the same network.

As is illustrated at the bottom of Fig.\ref{fig:network}, our reflectance estimation network adopts a typical U-Net \cite{ronneberger2015u} with 19 convolutional layers, four downsampling steps and four upsampling steps. Each downsampling step is a $2 \times 2$ max pooling operation with stride 2. And each upsampling step contains a bilinear interpolation to expand the height and width of the feature map to twice the original. Besides, two cascaded convolutional layers are included between two spatial resolution regulation operations. Each convolutional layer consists of a $3\times3$ convolution operation with padding, followed by a rectified linear unit (ReLU) activation function. We discard any batch normalization layers. In addition, skip connections concatenate the feature map of the downsampling process to the corresponding feature map of the upsampling process according to space resolution to increase the amount of information in the upsampling steps. Finally, the reflectance is obtained by an element-wise product between the input and the inverse of illumination marked as green in Fig\ref{fig:pipeline}.

\subsubsection{Loss Functions}
We define a set of loss functions that self-regularize the reflectance and the inverse of illumination, consisting of four following components.

\textbf{Reflectance Consistency Loss.}
According to Retinex theory \cite{land1977retinex}, the paired $\mathbf{\{V,V_d\}}$ should have the same reflectance. Therefore, to measure the difference between two generated reflectances, reflectance consistency loss $L_{rc}$ can be defined as: 
\begin{equation}
\label{equ:rc}
L_{rc} =  \|{\mathbf{R}-\mathbf{R^{'}}}\|_2^2,
\end{equation}
where $\mathbf{R}$ and $\mathbf{R^{'}}$ denotes the generated reflectances of $\mathbf{V}$ and $\mathbf{V^{'}}$, and $\|\cdot\|_2$ denotes the $l_2$ norm(MSE). 

\textbf{Exposure Control Loss.}
To constrain the general brightness of the generated reflectance, we set exposure control loss. It measures the distance between the average brightness of a local region of generated reflectance and a given well-exposed value $E$. $E$ is set to 0.7, which is slightly different from Zero-DCE \cite{guo2020zero}. Formally, the exposure control loss $L_{ec}$ can be expressed as:
\begin{equation}
\label{equ:ec}
L_{ec} =  \|{\mathbf{R_n}-\mathbf{E}}\|_2^2,
\end{equation}
where $\mathbf{R_n}$ denotes the result of $n \times n$ average pooling of the generated reflectance and $\mathbf{E}$ denotes a matrix with the same size as $\mathbf{R_n}$ and all 0.7 values. We set $n=16$ in the experiment to keep the balance of brightness and details empirically. 

\textbf{Spatial Structure Loss.}
To inherit the spatial structure of input onto the generated reflectance, we introduce spatial structure loss, which evaluates the difference between the horizontal and vertical gradients of each pixel of the input and the generated reflectance. Thus, spatial structure loss $L_{ss}$ is formalized as:
\begin{equation}
\label{equ:ss}
L_{ss} =  \|{\nabla\mathbf{R_m}-\nabla\mathbf{V_m}}\|_2^2,
\end{equation}
where $\mathbf{R_m}$ and $\mathbf{V_m}$ denote the result of $m \times m$ average pooling of the generated reflectance and input, respectively, and $\nabla$ stands for the first-order difference operation including $\nabla_x$ (horizontal) and $\nabla_y$ (vertical) directions. We set $m=4$ empirically.

Noted that although spatial structure loss is analogous to spatial consistency loss in Zero-DCE \cite{guo2020zero}, we improve it by considering the gradient direction.

\textbf{Illumination Smoothness Loss.}
According to Retinex theory, the illumination should be smooth, by which all the details can be preserved on the reflectance. The same applies to the inverse of illumination. Total variation loss (TV) \cite{chan2011augmented}, which calculates the gradient of adjacent pixels, is suitable for smoothness loss. Therefore, illumination smoothness loss $L_{is}$ can be defined as:
\begin{equation}
\label{equ:is}
L_{is} =  \|{\nabla\mathbf{L}\|}_2^2+\|{\nabla\mathbf{L^{'}}\|}_2^2,
\end{equation}
where $\mathbf{L}$ and $\mathbf{L^{'}}$ denote the generated illumination component of $\mathbf{V}$ and $\mathbf{V^{'}}$ in inverse, and $\nabla$ is the same as Equ.\ref{equ:ss}.

\textbf{Total Loss.} Finally, the total loss can be expressed as:

\begin{equation}
\label{equ:total}
L = L_{rc} + L_{ec}+L_{ss}+W_{is}L_{is},
\end{equation}
where $W_{is}$ is the weight of illumination smoothness loss. Noted that the exposure control loss and spatial structure loss of $\mathbf{R^{'}}$ are not calculated repeatedly, since we have restricted $\mathbf{R}$ and $\mathbf{R^{'}}$ to be consistent by reflectance consistency loss. 

\section{Experiments}
\label{Experiments}

In this section, we conduct extensive experiments. To start with, the implementation details of our method are present in Sec.\ref{Implementation}. Besides, we compare our method with multiple state-of-the-art methods quantitatively and qualitatively in Sec.\ref{comp}. What's more, we perform ablation studies in Sec.\ref{Ablation} to demonstrate the effectiveness of the loss functions and the brightness disturbance approach. Finally, we show the adaptability of our method and Zero-DCE \cite{guo2020zero} under various lighting conditions in Sec.\ref{Adapt}.

\subsection{Implementation Details}
\label{Implementation}

For a fair comparison with other methods, we employ the same training dataset as Zero-DCE \cite{guo2020zero} and EnlightenGAN \cite{jiang2019enlightengan}. From 360 multi-exposed image sequences in the Part1 of SICE dataset \cite{cai2018learning}, we randomly select one image in the first third of each sequence (corresponding to the low-light image) for validation and the rest for training. All the images are resized to $512\times512$. Besides, each sequence of the Part1 of SICE dataset \cite{cai2018learning} collects under-exposed and over-exposed images. As is mentioned in \cite{guo2020zero}, the usage of over-exposed images is beneficial for under-exposed image enhancement task.

Furthermore, we adopt similar training settings as Zero-DCE \cite{guo2020zero}.  We implement our method with PyTorch on an NVIDIA 1080Ti GPU. The batch size is set to 8. The filter weights of each layer are initialized by kaiming initialization \cite{he2015delving}. Bias is initialized as a constant. We use ADAM optimizer with default parameters. The learning rate is set to $10^{-4}$. We conduct one disturbed brightness for each input and set the weight $W_{is}$ to 10. We train our method for 500 epochs and evaluate it every 50 epochs, treating the best model as our final model.

\subsection{Comparison with State-of-the-Art Methods}
\label{comp}
We compare our method with several state-of-the-art (SOTA) methods, including two conventional methods(LIME \cite{guo2016lime}, Li \etal \cite{li2018structure}), one supervised learning method (RetinexNet \cite{wei2018deep}), one GAN-based method (EnlightenGAN \cite{jiang2019enlightengan}) and one self-regularized learning method (Zero-DCE \cite{guo2020zero}). 

For quantitative evaluation, we employ the Part2 of SICE dataset \cite{cai2018learning}. It contains 229 multi-exposed sequences and the corresponding reference image for each sequence. We select the first three or four low-light images if there are seven or nine images in a multi-exposure sequence, discarding seven unaligned sequences (127, 169, 193, 194, 195, 208 and 212) and resize all the images to $640\times480$. Besides, we conduct qualitative experiments on four publicly available image datasets used by previous works including LIME \cite{guo2016lime}, MEF \cite{ma2015perceptual}, NPE \cite{wang2013naturalness} and DICM \cite{lee2012contrast}.

\subsubsection{Quantitative Comparison}

\begin{table}[t]
	\caption{Quantitative comparison with SOTA methods in terms of PSNR and SSIM. The optimal results are highlighted in red whereas the second-optimal results are highlighted in blue.}
	\centering
	\begin{tabular}{c|c c}
		\hline
		\textbf{Method} & \textbf{PSNR$\uparrow$ } & \textbf{SSIM$\uparrow$}   \\
		\hline
		LIME \cite{guo2016lime}                    &  14.77     & 0.512 \\
		Li \etal \cite{li2018structure}           &  14.80     &   0.467 \\
		RetinexNet \cite{wei2018deep}              &  16.15     & 0.499  \\
		EnlightenGAN \cite{jiang2019enlightengan}  &  15.87      &  \color{red}0.546  \\
		Zero-DCE \cite{guo2020zero}                &  16.28      & 0.525	  \\
       Ours-half                                 &  \color{blue}16.99      & \color{blue}0.530 \\
		Ours                                      &  \color{red}17.06      & \color{blue}0.530  \\
		\hline
	\end{tabular}
	\label{table:comp}
\end{table}

For quantitative comparison, full reference metrics Peak Signal to Noise Ratio (PSNR) and Structure SIMilarity index (SSIM) \cite{wang2004image} are used to measure signal and structure fidelity. As reported in Table \ref{table:comp}, our method wins first place in PSNR and second place in SSIM. It is worth noting that RetinexNet \cite{wei2018deep} and EnlightenGAN \cite{jiang2019enlightengan} employ paired and unpaired low-/normal-light images in training. The normal-light images provide additional information. As for self-regularized learning, our method outperforms Zero-DCE about 0.78dB in PSNR, demonstrating the effectiveness of our method.

Besides, we consider the size of the training dataset. To demonstrate the brightness disturbance approach is not equivalent to data augmentation, we train another model with half of the training data. We report the performance at the penultimate row of Table \ref{table:comp}, namely Ours-half. It has nearly the same performance as the model that uses the full dataset and surpasses Zero-DCE about 0.71dB in PSNR.

\subsubsection{Qualitative Comparison}
\begin{figure*}[t]
    \centering
	  \subfloat[Input]{ \label{qua_gt:51_input}
       \includegraphics[width=0.22\linewidth]{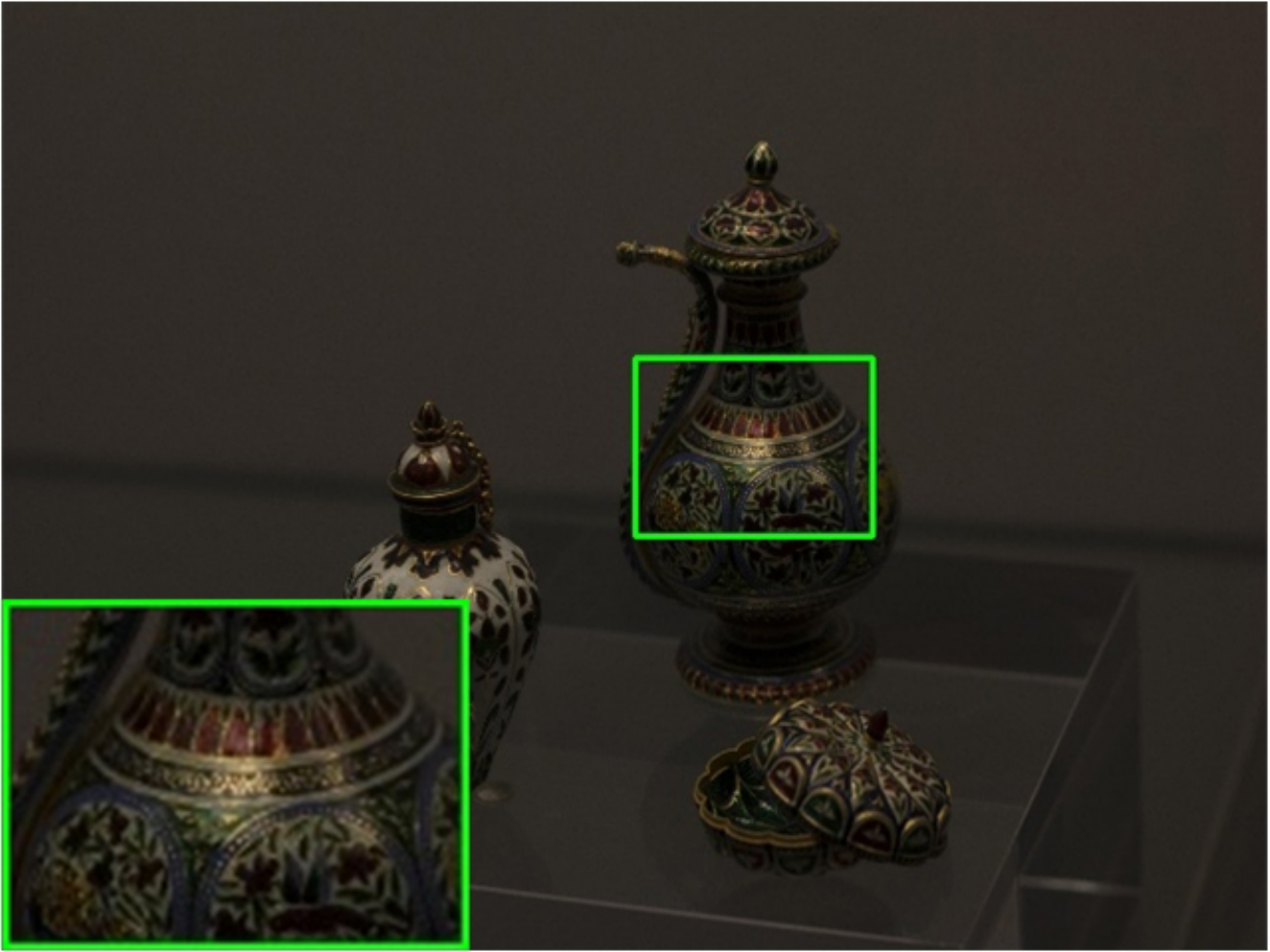}}
	  \quad
	  \subfloat[LIME \cite{guo2016lime}]{\label{qua_gt:51_lime}
        \includegraphics[width=0.22\linewidth]{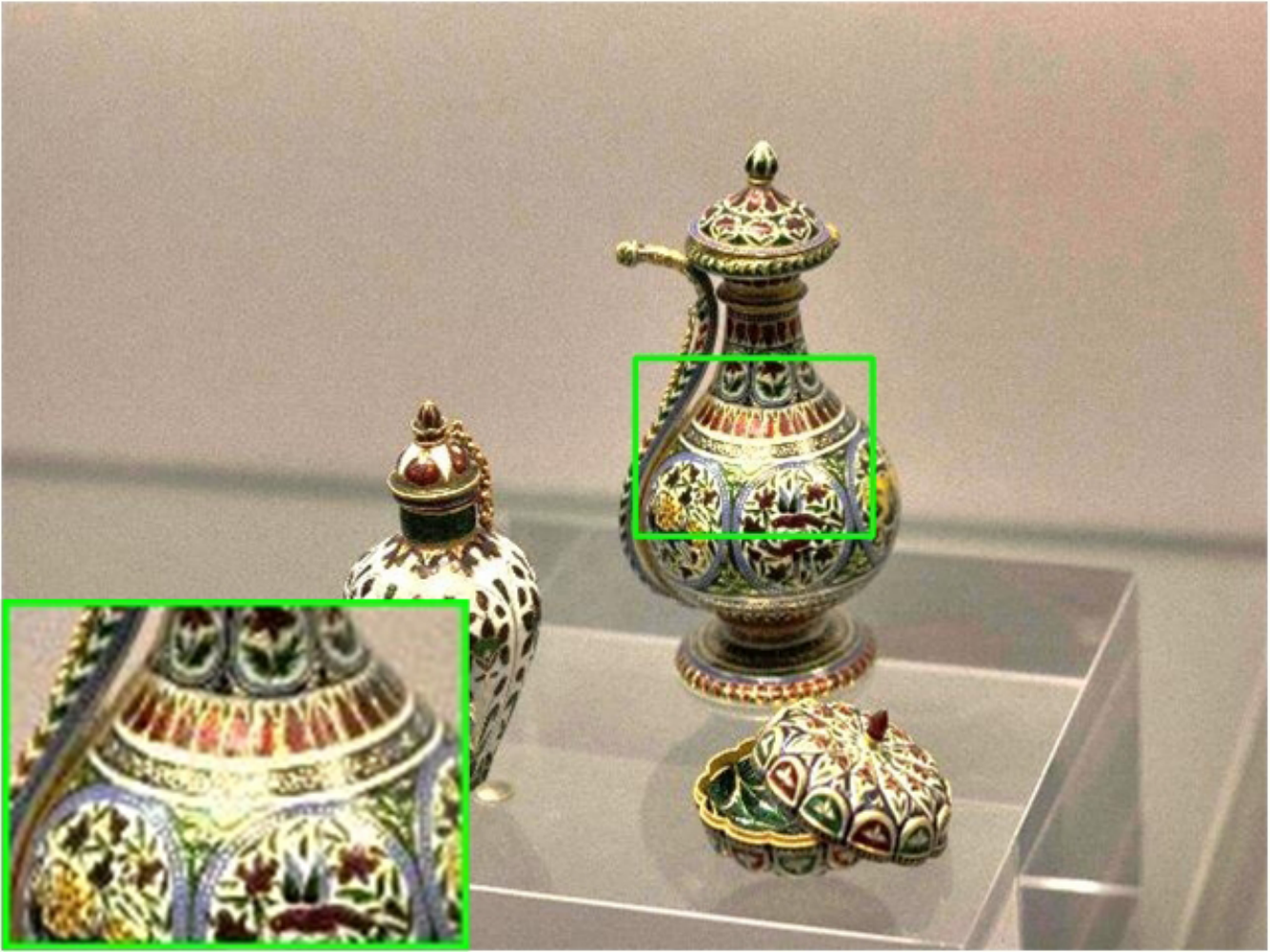}}
      \quad
	  \subfloat[Li \etal \cite{li2018structure}]{ \label{qua_gt:51_li}
       \includegraphics[width=0.22\linewidth]{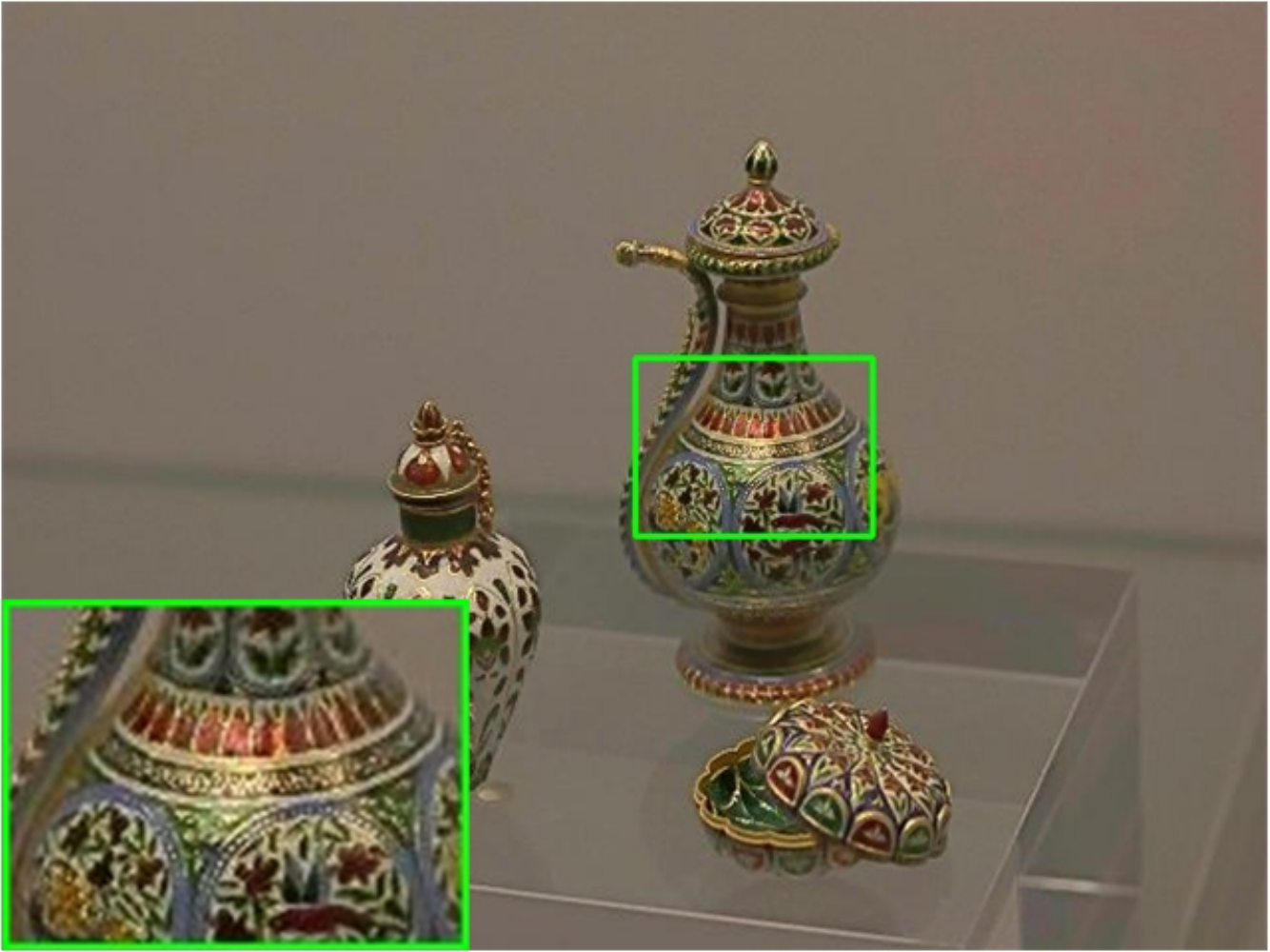}}
	  \quad
	  \subfloat[RetinexNet \cite{wei2018deep}]{\label{qua_gt:51_retinex}
        \includegraphics[width=0.22\linewidth]{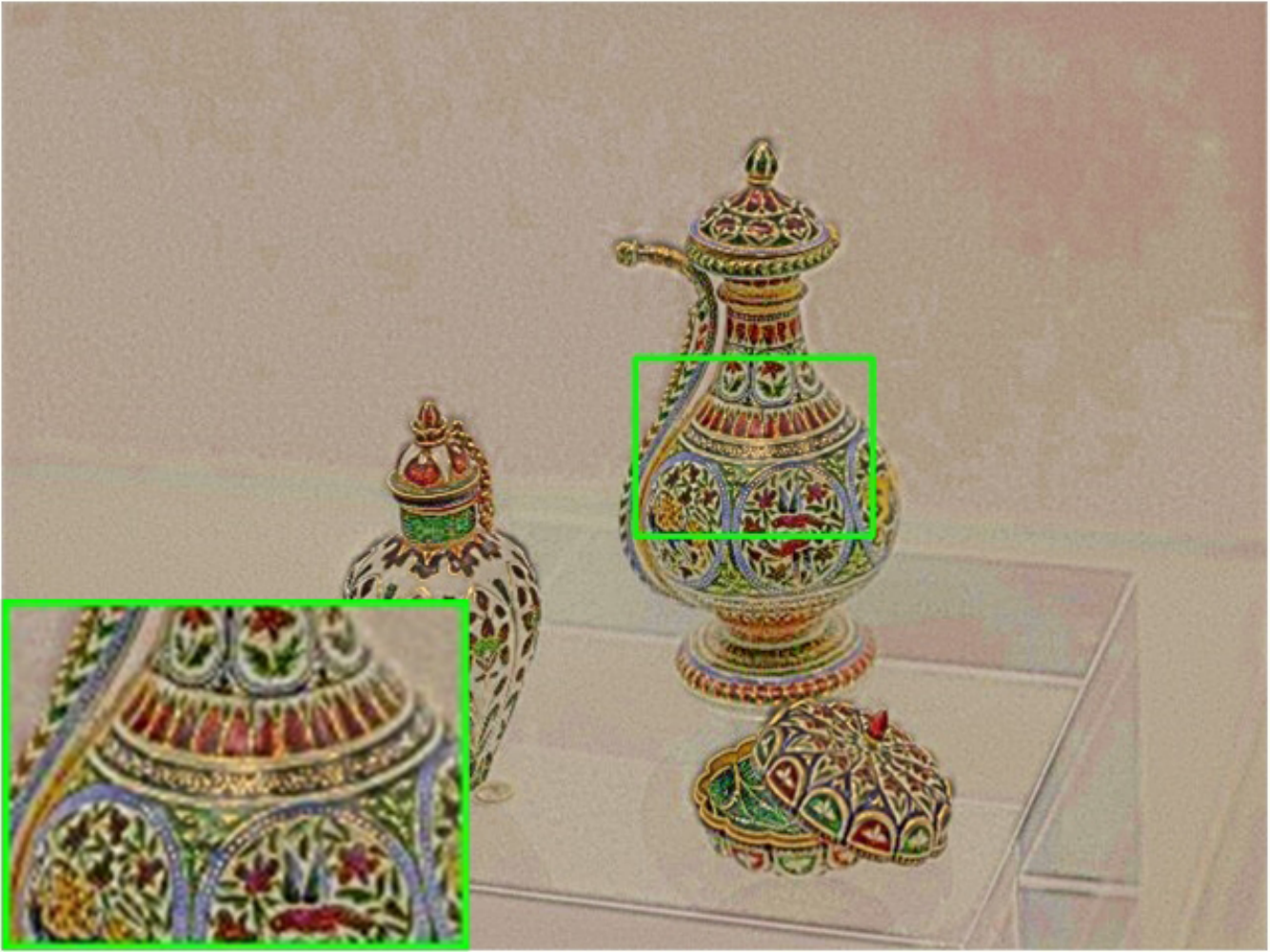}}\\
	  \subfloat[EnlightenGAN \cite{jiang2019enlightengan}]{\label{qua_gt:51_enlightengan}
        \includegraphics[width=0.22\linewidth]{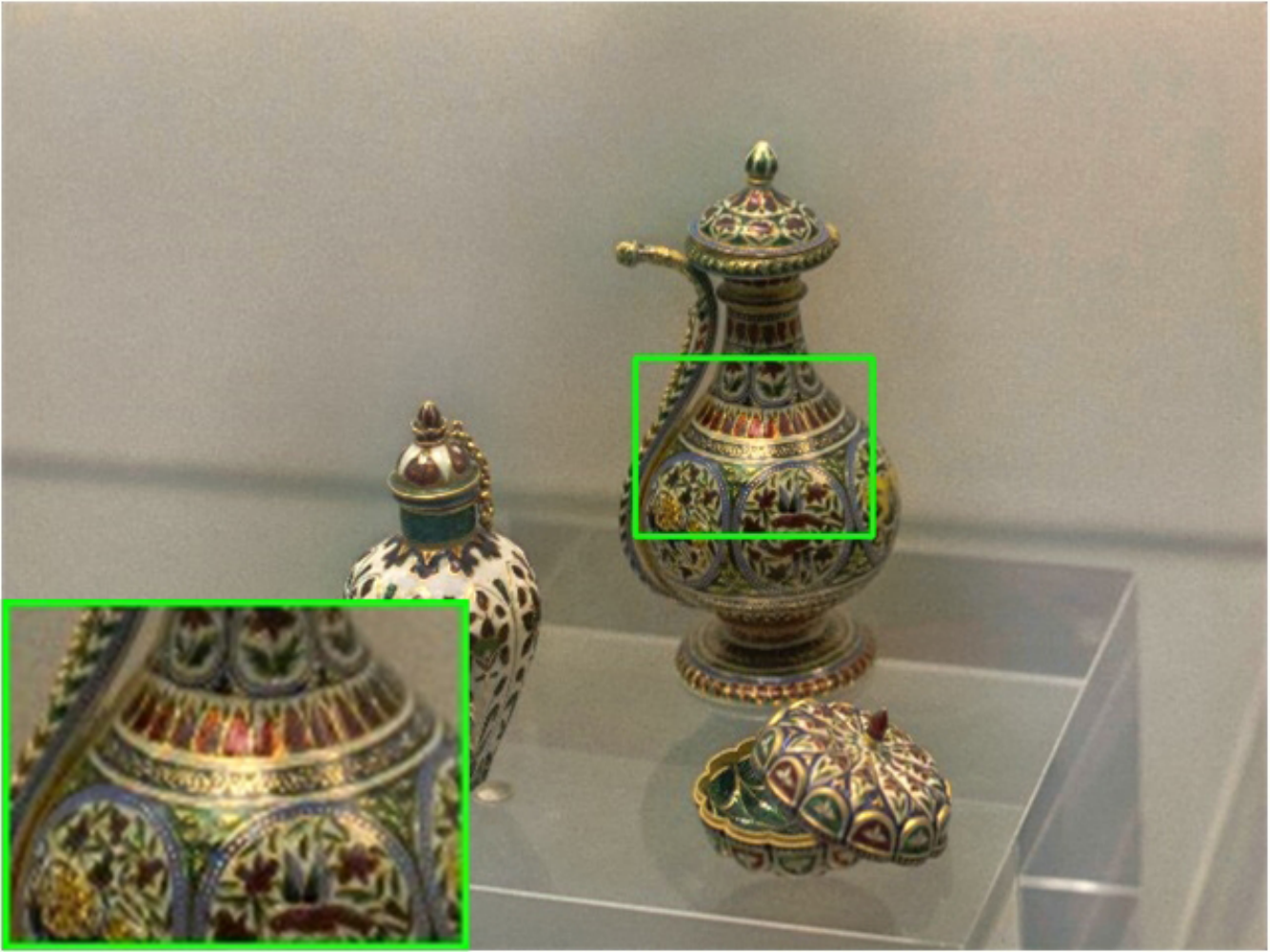}}
      \quad
     \subfloat[Zero-DCE \cite{guo2020zero}]{ \label{qua_gt:51_zerodce}
       \includegraphics[width=0.22\linewidth]{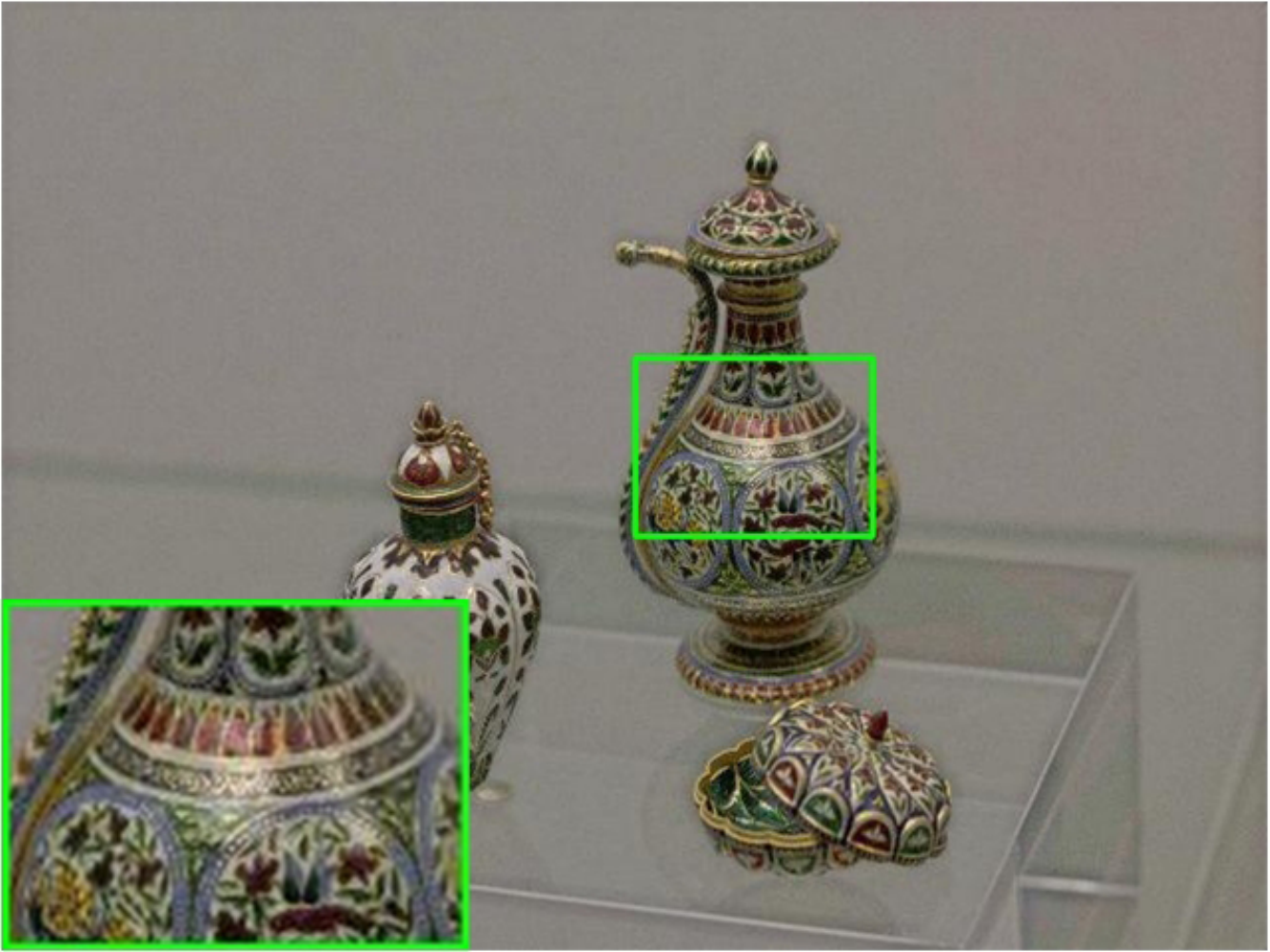}}
	  \quad
	  \subfloat[Ours]{\label{qua_gt:51_ours}
        \includegraphics[width=0.22\linewidth]{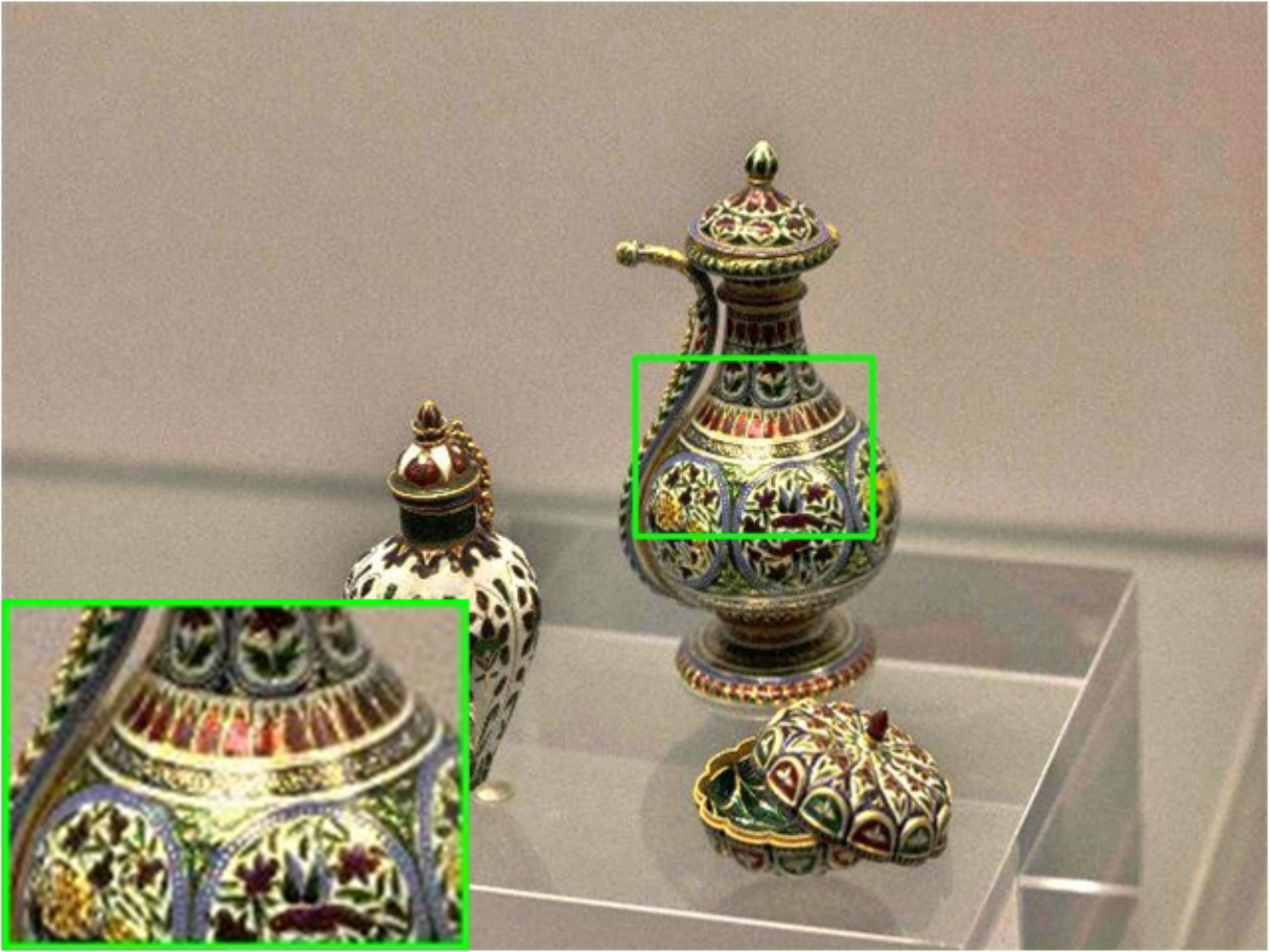}}
	  \quad
	  \subfloat[Ground Truth]{\label{qua_gt:51_gt}
        \includegraphics[width=0.22\linewidth]{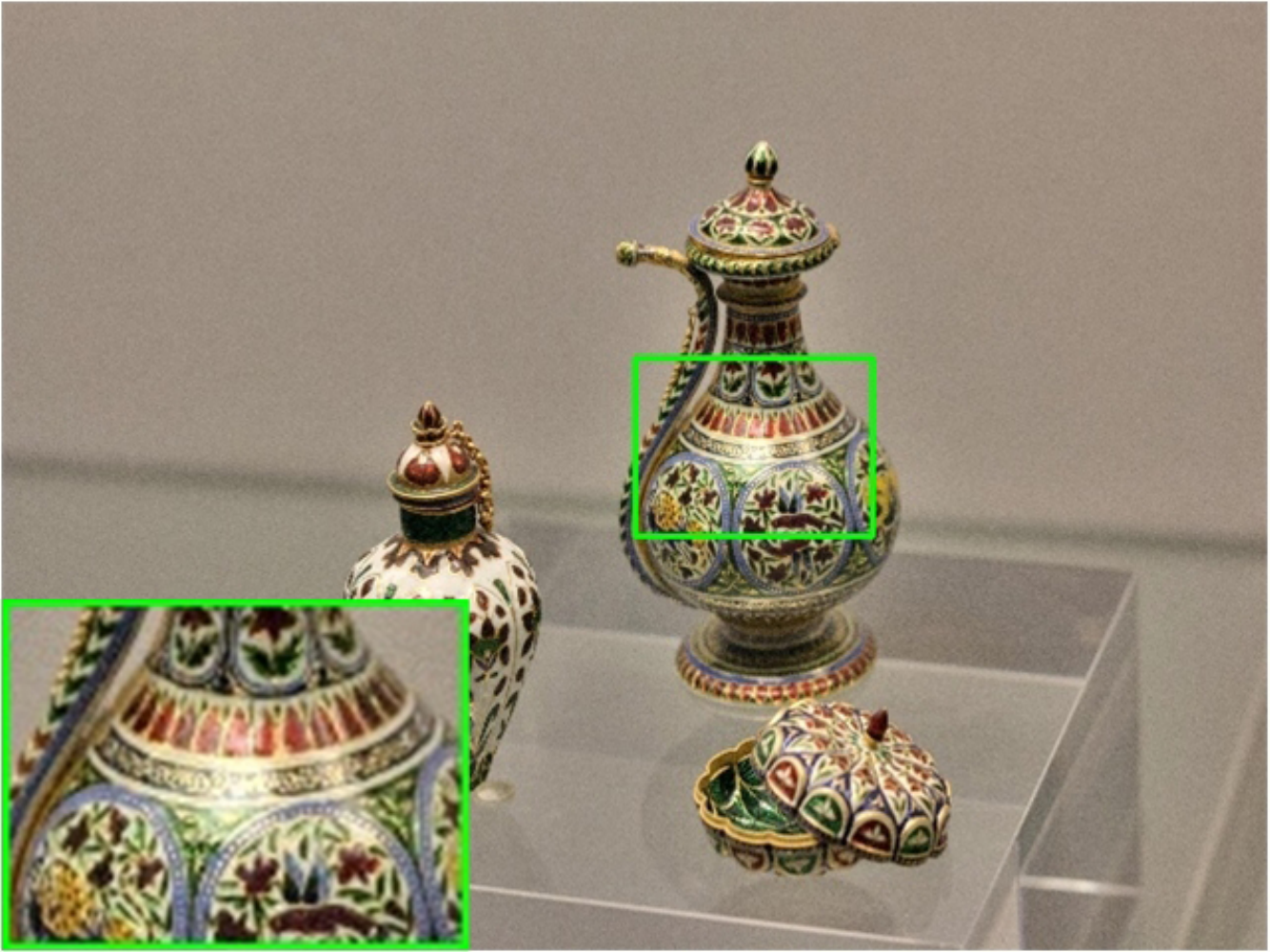}}\\
      \subfloat[Input]{ \label{qua_gt:90_input}
       \includegraphics[width=0.22\linewidth]{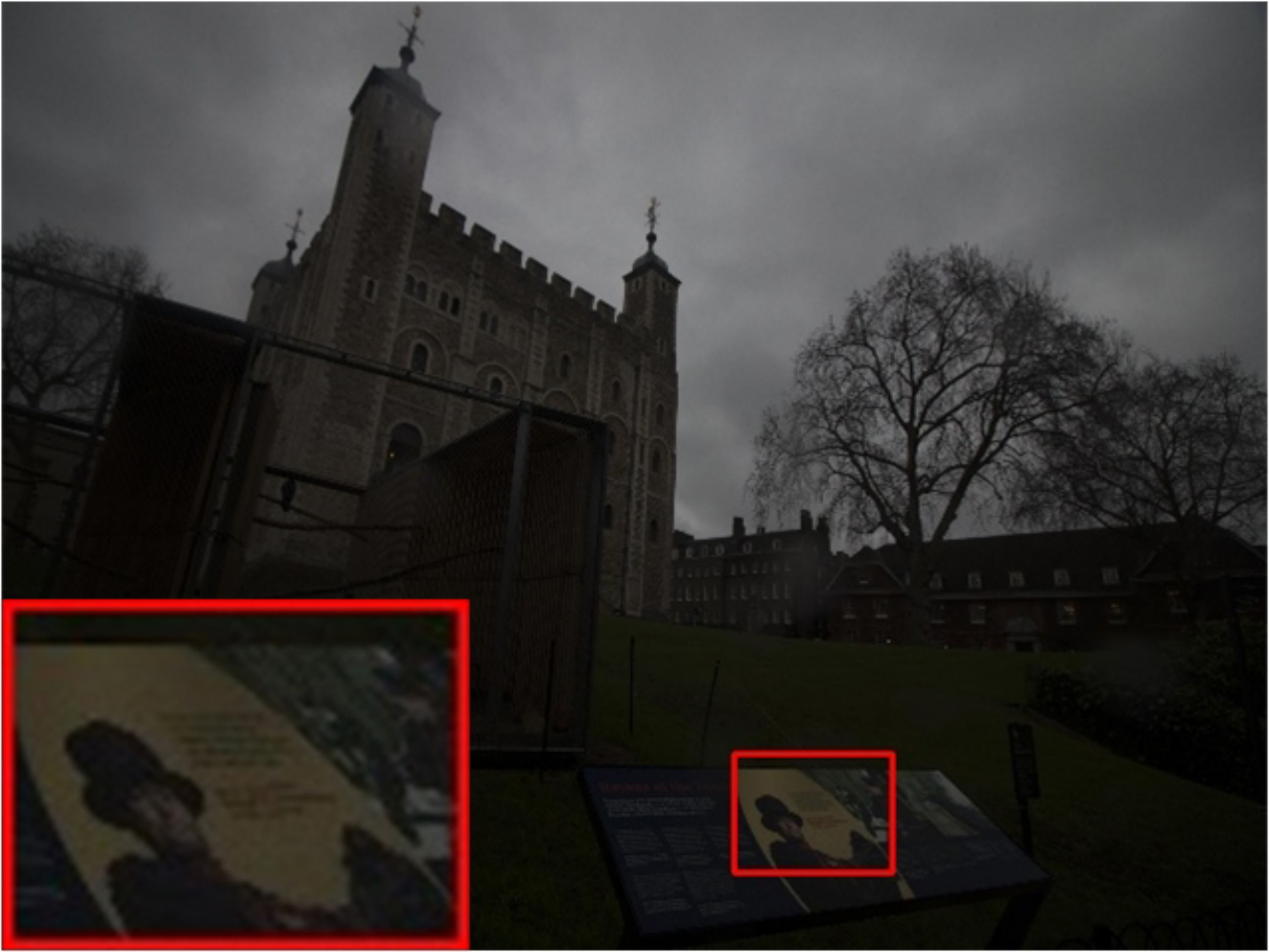}}
	  \quad
	  \subfloat[LIME \cite{guo2016lime}]{\label{qua_gt:90_lime}
        \includegraphics[width=0.22\linewidth]{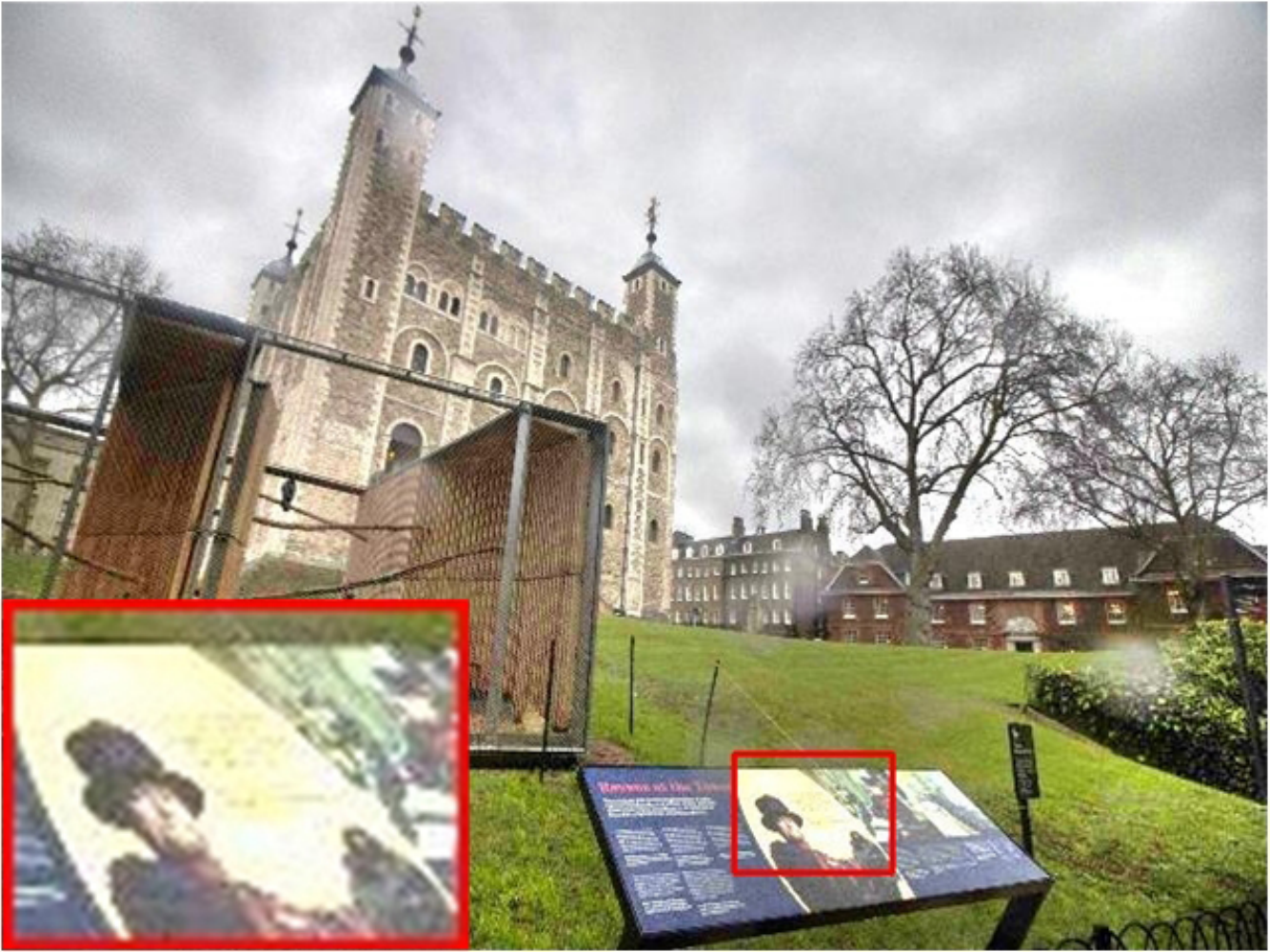}}
      \quad
	  \subfloat[Li \etal \cite{li2018structure}]{ \label{qua_gt:90_li}
       \includegraphics[width=0.22\linewidth]{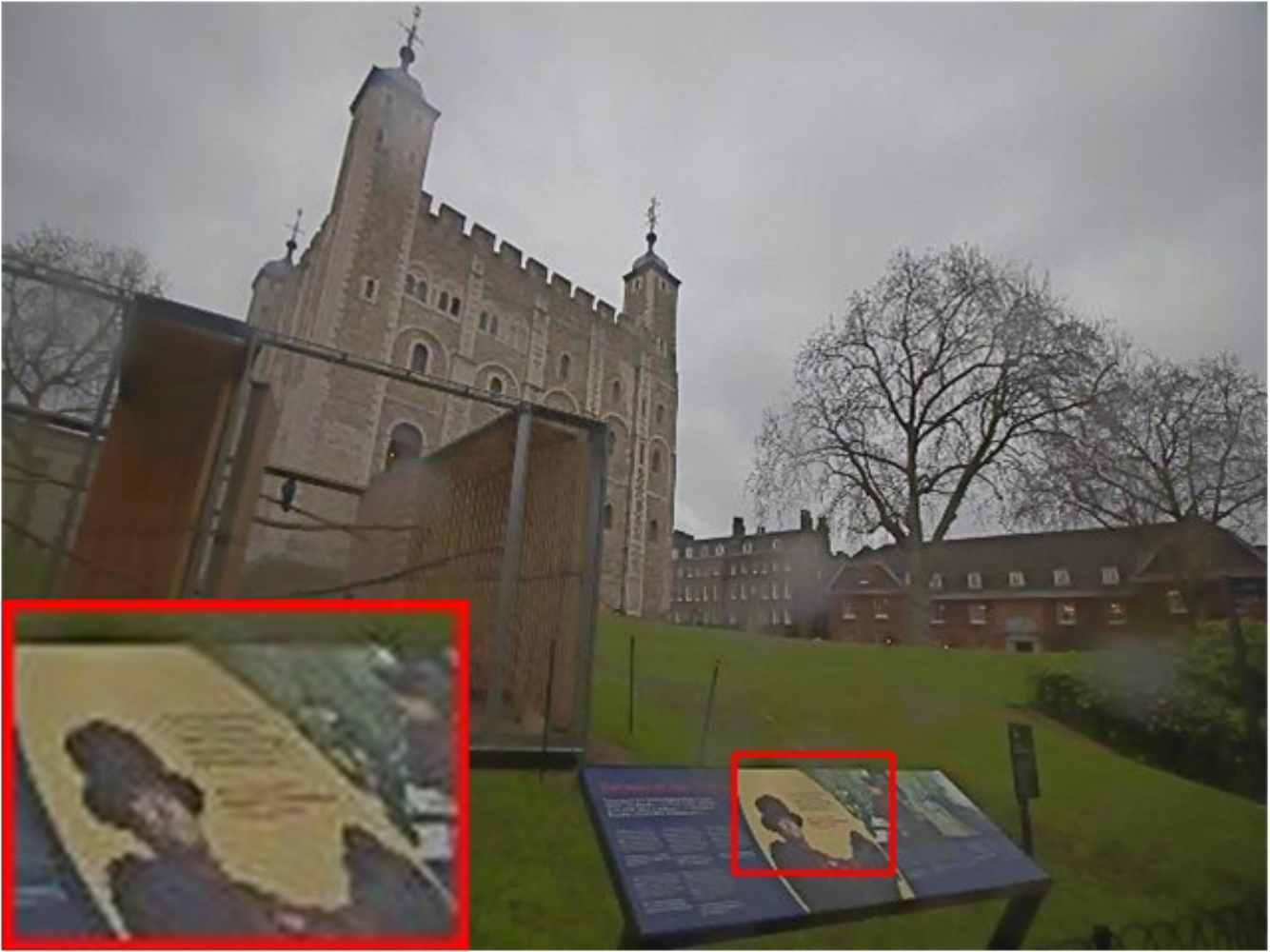}}
	  \quad
	  \subfloat[RetinexNet \cite{wei2018deep}]{\label{qua_gt:90_retinex}
        \includegraphics[width=0.22\linewidth]{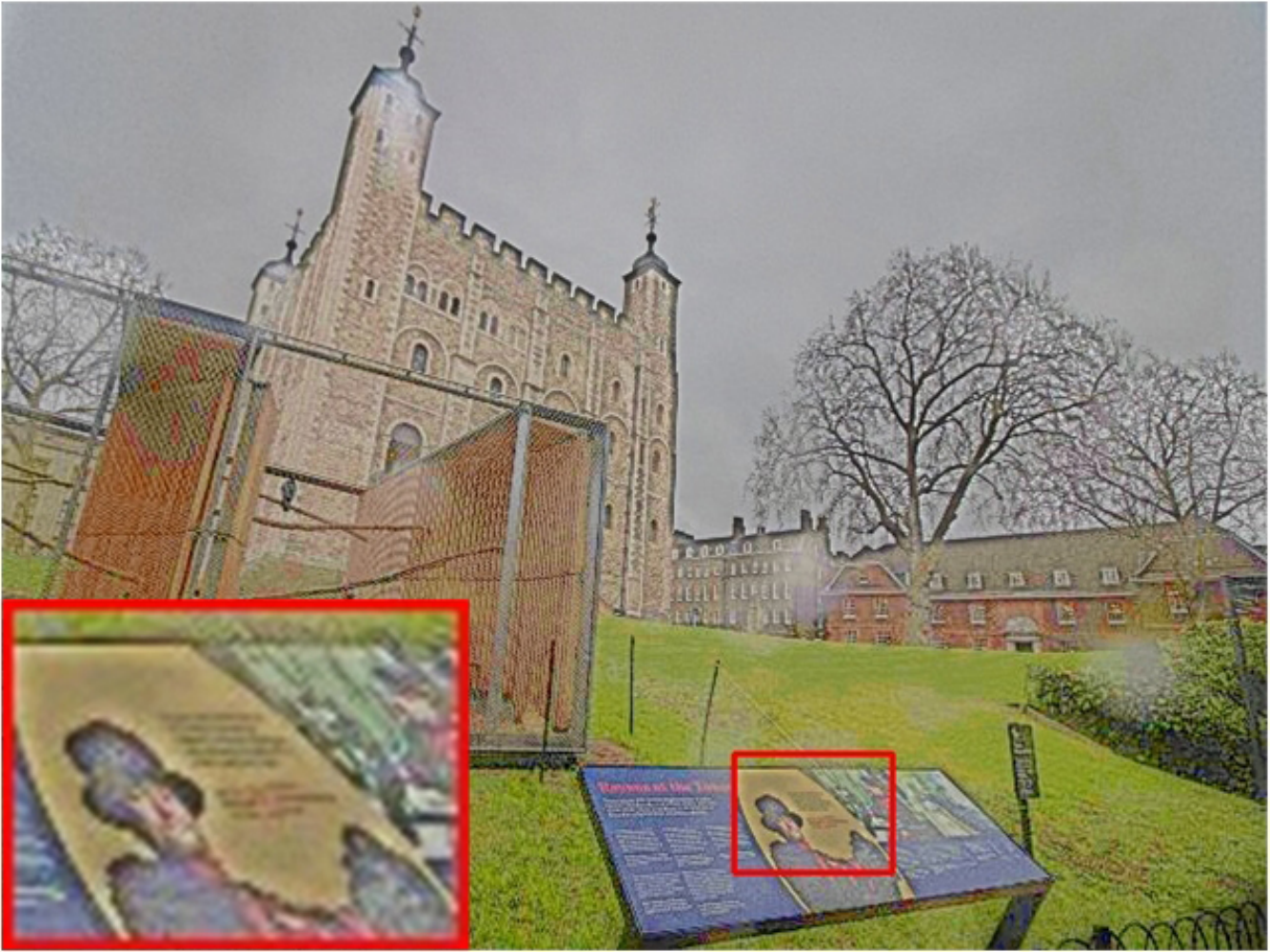}}\\
	  \subfloat[EnlightenGAN \cite{jiang2019enlightengan}]{\label{qua_gt:90_enlightengan}
        \includegraphics[width=0.22\linewidth]{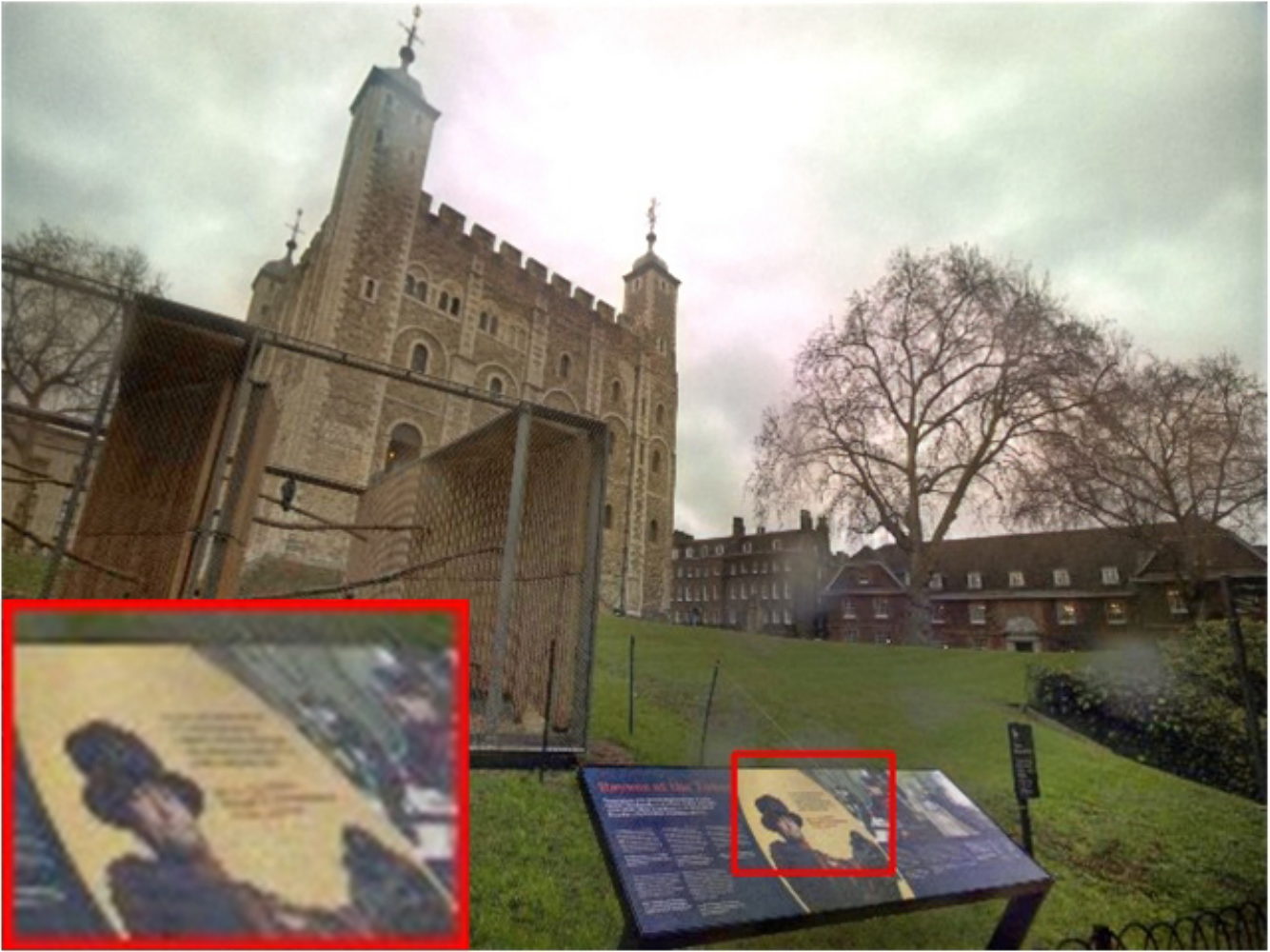}}
      \quad
     \subfloat[Zero-DCE \cite{guo2020zero}]{ \label{qua_gt:90_zerodce}
       \includegraphics[width=0.22\linewidth]{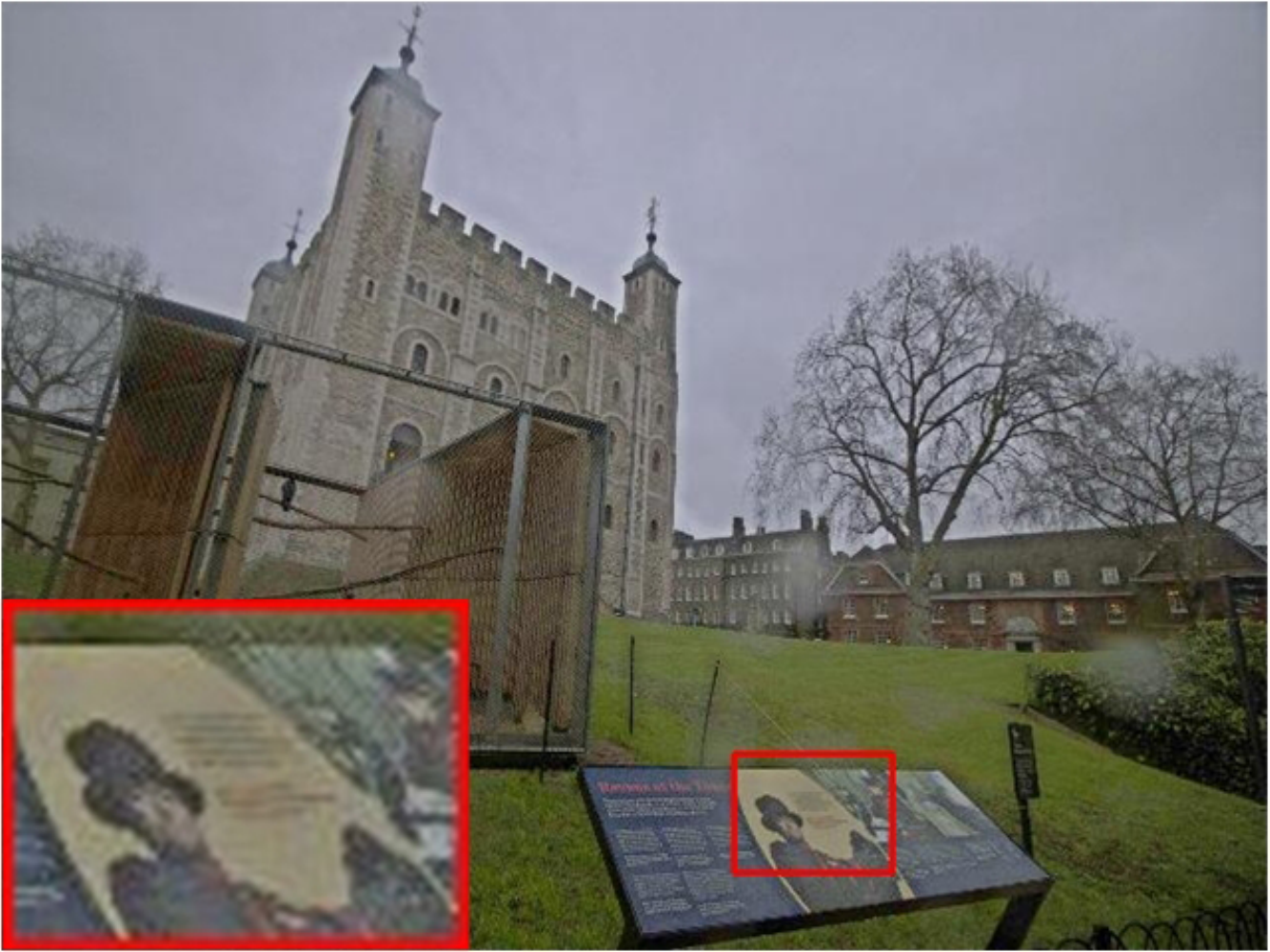}}
	  \quad
	  \subfloat[Ours]{\label{qua_gt:90_ours}
        \includegraphics[width=0.22\linewidth]{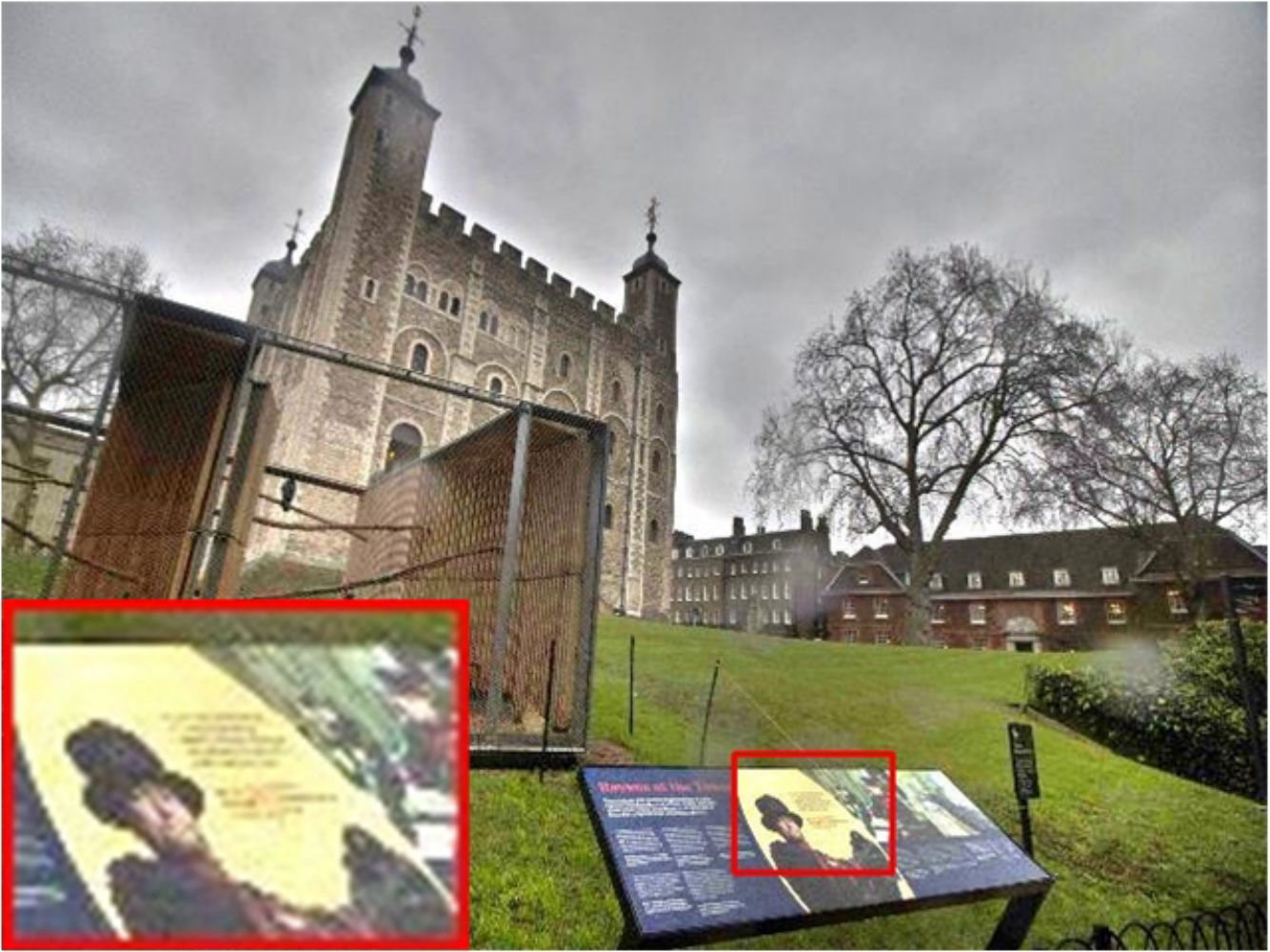}}
	  \quad
	  \subfloat[Ground Truth]{\label{qua_gt:90_gt}
        \includegraphics[width=0.22\linewidth]{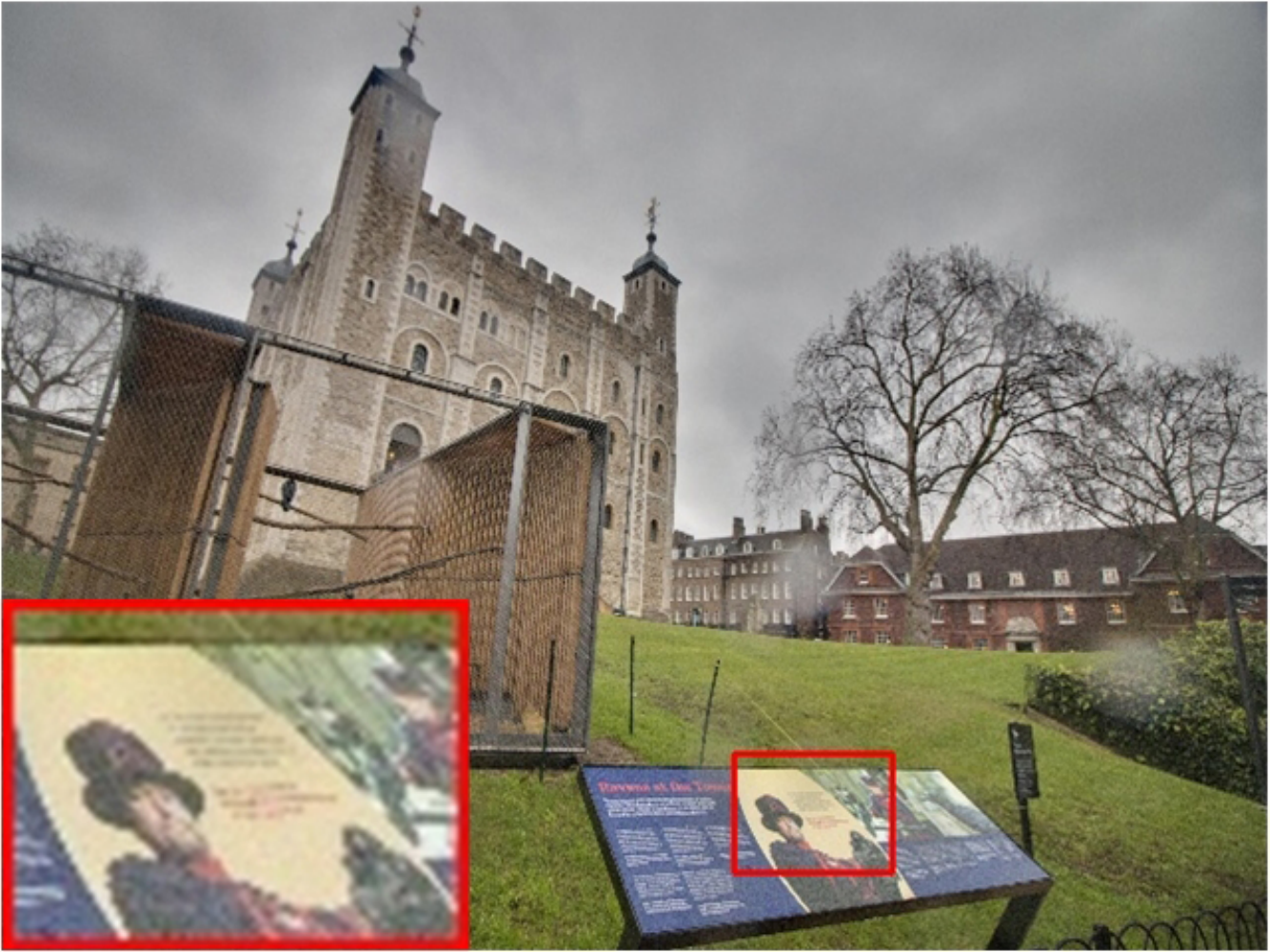}}
	  \caption{Visual comparisons with SOTA methods on the Part2 of SICE dataset \cite{cai2018learning}.}
	  \label{fig:qua_gt} 
\end{figure*}

Fig.\ref{fig:qua_gt} shows the results of different methods on two typical low-light images of the Part2 of SICE dataset \cite{cai2018learning}. We evaluate the visual quality in terms of brightness and color. For conventional methods, LIME \cite{guo2016lime} can generate pleasant visual performance, but the bright regions are over-exposed and the details are lost (\eg, the words in the red box of Fig.\ref{qua_gt:90_lime}). Although Li \etal \cite{li2018structure} enhances the low-light images to some extent, they are still under-exposed. As for deep learning-based methods, RetinexNet \cite{wei2018deep} generates weird colors and details. EnlightenGAN \cite{jiang2019enlightengan} can obtain satisfying results, but the color is mixed with yellow (\eg, the building and exhibits in Fig.\ref{qua_gt:90_enlightengan} and Fig.\ref{qua_gt:51_enlightengan}). Zero-DCE \cite{guo2020zero} has similar performance with Li \etal. It fails to enhance low-light images sufficiently. Compared to other methods, our method outperforms in terms of brightness and faithfully restores the color hidden in the low-light image.

\begin{figure*}[t]
    \centering
	  \subfloat[Input]{ \label{qua_no_gt:DICM_input}
       \includegraphics[width=0.18\linewidth]{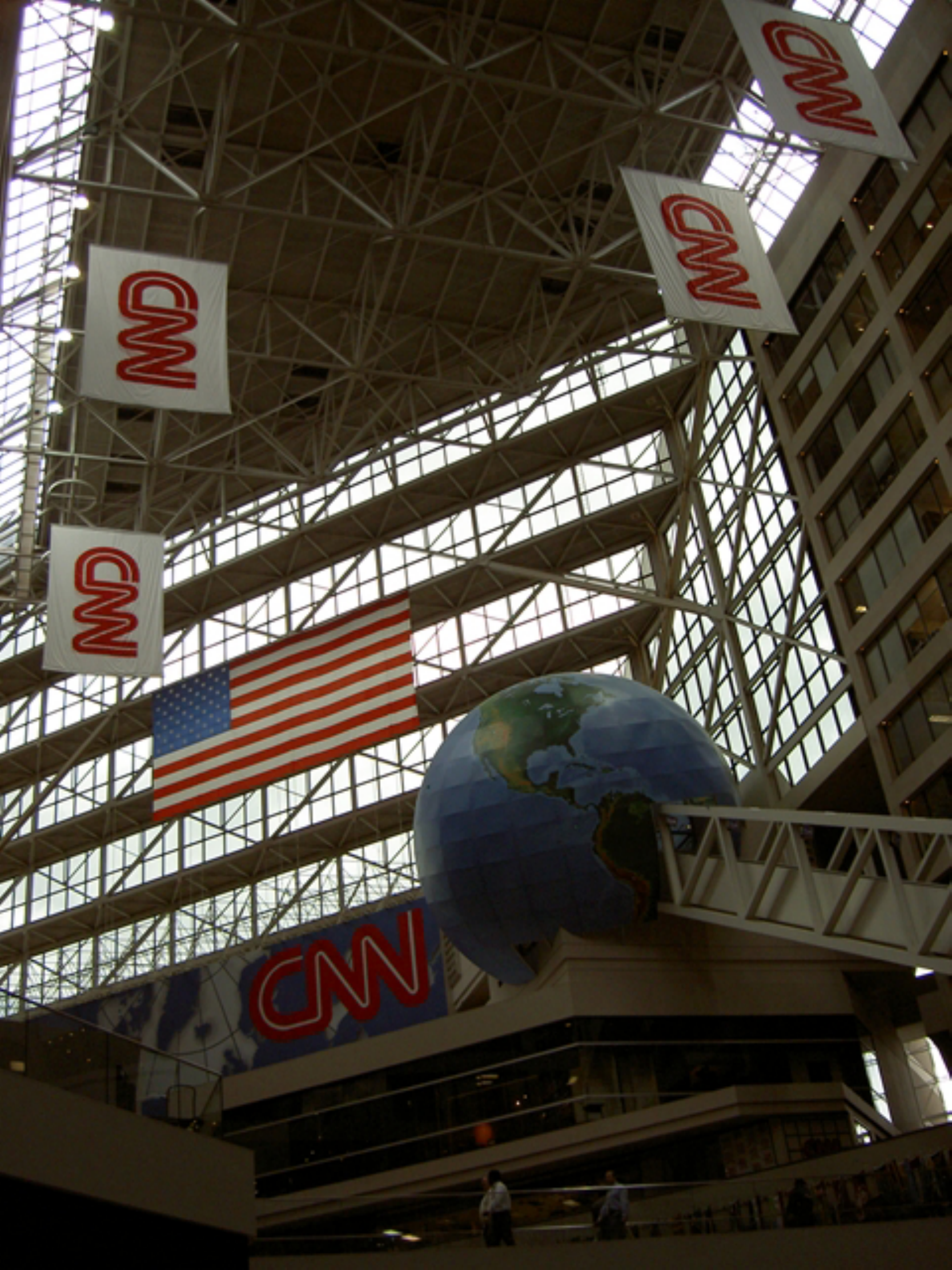}}
	  \quad
	  \subfloat[LIME \cite{guo2016lime}]{\label{qua_no_gt:DICM_lime}
        \includegraphics[width=0.18\linewidth]{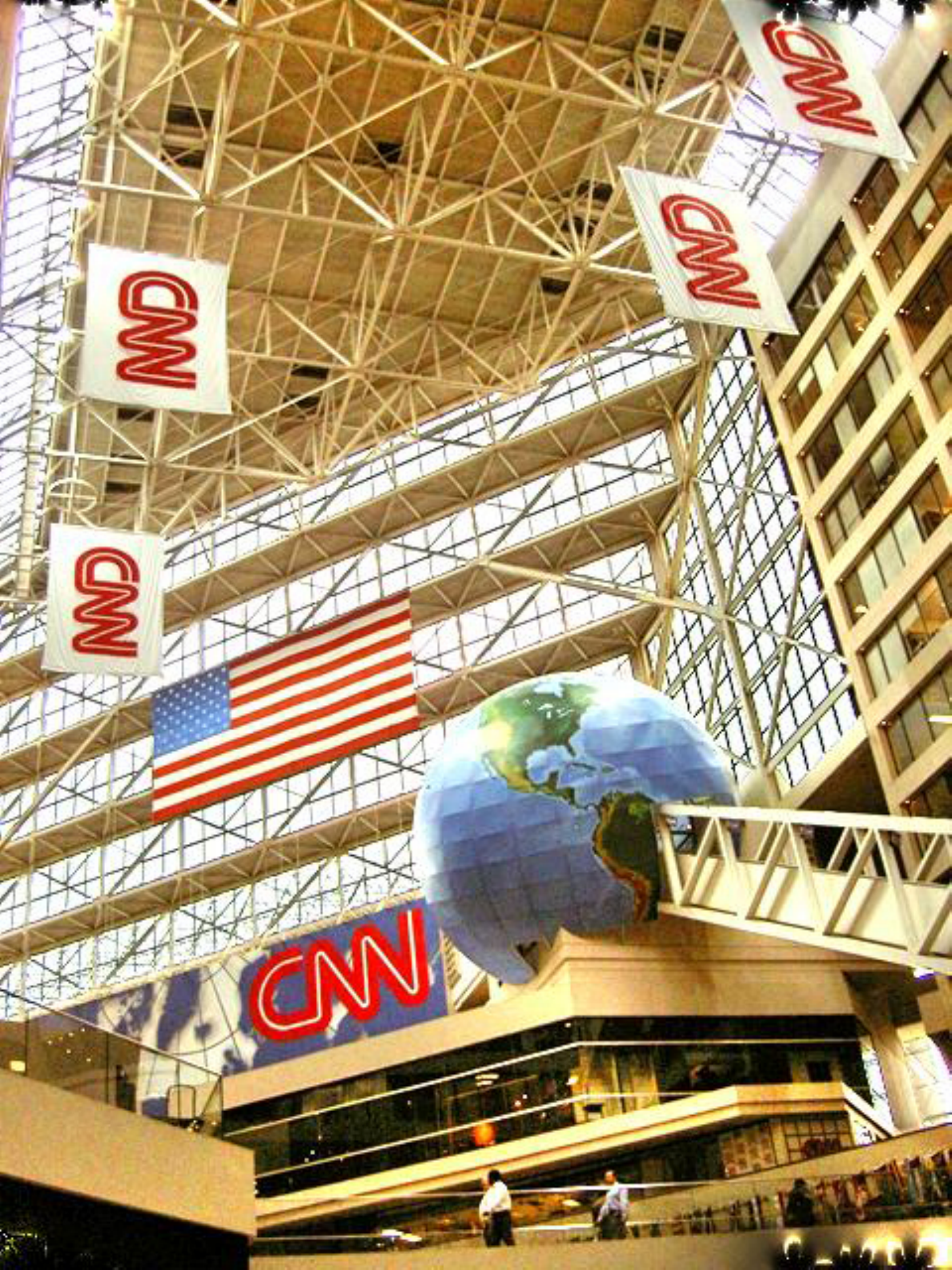}}
	  \quad
	  \subfloat[EnlightenGAN \cite{jiang2019enlightengan}]{\label{qua_no_gt:DICM_enlighten}
        \includegraphics[width=0.18\linewidth]{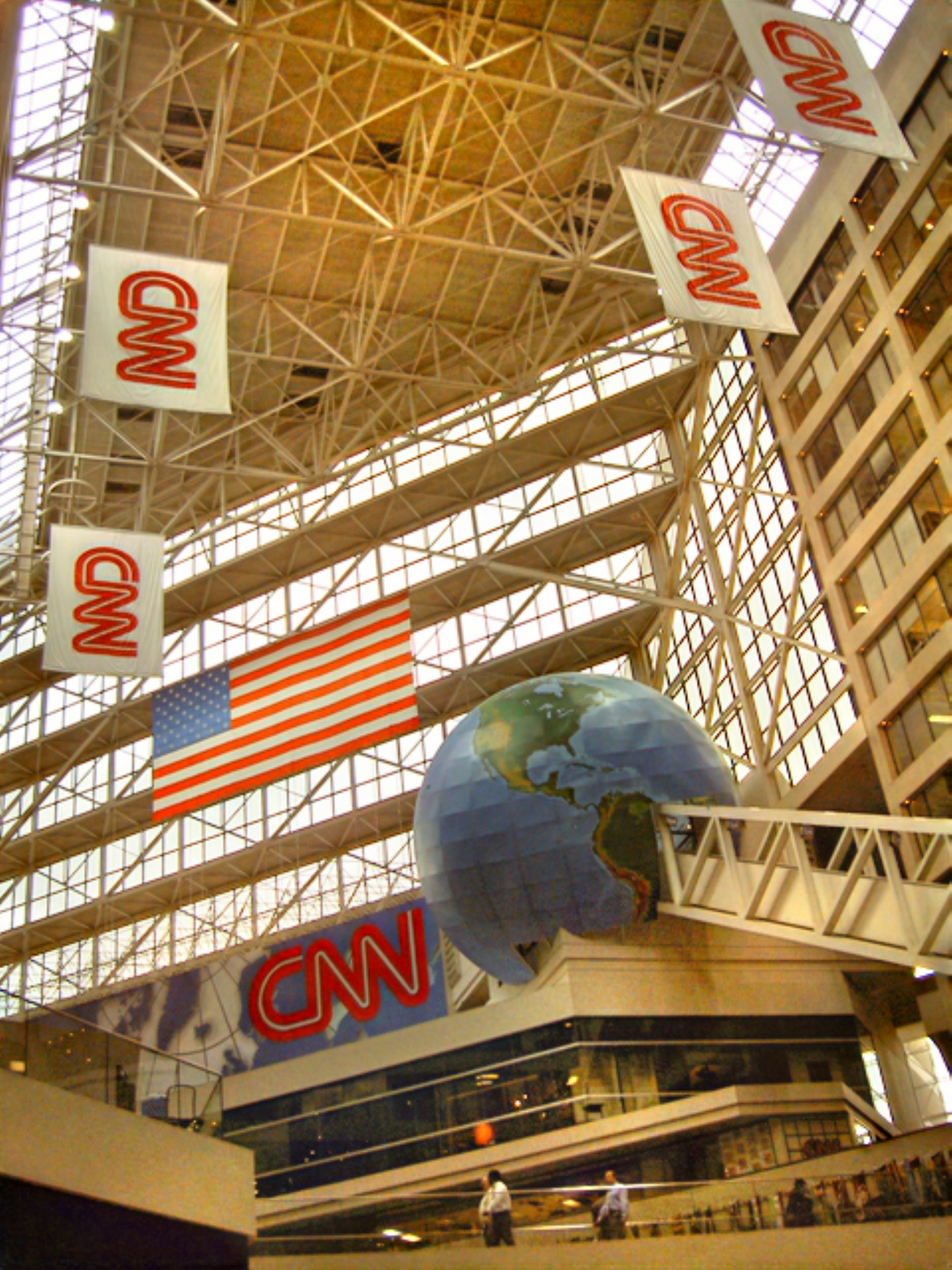}}
     \quad
     \subfloat[Zero-DCE \cite{guo2020zero}]{ \label{qua_no_gt:DICM_zerodce}
       \includegraphics[width=0.18\linewidth]{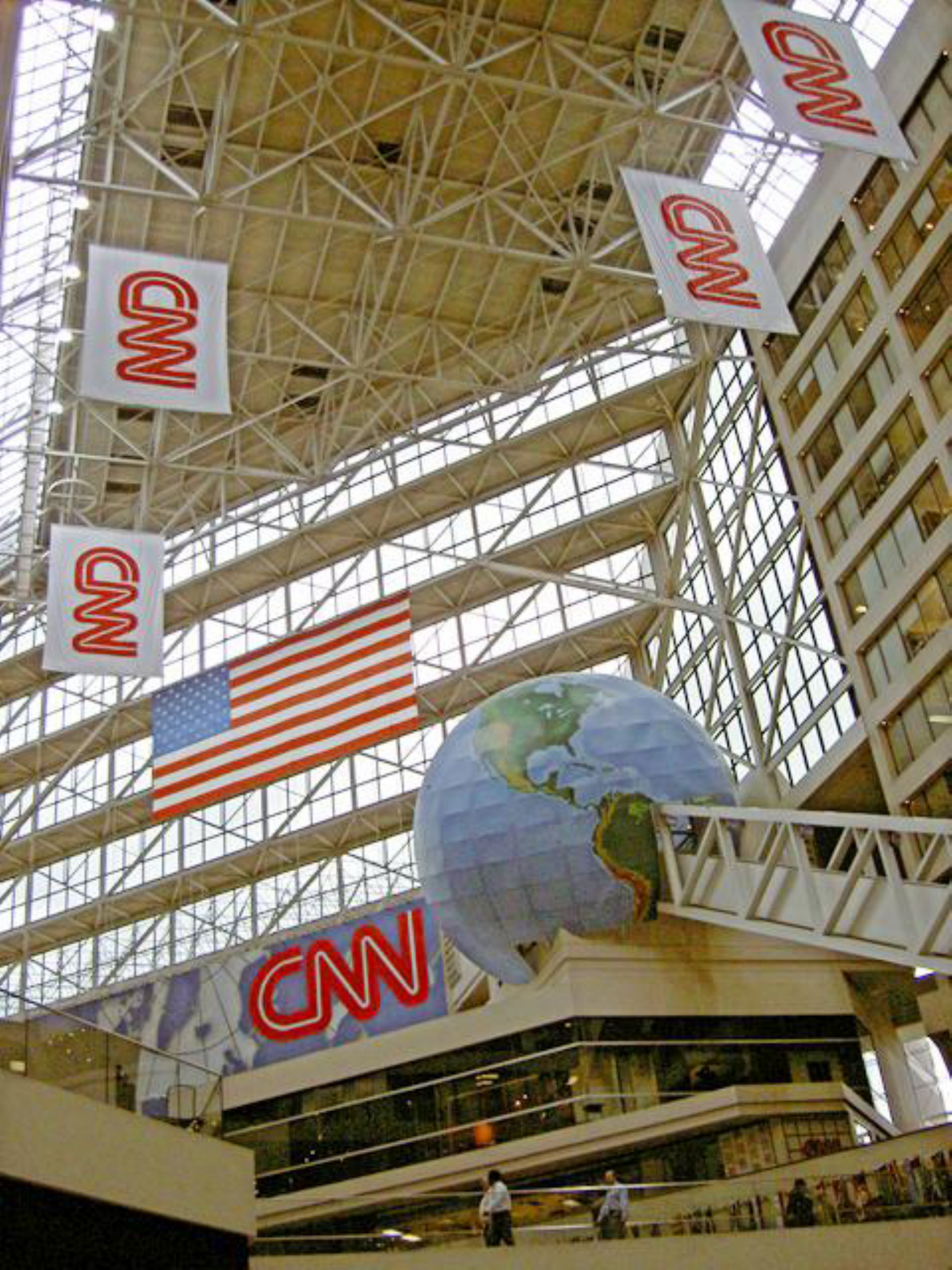}}
	  \quad
	  \subfloat[Ours]{\label{qua_no_gt:DICM_ours}
        \includegraphics[width=0.18\linewidth]{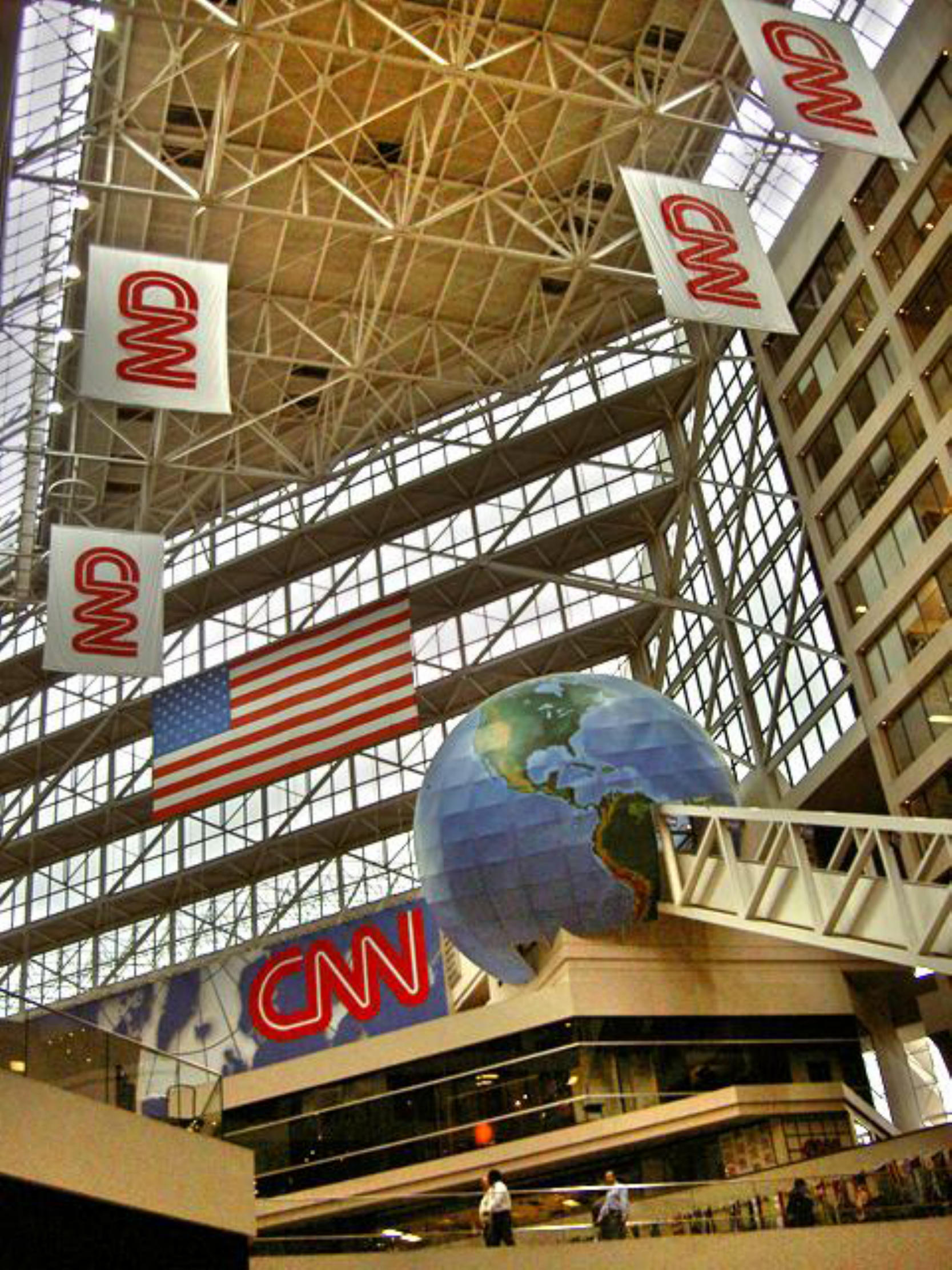}}\\
	  \subfloat[Input]{ \label{qua_no_gt:LIME_input}
       \includegraphics[width=0.18\linewidth]{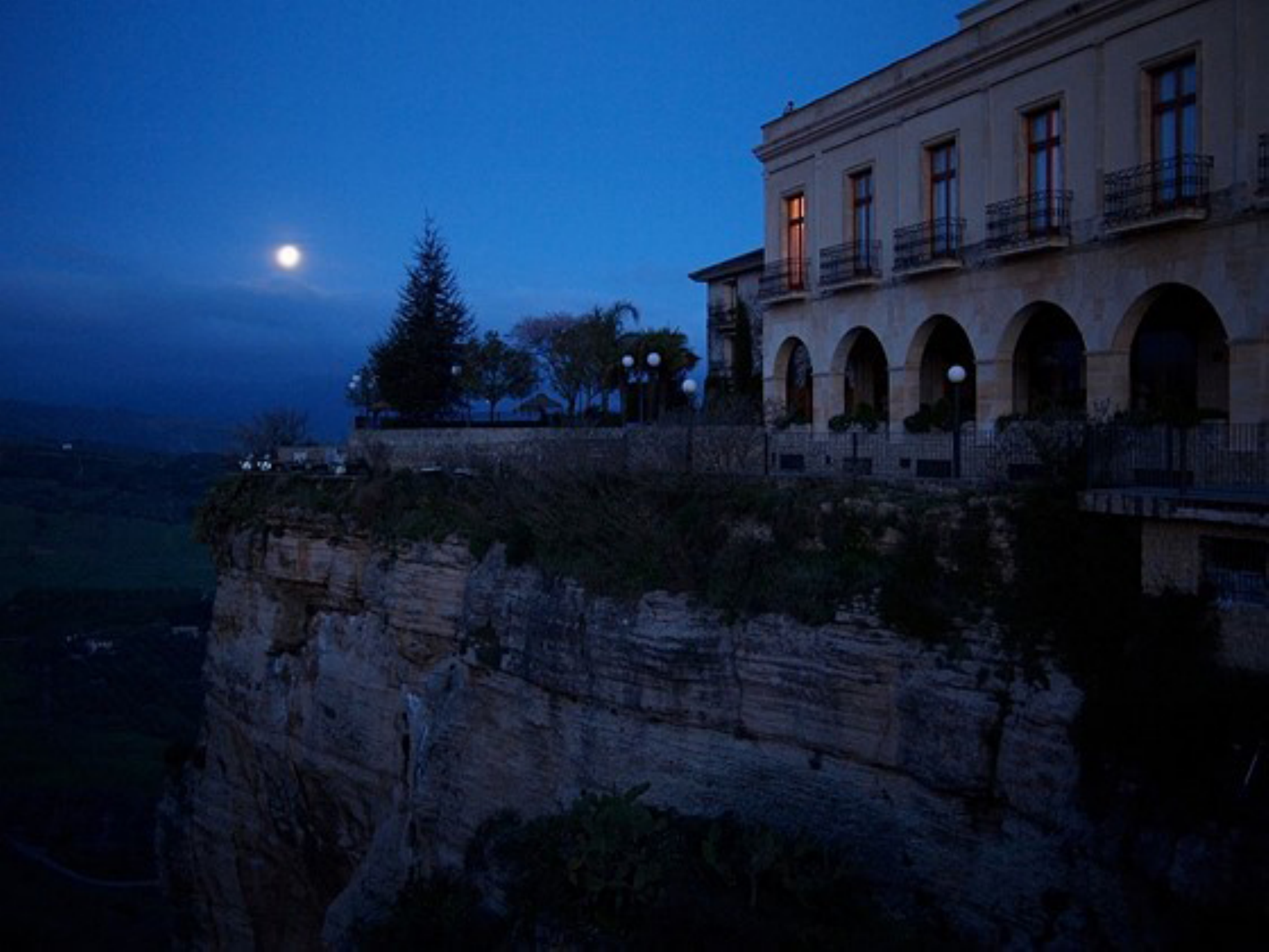}}
	  \quad
	  \subfloat[LIME \cite{guo2016lime}]{\label{qua_no_gt:LIME_lime}
        \includegraphics[width=0.18\linewidth]{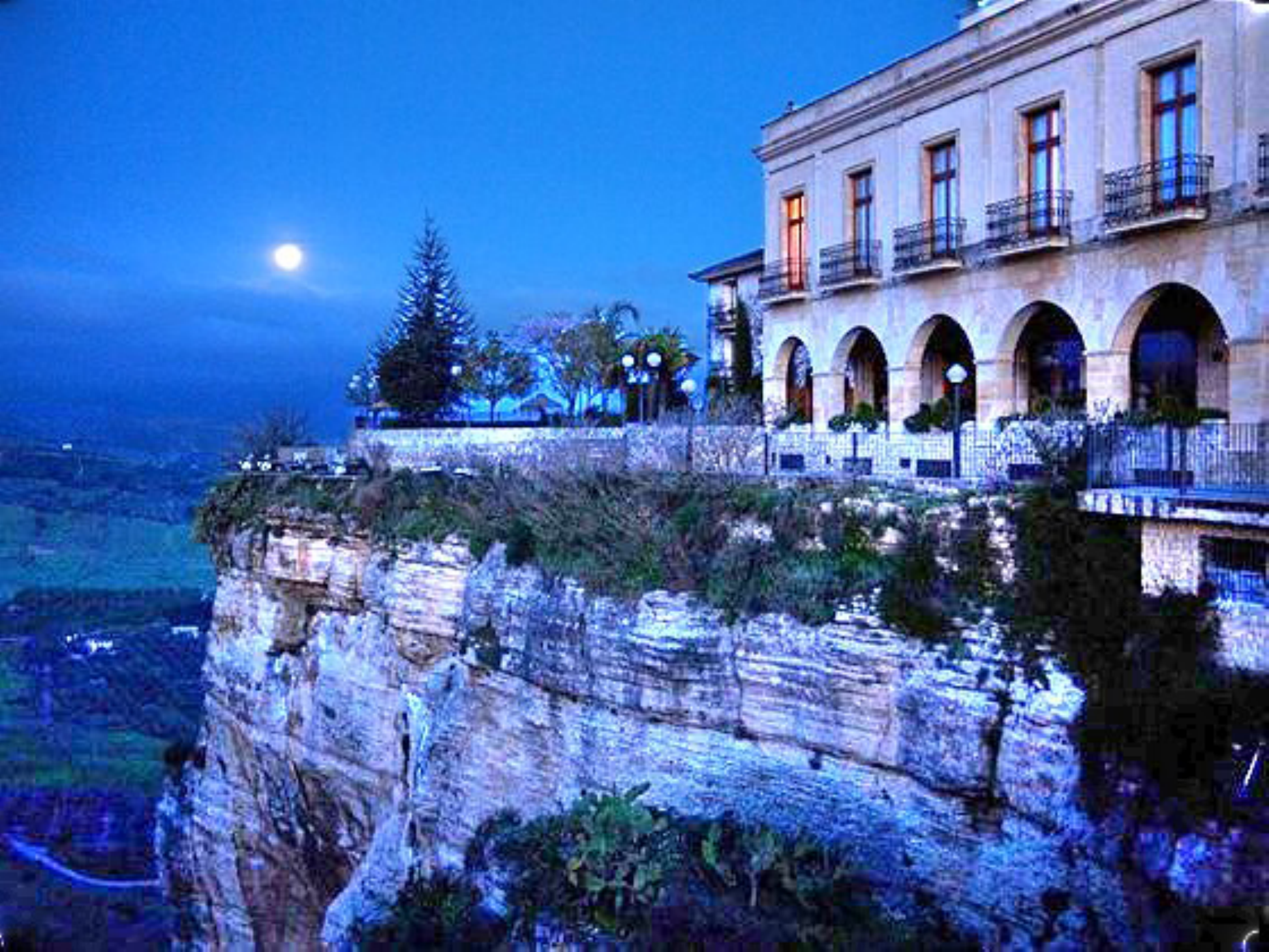}}
	  \quad
	  \subfloat[EnlightenGAN \cite{jiang2019enlightengan}]{\label{qua_no_gt:LIME_enlighten}
        \includegraphics[width=0.18\linewidth]{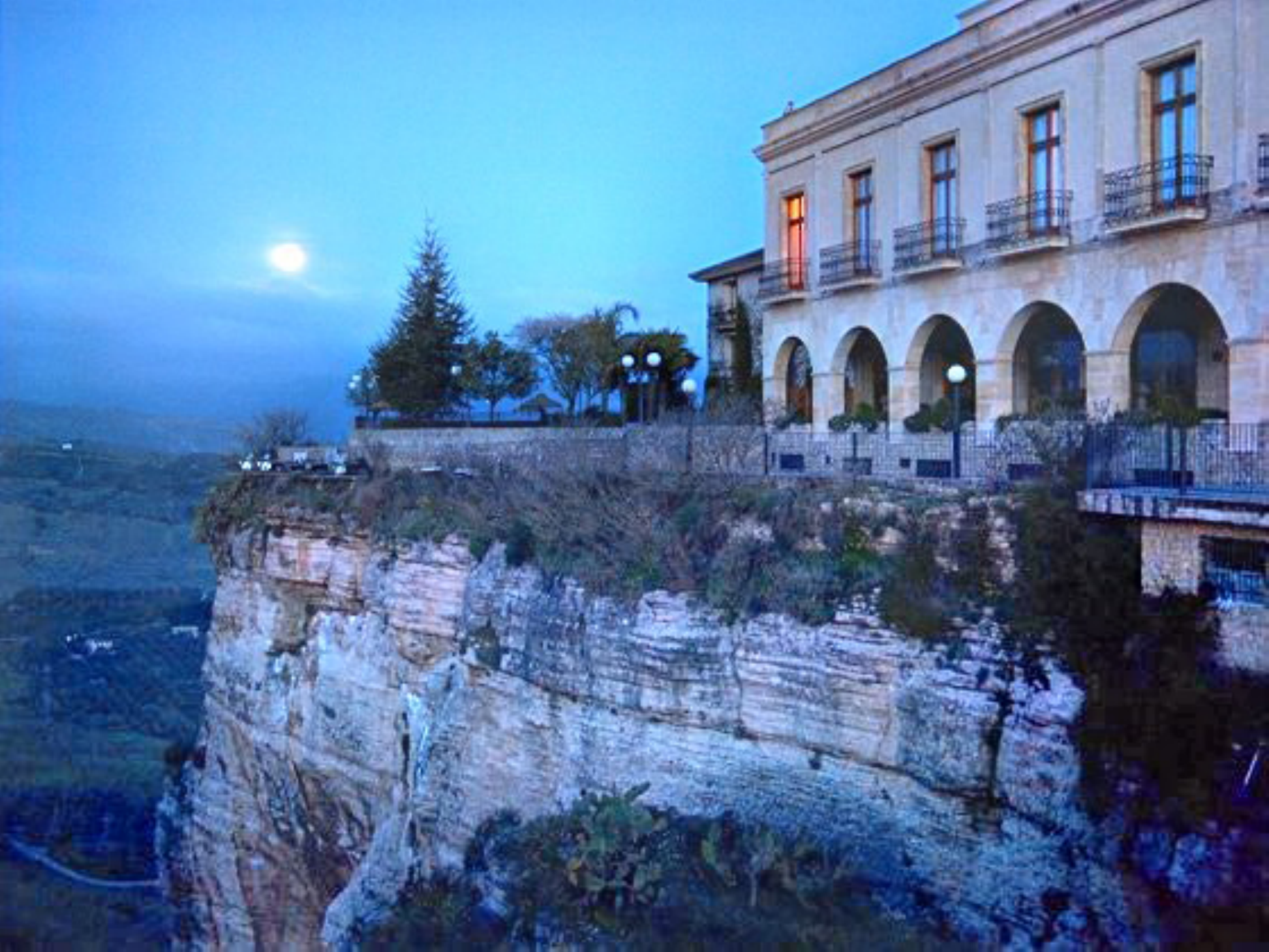}}
     \quad
     \subfloat[Zero-DCE \cite{guo2020zero}]{ \label{qua_no_gt:LIME_zerodce}
       \includegraphics[width=0.18\linewidth]{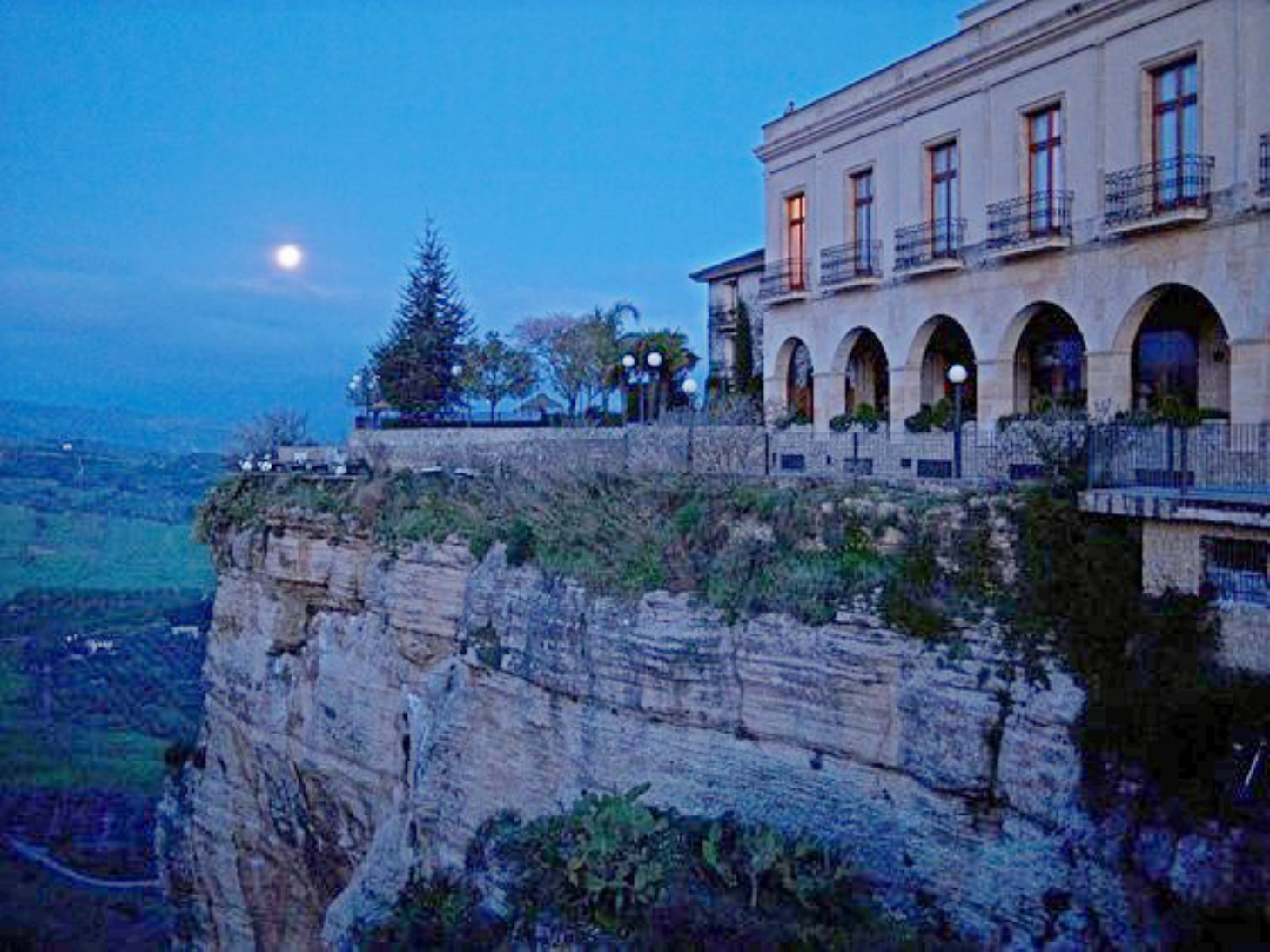}}
	  \quad
	  \subfloat[Ours]{\label{qua_no_gt:LIME_ours}
        \includegraphics[width=0.18\linewidth]{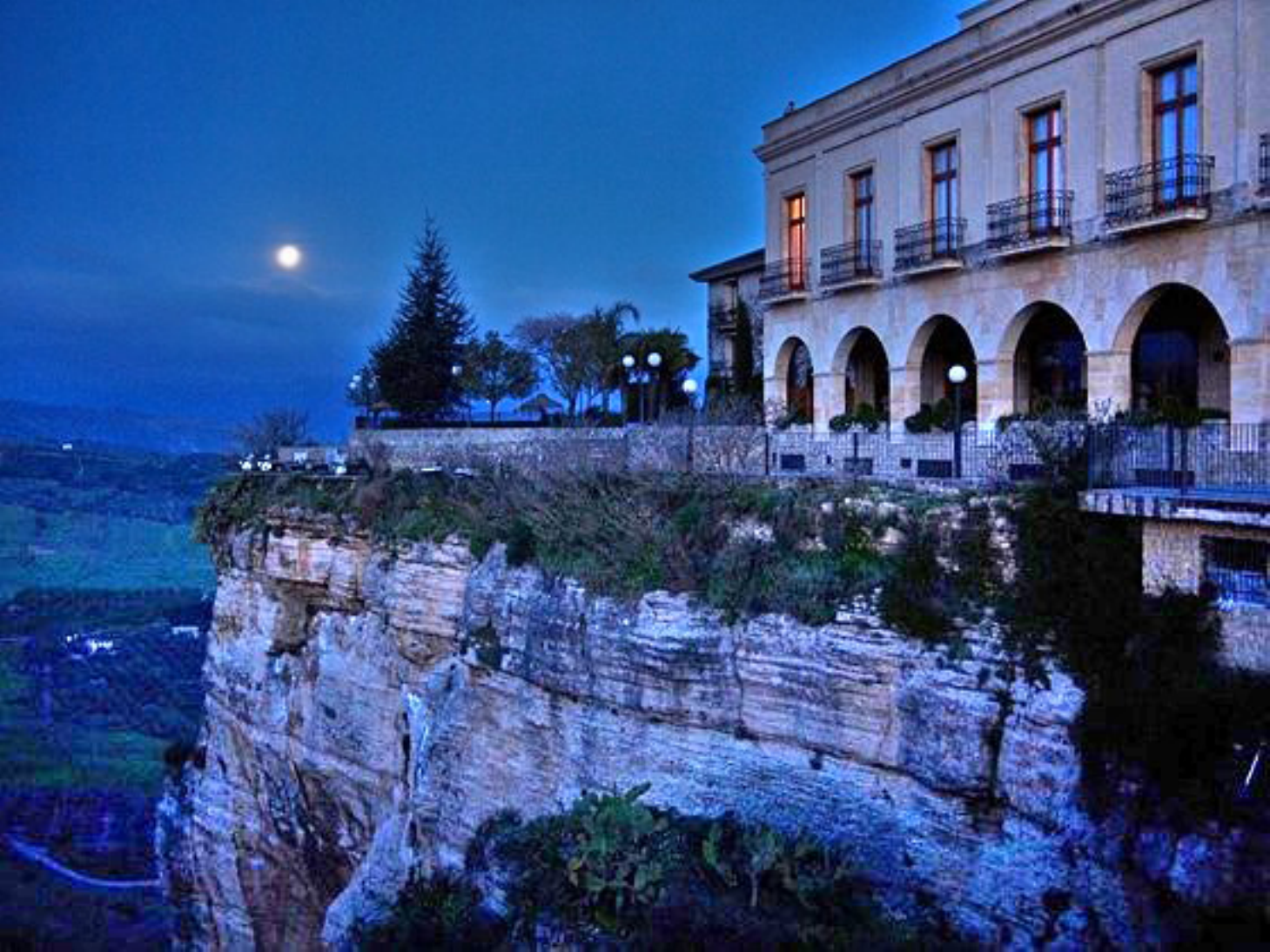}}\\
	  \subfloat[Input]{ \label{qua_no_gt:MEF_input}
       \includegraphics[width=0.18\linewidth]{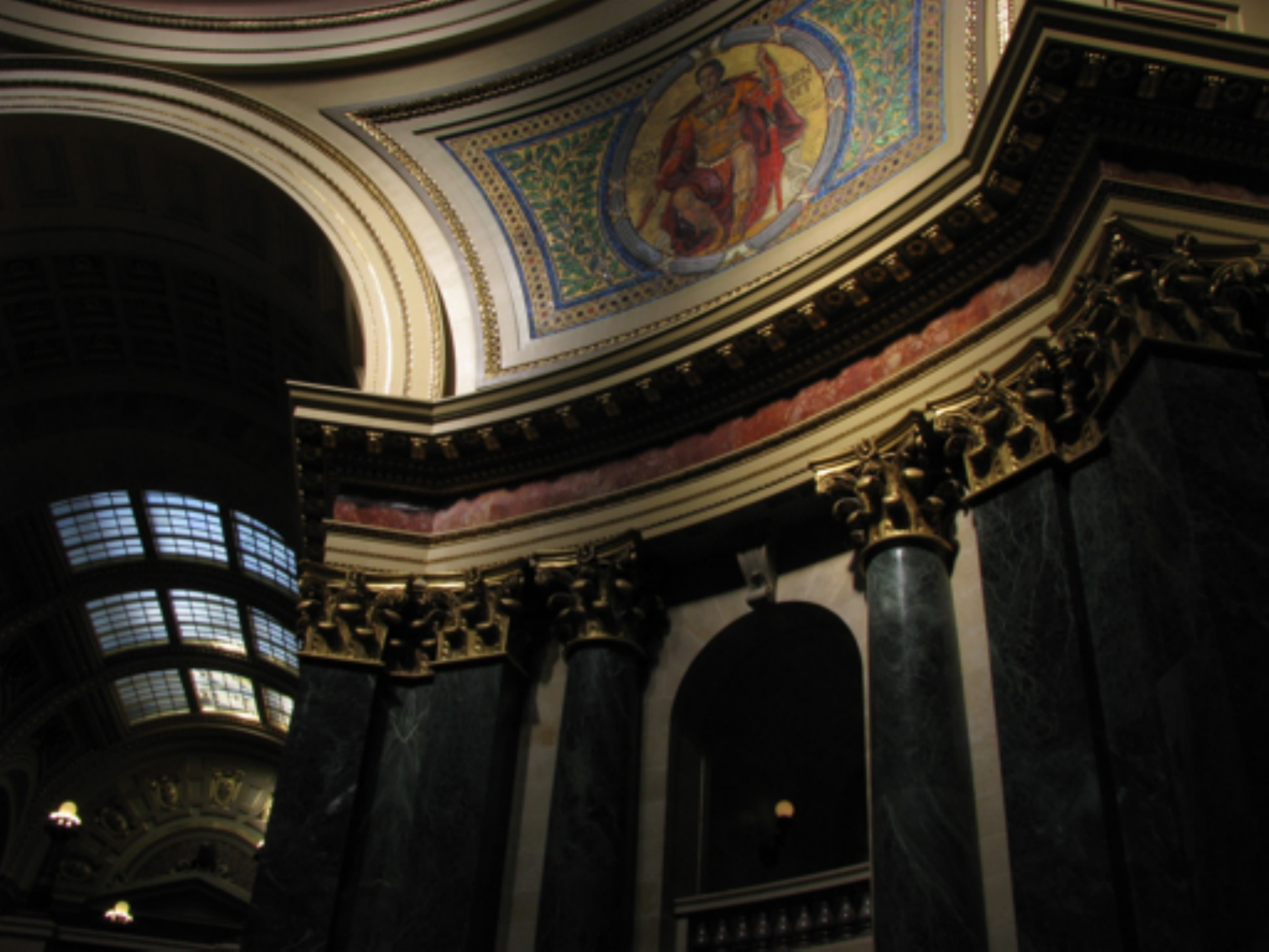}}
	  \quad
	  \subfloat[LIME \cite{guo2016lime}]{\label{qua_no_gt:MEF_lime}
        \includegraphics[width=0.18\linewidth]{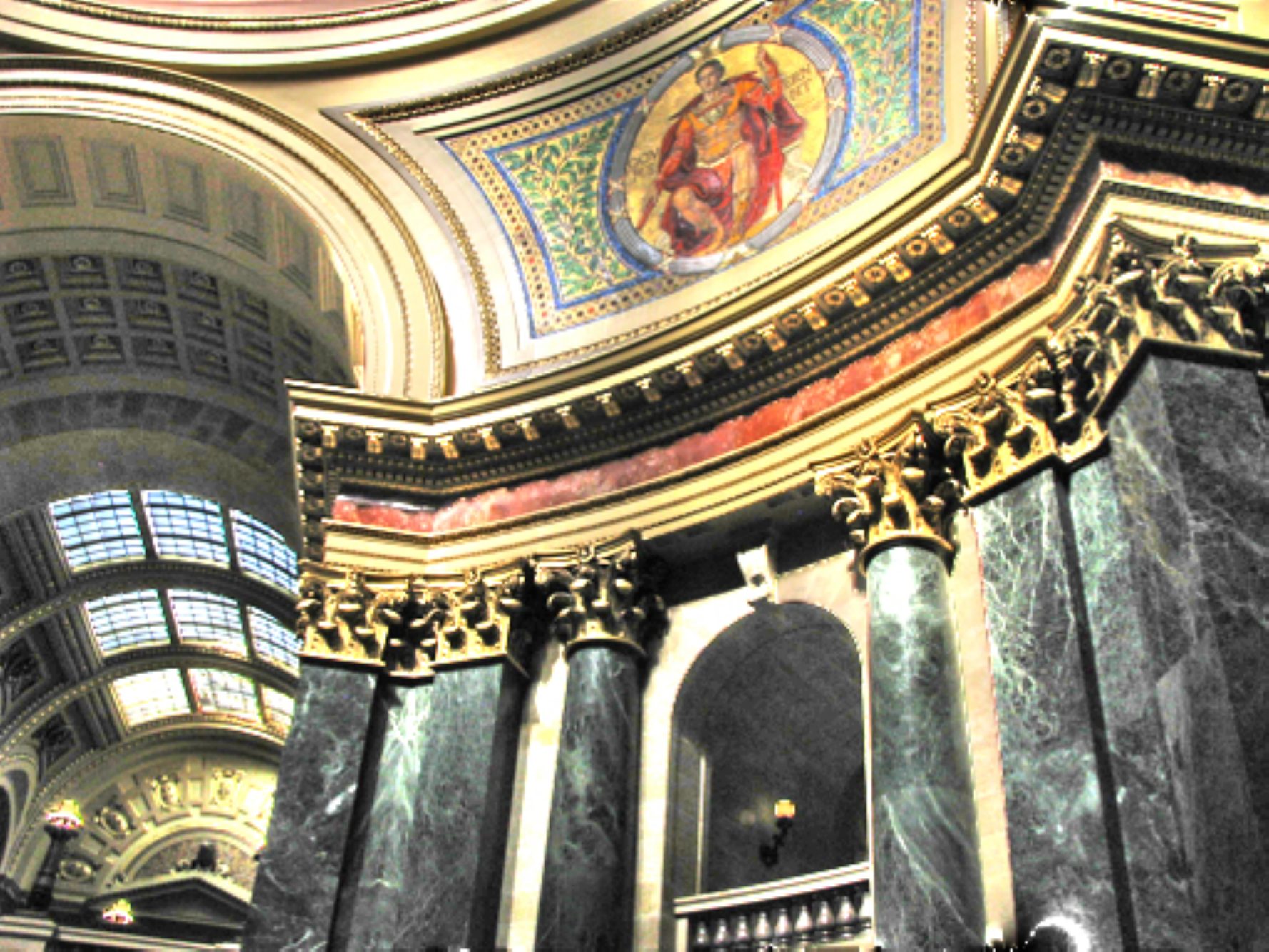}}
	  \quad
	  \subfloat[EnlightenGAN \cite{jiang2019enlightengan}]{\label{qua_no_gt:MEF_enlighten}
        \includegraphics[width=0.18\linewidth]{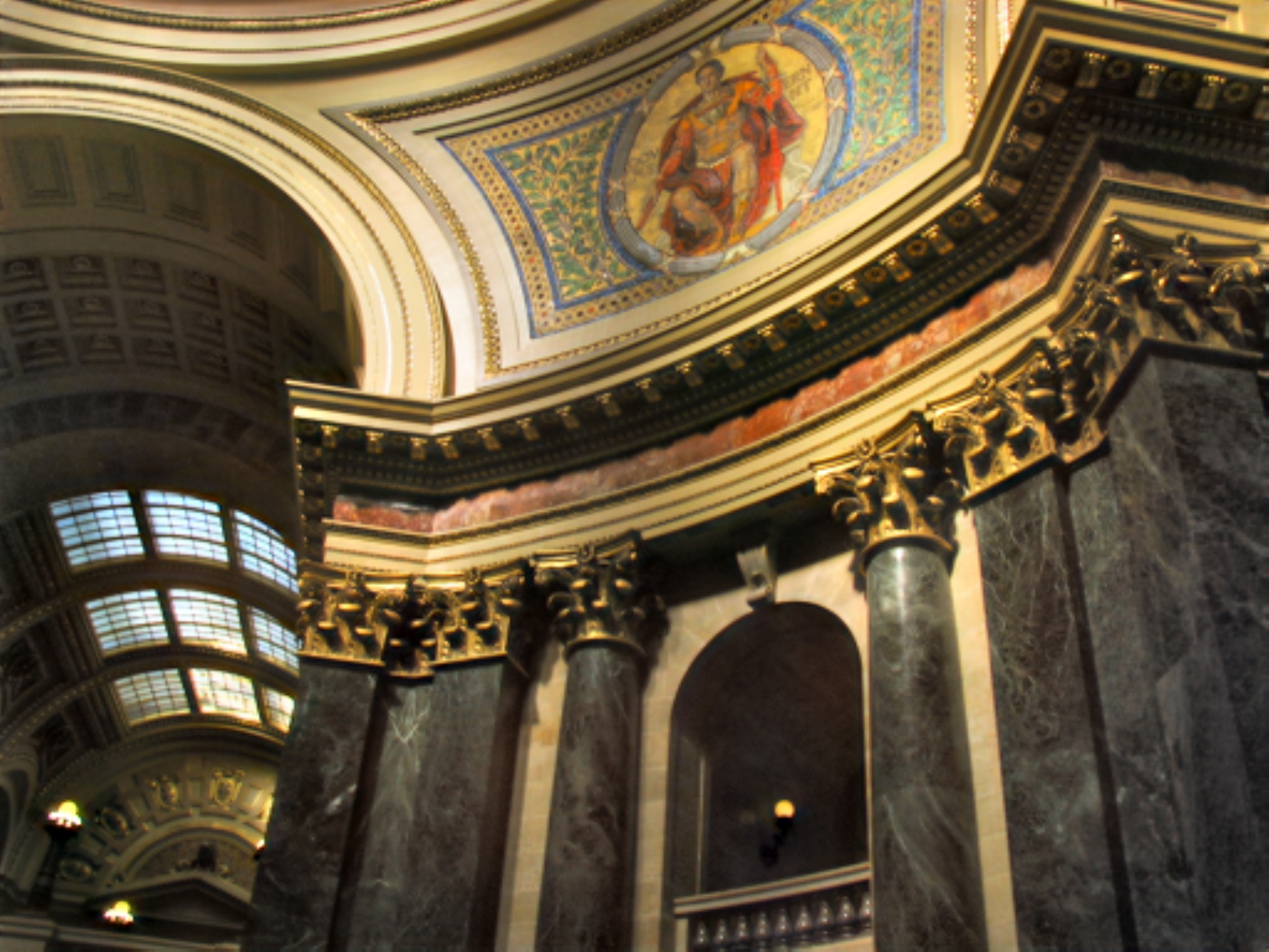}}
     \quad
     \subfloat[Zero-DCE \cite{guo2020zero}]{ \label{qua_no_gt:MEF_zerodce}
       \includegraphics[width=0.18\linewidth]{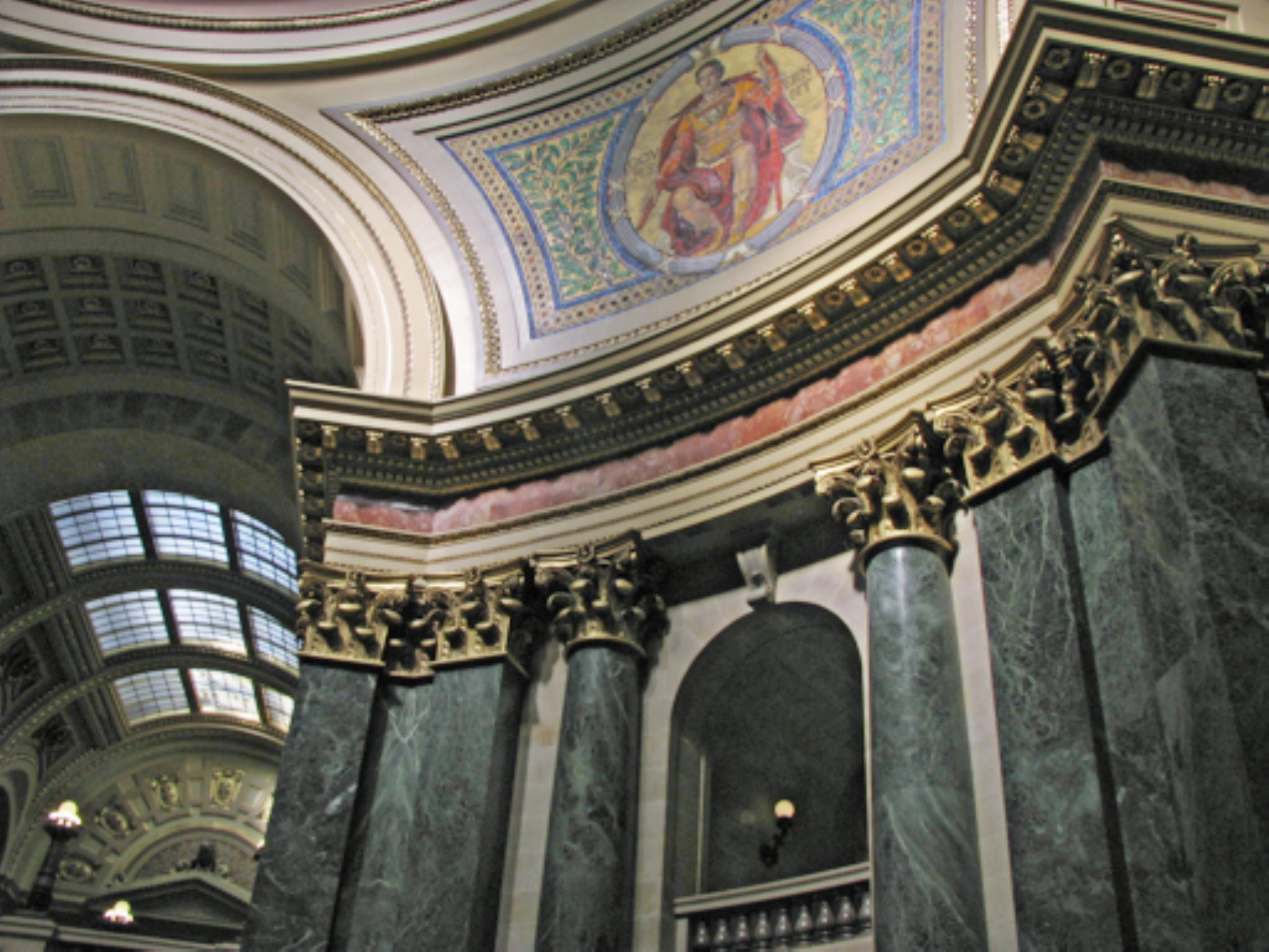}}
	  \quad
	  \subfloat[Ours]{\label{qua_no_gt:MEF_ours}
        \includegraphics[width=0.18\linewidth]{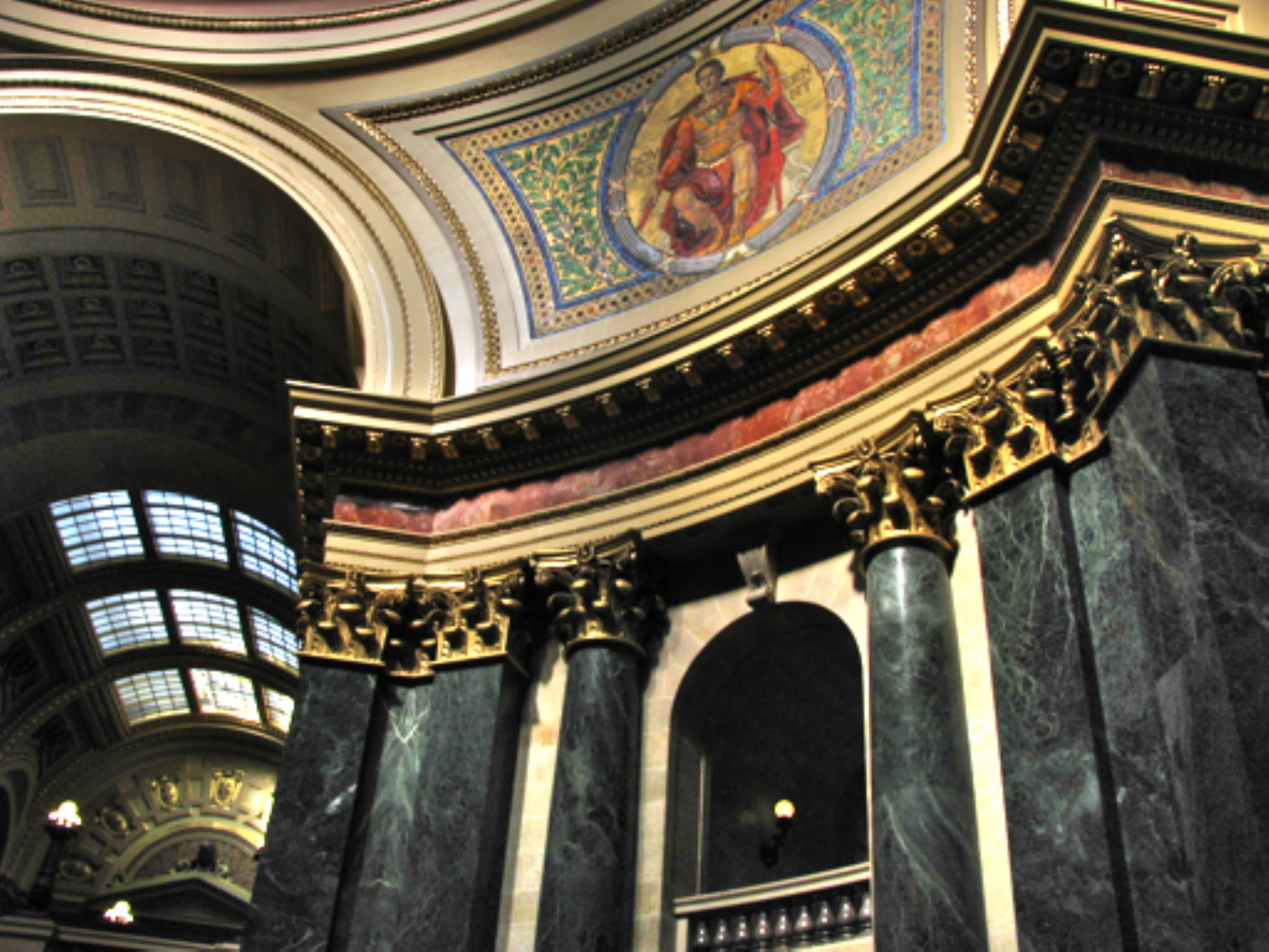}}\\
	  \subfloat[Input]{ \label{qua_no_gt:NPE_input}
       \includegraphics[width=0.18\linewidth]{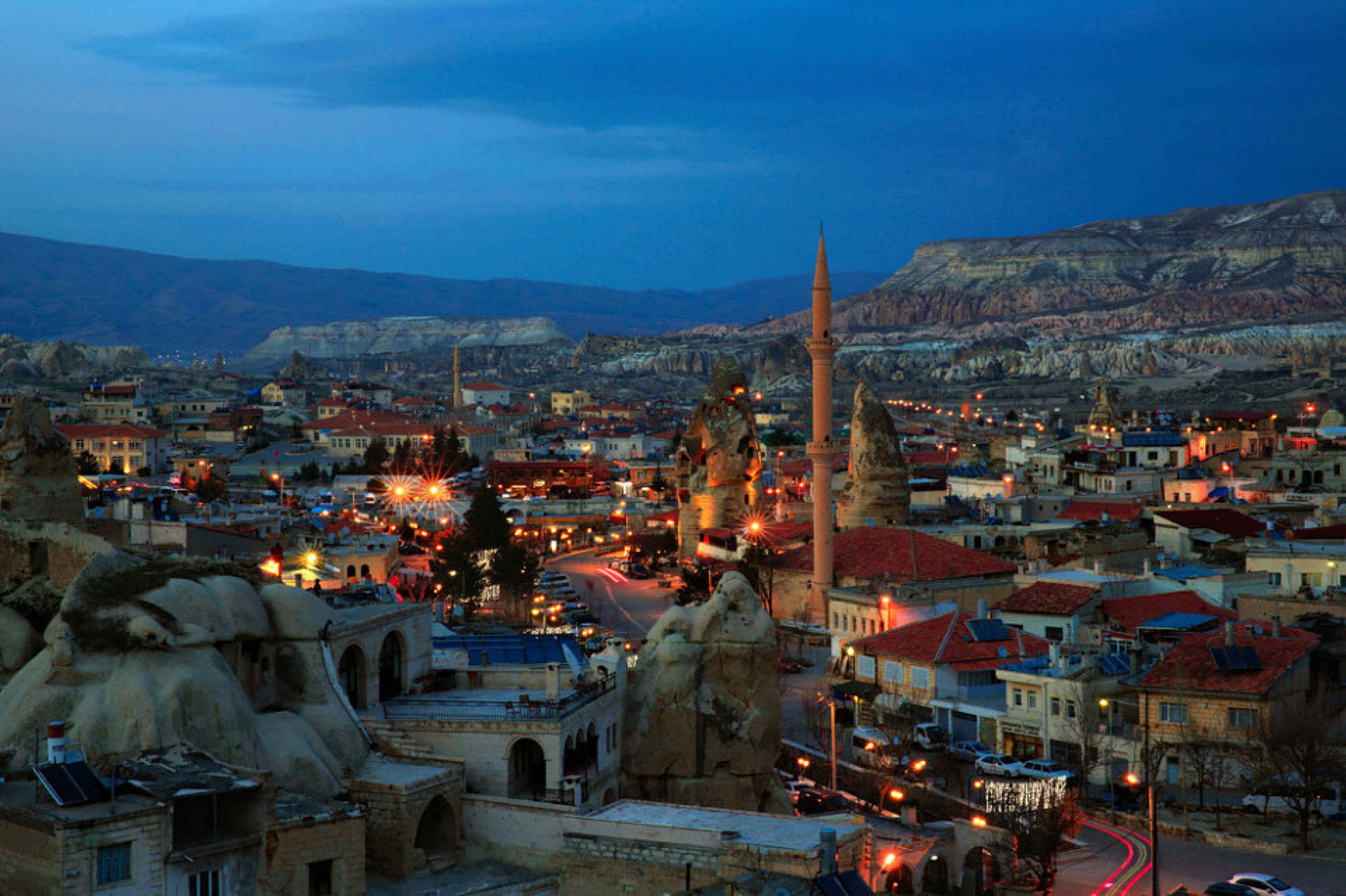}}
	  \quad
	  \subfloat[LIME \cite{guo2016lime}]{\label{qua_no_gt:NPE_lime}
        \includegraphics[width=0.18\linewidth]{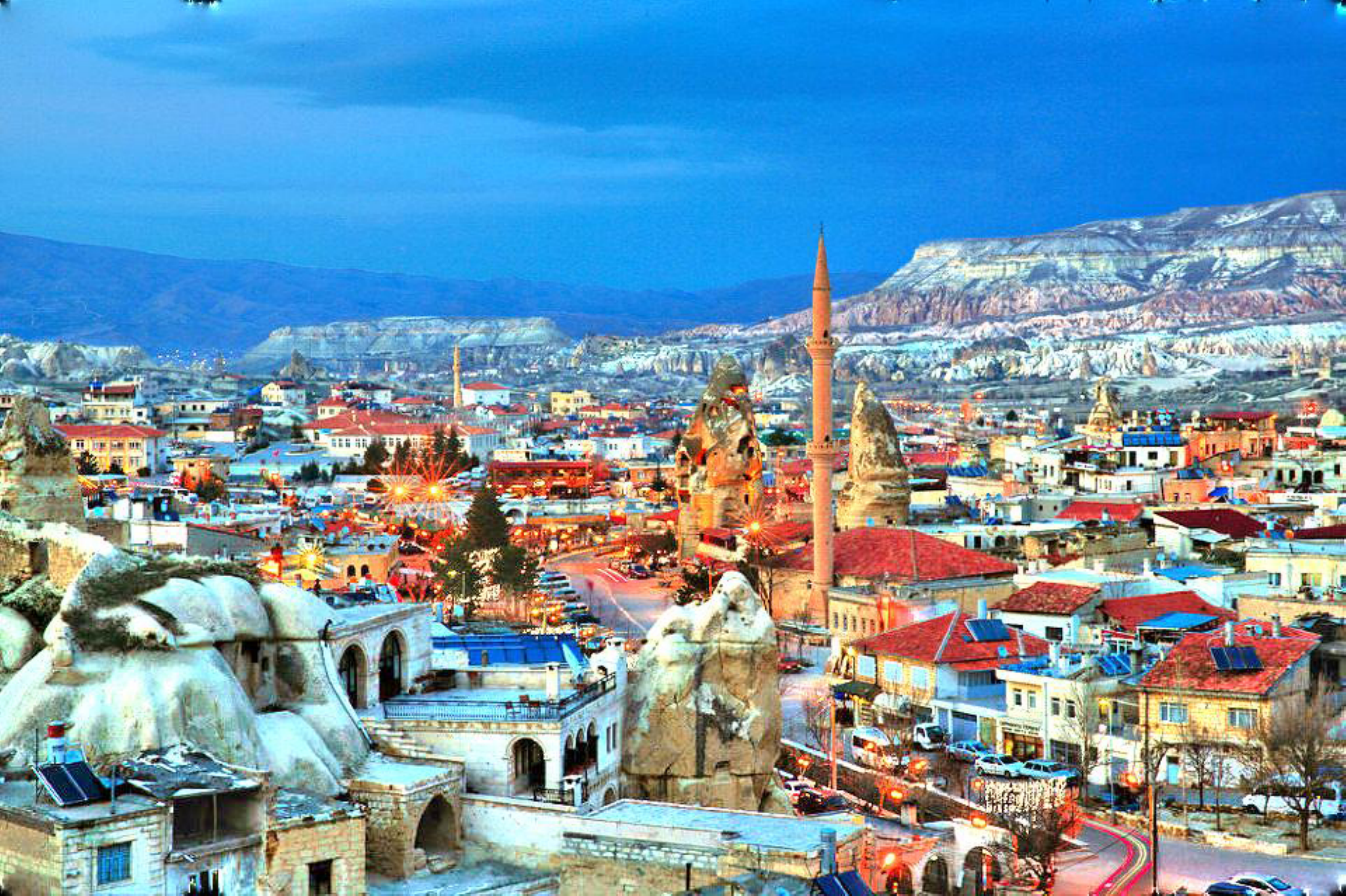}}
	  \quad
	  \subfloat[EnlightenGAN \cite{jiang2019enlightengan}]{\label{qua_no_gt:NPE_enlighten}
        \includegraphics[width=0.18\linewidth]{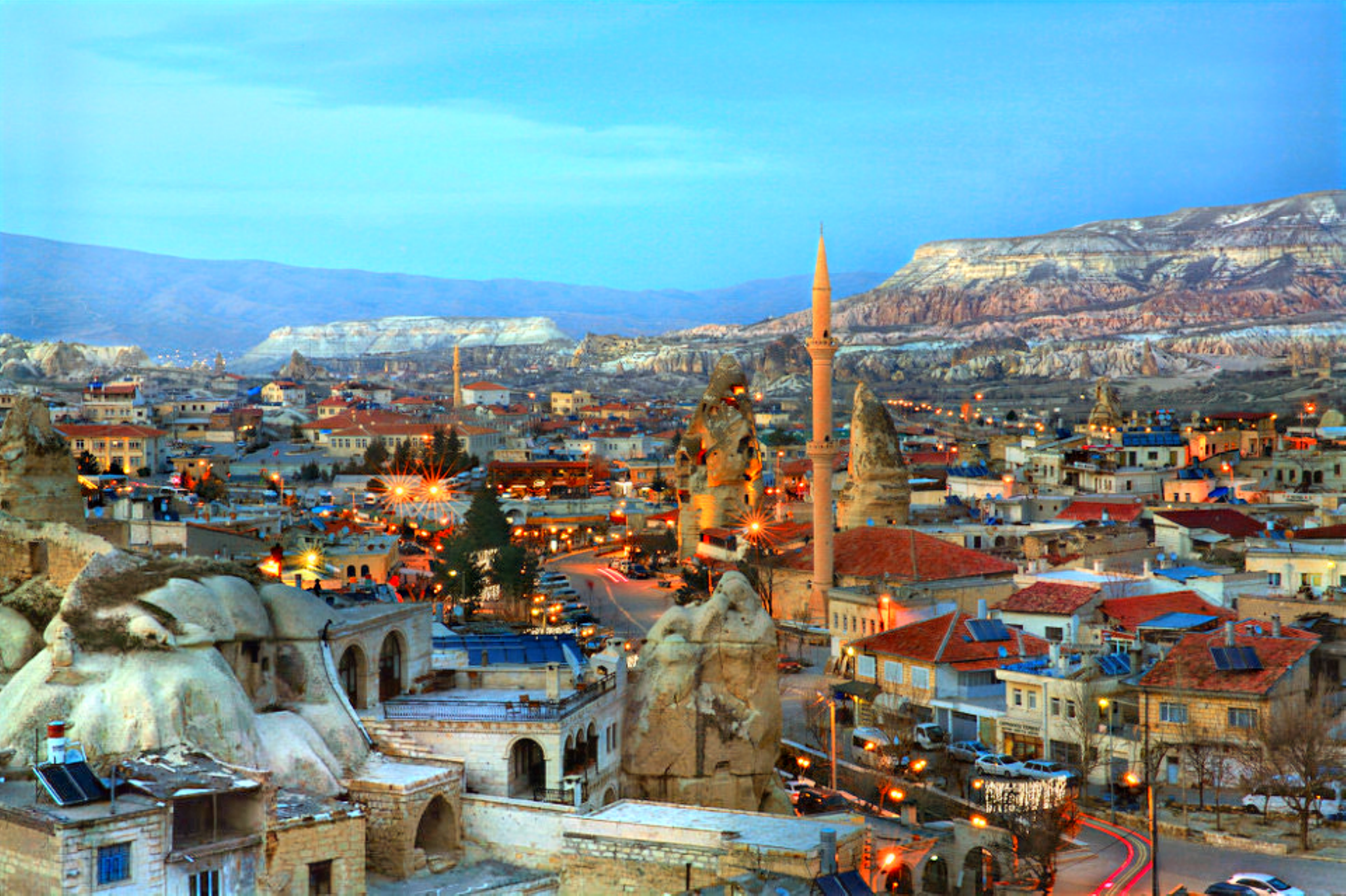}}
     \quad
     \subfloat[Zero-DCE \cite{guo2020zero}]{ \label{qua_no_gt:NPE_zerodce}
       \includegraphics[width=0.18\linewidth]{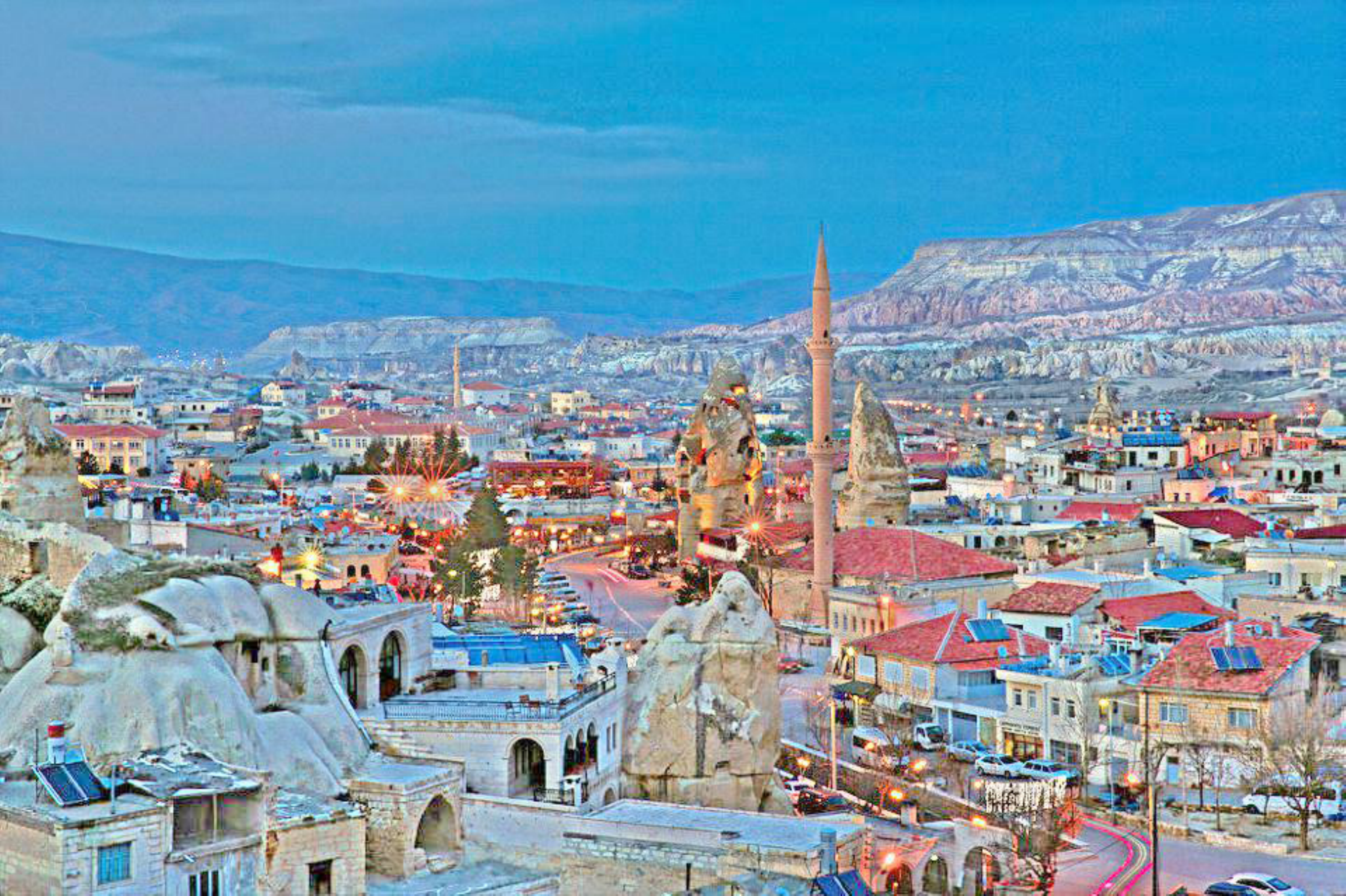}}
	  \quad
	  \subfloat[Ours]{\label{qua_no_gt:NPE_ours}
        \includegraphics[width=0.18\linewidth]{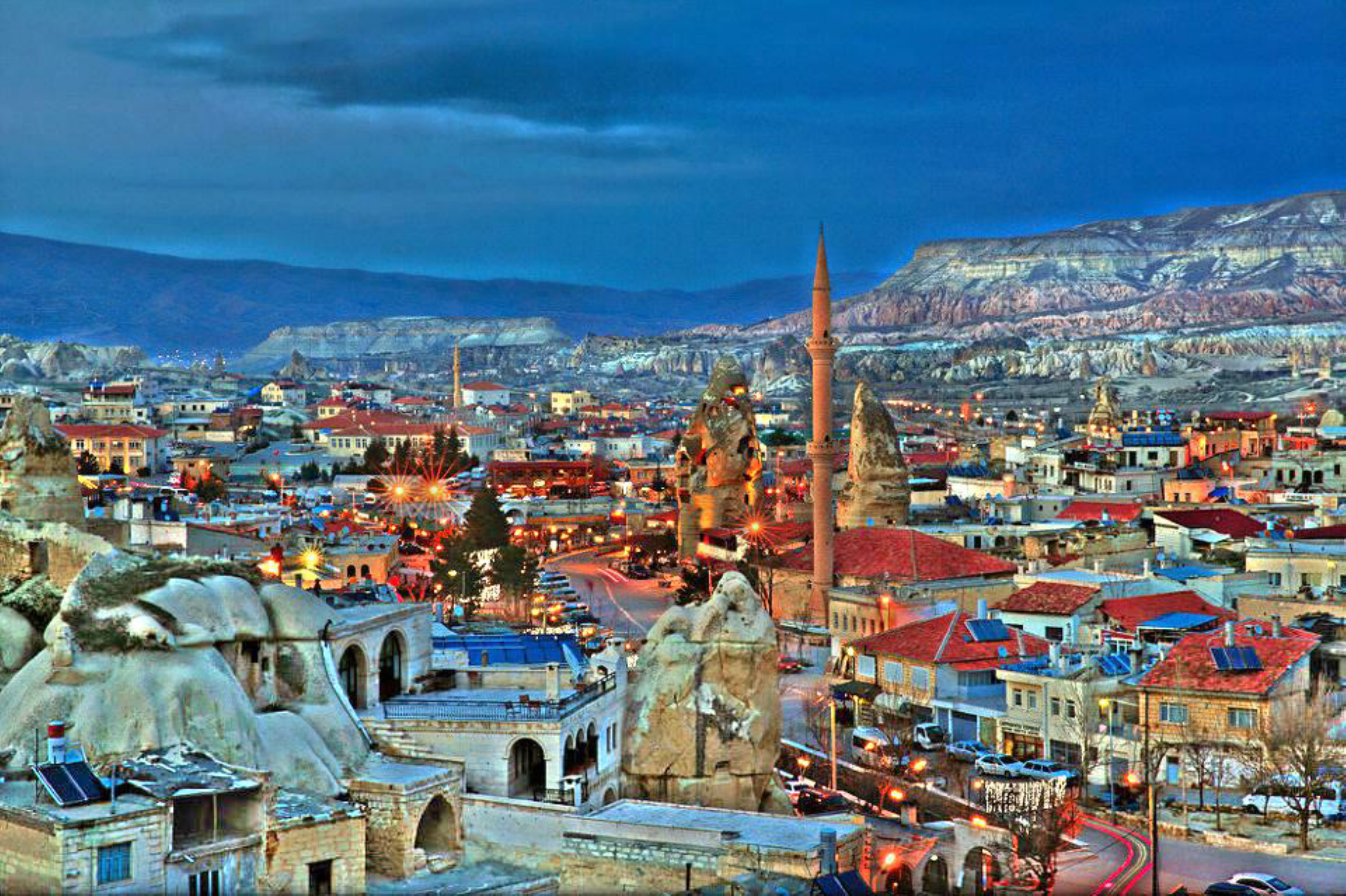}}
	  \caption{Visual comparisons with SOTA methods on DICM \cite{lee2012contrast}, LIME \cite{guo2016lime}, MEF \cite{ma2015perceptual} and NPE \cite{wang2013naturalness} datasets respectively.}
	  \label{fig:qua_no_gt} 
\end{figure*}

We also present visual results on DICM \cite{lee2012contrast}, LIME \cite{guo2016lime}, MEF \cite{ma2015perceptual} and NPE \cite{wang2013naturalness} datasets in Fig.\ref{fig:qua_no_gt}. Here we focus on the comparison with Zero-DCE, which is the SOTA method in self-regularized learning. In the case of similar brightness, our enhanced images have more vivid colors (\eg, the blue color of the tellurion in Fig.\ref{qua_no_gt:DICM_ours}, the golden decorations in Fig.\ref{qua_no_gt:MEF_ours}, the colorful town in Fig.\ref{qua_no_gt:NPE_ours}). This improvement is attributed to preserving the color by HSV color space in our method, instead of using Gray-World color constancy hypothesis \cite{buchsbaum1980spatial} as Zero-DCE \cite{guo2016lime}.

\subsection{Ablation Studies}
\label{Ablation}

\begin{table}[t]
	\caption{Quantitative result of ablation studies on the effectiveness of each loss function and the brightness disturbance. The optimal results are highlighted in red whereas the second-optimal results are highlighted in blue.}
	\centering
	\begin{tabular}{c|c c}
		\hline
		\textbf{Method} & \textbf{PSNR$\uparrow$ } & \textbf{SSIM$\uparrow$}   \\
		\hline
		baseline               &  \color{red}17.06    & \color{red}0.530  \\
		w/o $L_{rc}$           &  16.71               & 0.527             \\ 
		w/o $L_{ec}$           &  8.24                &   0.170           \\
		w/o $L_{ss}$           &  16.77               & 0.527             \\
		w/o $L_{is}$           &  12.82               &  0.410            \\
     w/o disturbance        &  16.73               & 0.527
\\
		two disturbed brightnesses   &  \color{blue}17.00   & \color{blue}0.528 \\
		\hline
	\end{tabular}
	\label{table:abla}
\end{table}

\begin{figure*} [t]
    \centering
	  \subfloat[baseline]{ \label{abla:baseline}
       \includegraphics[width=0.18\linewidth]{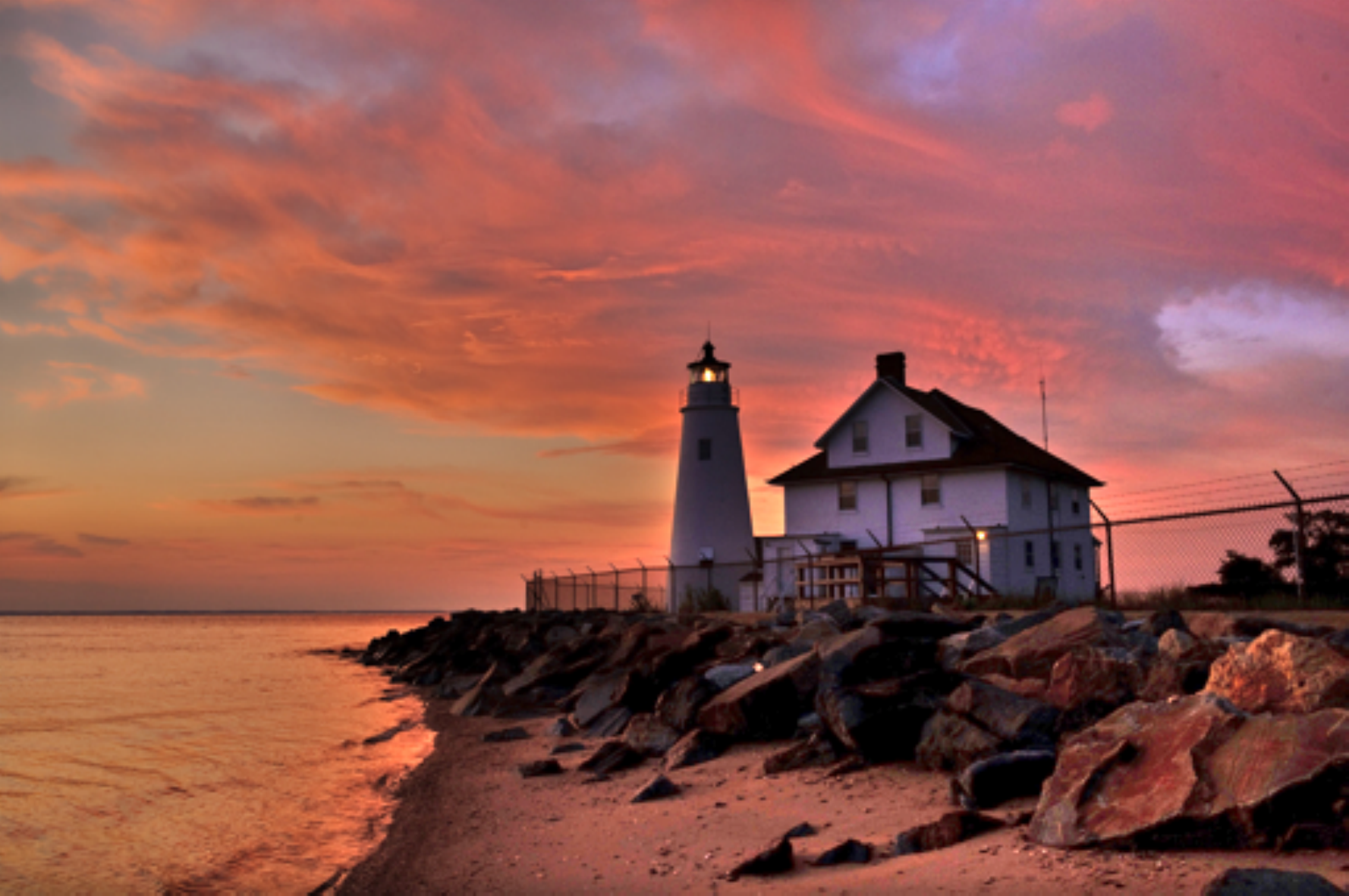}}
	  \quad
	  \subfloat[w/o $L_{rc}$]{\label{abla:norc}
        \includegraphics[width=0.18\linewidth]{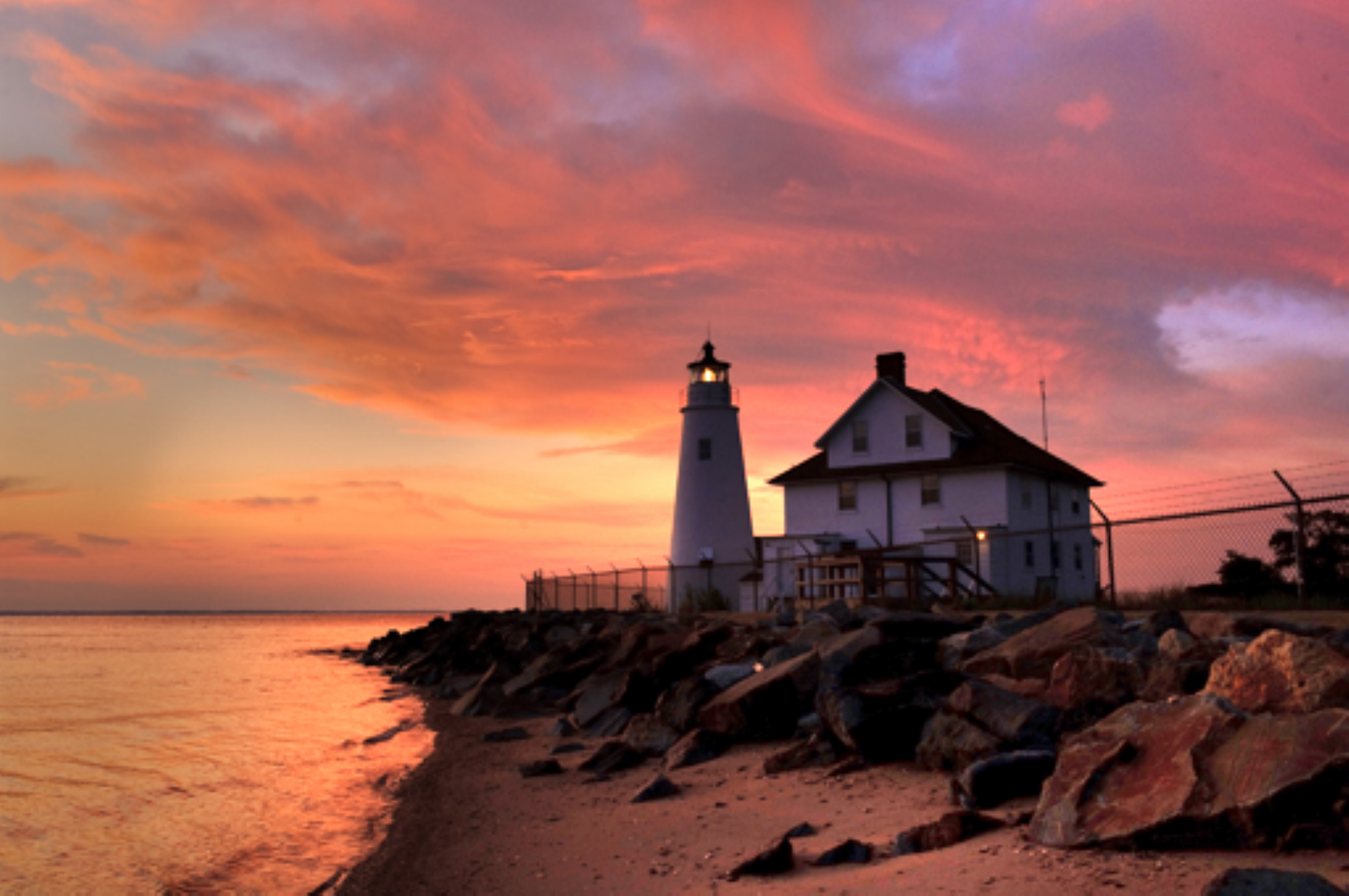}}
	  \quad
	  \subfloat[w/o $L_{ec}$]{\label{abla:noec}
        \includegraphics[width=0.18\linewidth]{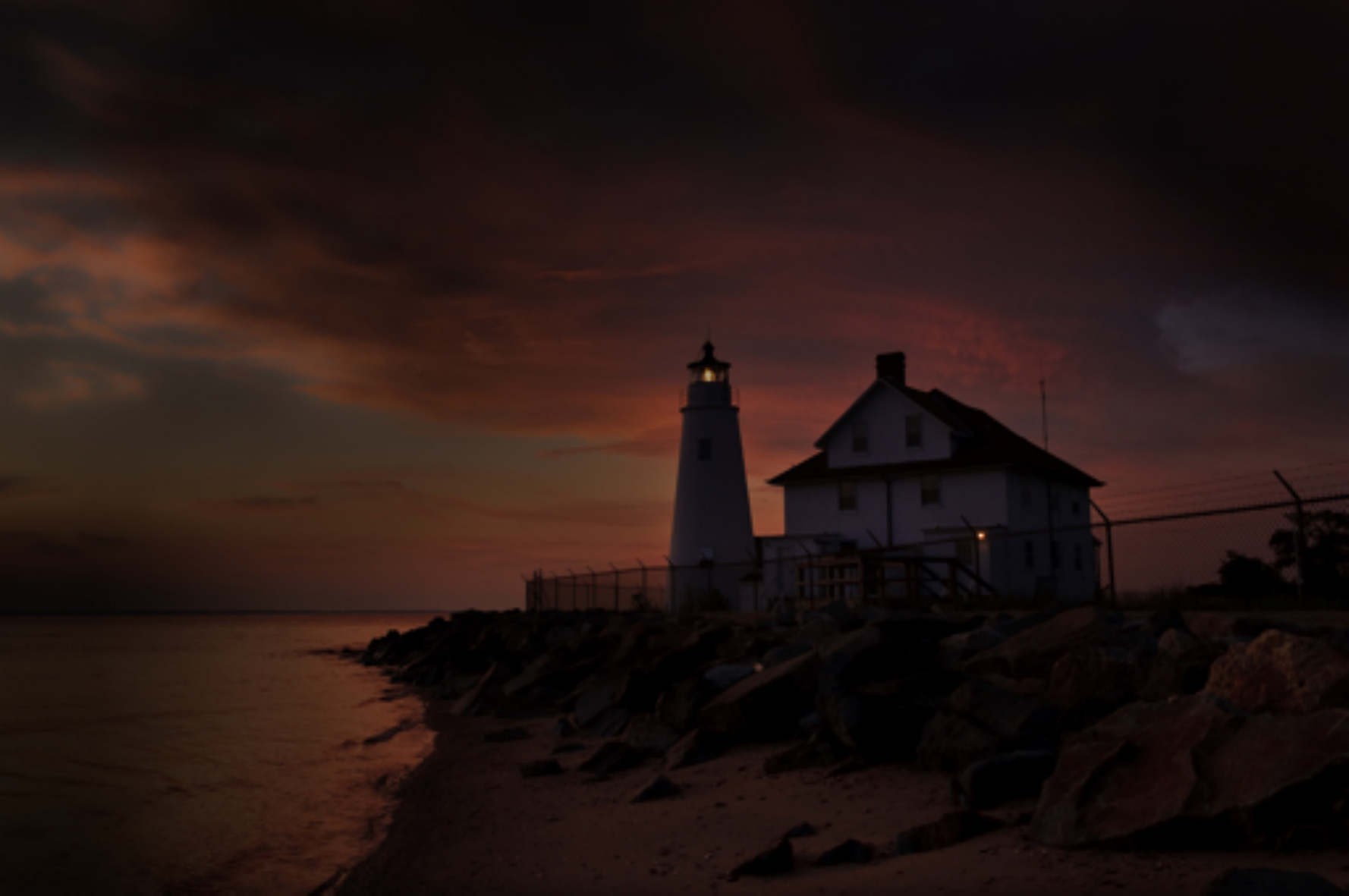}}
     \quad
	  \subfloat[w/o $L_{ss}$]{\label{abla:noss}
        \includegraphics[width=0.18\linewidth]{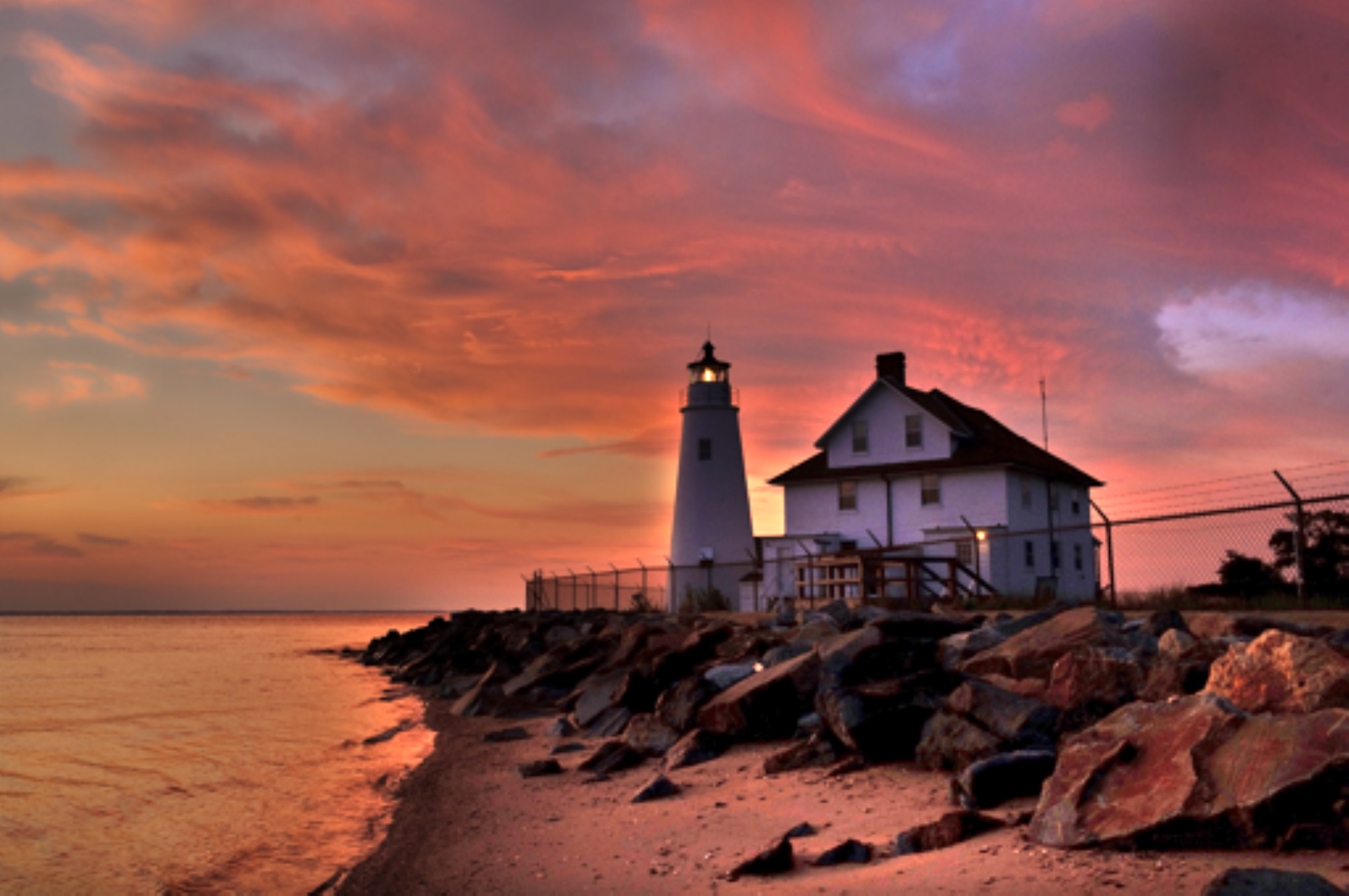}}
    \quad
	  \subfloat[w/o $L_{is}$]{\label{abla:nois}
        \includegraphics[width=0.18\linewidth]{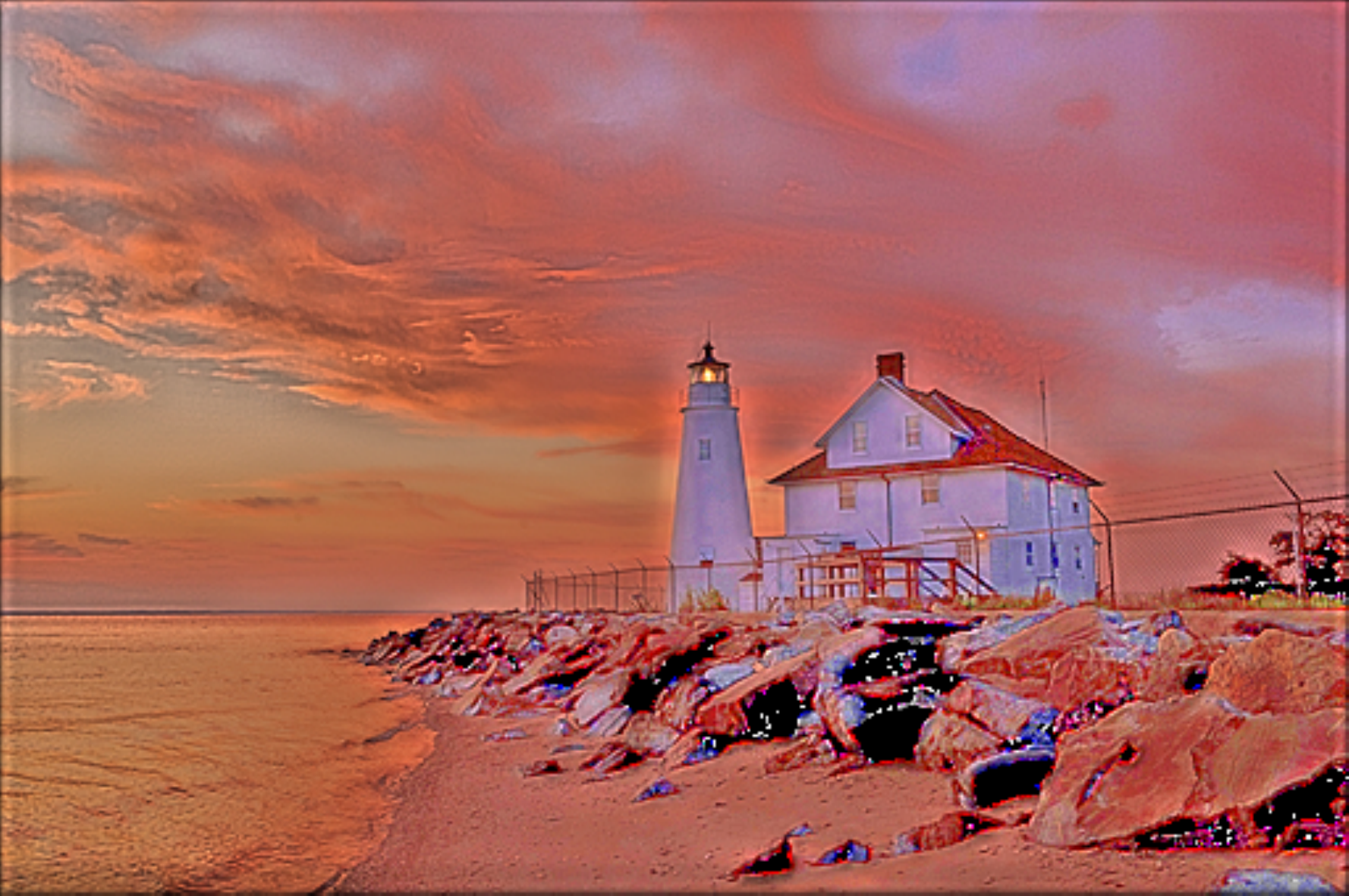}}\\
	  \caption{Visual comparison of ablation study on the effectiveness of each loss function.}
	  \label{fig:abla_loss} 
\end{figure*}

\begin{figure} [t]
    \centering
	  \subfloat[w/o disturbance]{ \label{abla:nodis}
       \includegraphics[width=0.3\linewidth]{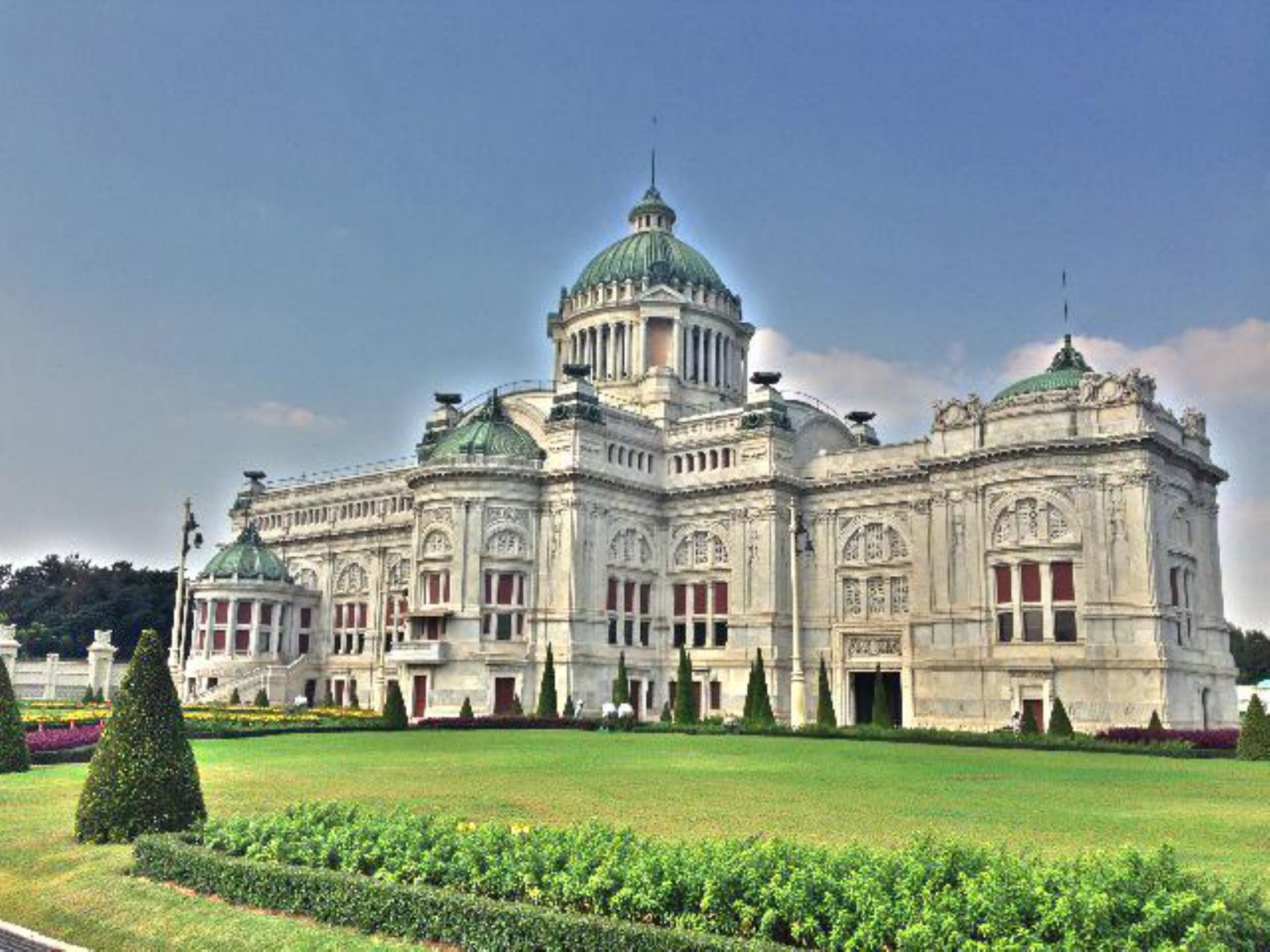}}
	  \quad
	  \subfloat[one disturbed brightness (baseline)]{\label{abla:1dis}
        \includegraphics[width=0.3\linewidth]{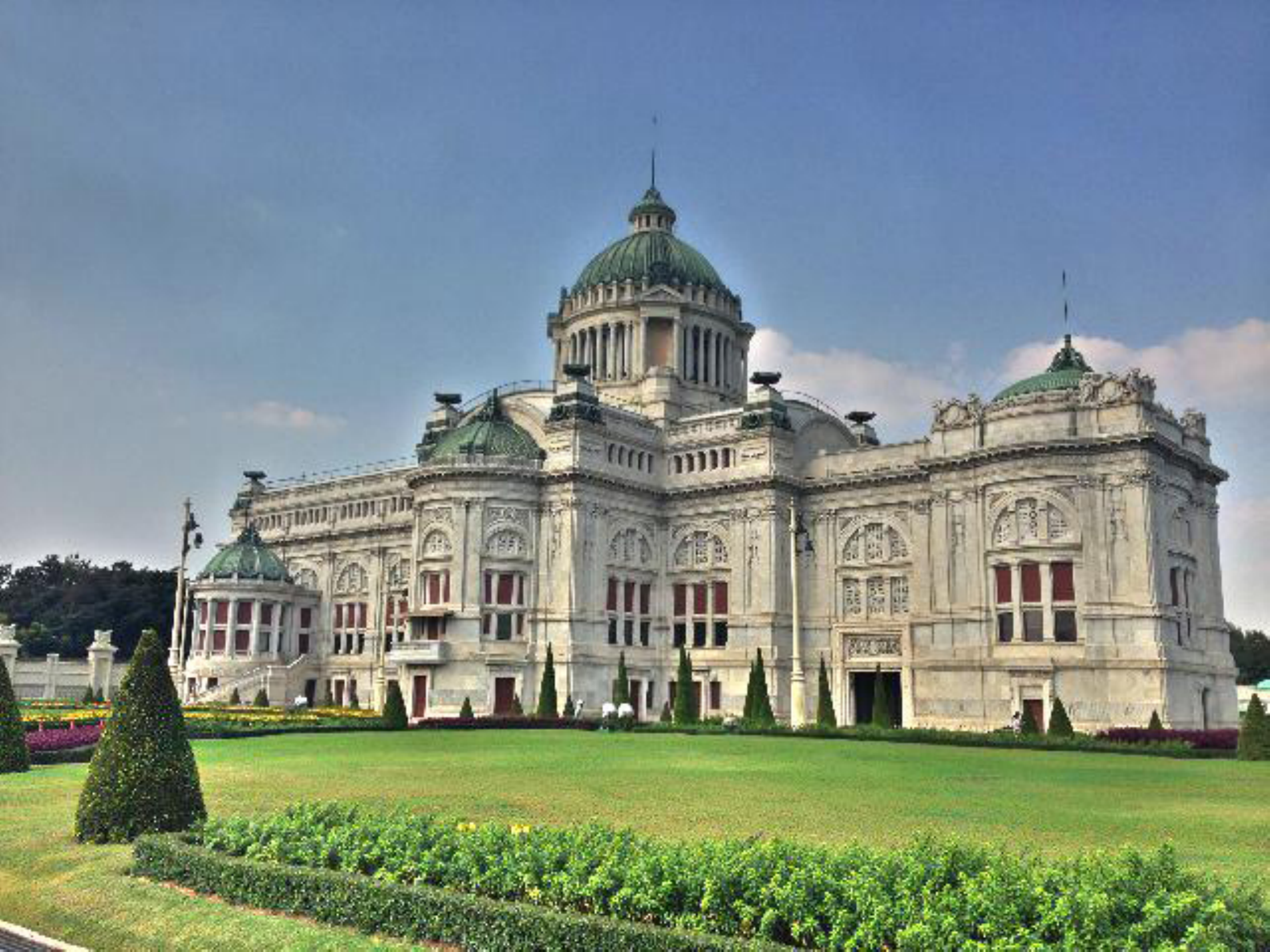}}
	  \quad
	  \subfloat[two disturbed brightnesses]{\label{abla:2dis}
        \includegraphics[width=0.3\linewidth]{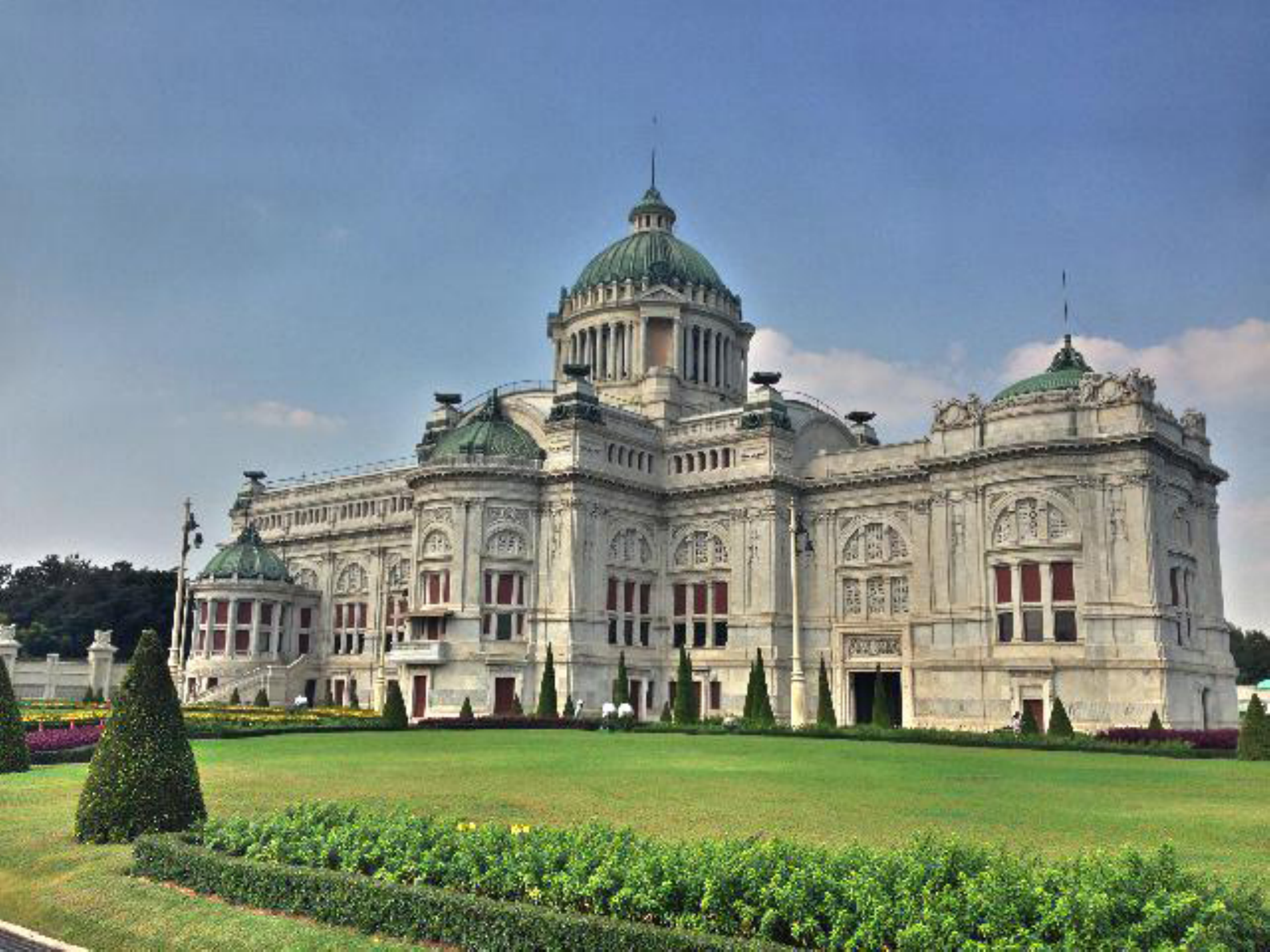}}
	  \caption{Visual comparison of ablation study on the effectiveness of brightness  disturbance approach.}
	  \label{fig:abla_disturbance} 
\end{figure}

We conduct ablation studies to demonstrate the effectiveness of the loss functions and the brightness disturbance in our method. We treat the model of full loss functions and one brightness disturbance as baseline. Quantitative and qualitative results are reported in Table.\ref{table:abla}, Fig.\ref{fig:abla_loss} and Fig.\ref{fig:abla_disturbance}, where the quantitative experiments uses the same test dataset in Sec.\ref{comp}.

\subsubsection{Effectiveness of Each Loss Function}
The first five rows of Table.\ref{table:abla} show that discarding any loss function will incur lower performance quantitatively. In qualitative terms, the result without reflectance consistency loss $L_{rc}$ has abnormal-exposed regions (\eg, the left region of the lighthouse in Fig.\ref{abla:norc}). It is worth noting that despite the added disturbance, due to the lack of reflectance consistency loss, the network essentially learns an image-to-image mapping instead of an image-to-reflectance mapping. Besides, removing exposure control loss $L_{ec}$ fails to brighten up low-light images. Furthermore, the absence of spatial structure loss $L_{ss}$ generates weird dark regions (\eg, the upper right corner of Fig.\ref{abla:noss}). Last but not least, severe color distortion emerges without illumination smoothness loss $L_{is}$.

\subsubsection{Effectiveness of Brightness Disturbance}

We study the effectiveness of the brightness disturbance by adjusting the number of disturbed brightness, including no disturbed brightness, one disturbed brightness and two disturbed brightnesses. The quantitative and qualitative results are showed in the last two row of Table \ref{table:abla} and Fig.\ref{fig:abla_disturbance}. 

Compared to our baseline method (Fig.\ref{abla:1dis}), the result without disturbance generates unnatural brightness. Specifically, in Fig.\ref{abla:nodis}, the building is too bright to match the brightness of the sky. Besides, the halo surrounding the building is more obvious. We also add more disturbed brightnesses for each scene. However, it fails to improve the performance further quantitatively or qualitatively but consumes more time in training.

\subsection{Adaptation to More Lighting Conditions}
\label{Adapt}

\begin{figure} [t]
    \centering
	  \subfloat[Darker Input]{ \label{Light:1_116_input}
       \includegraphics[width=0.3\linewidth]{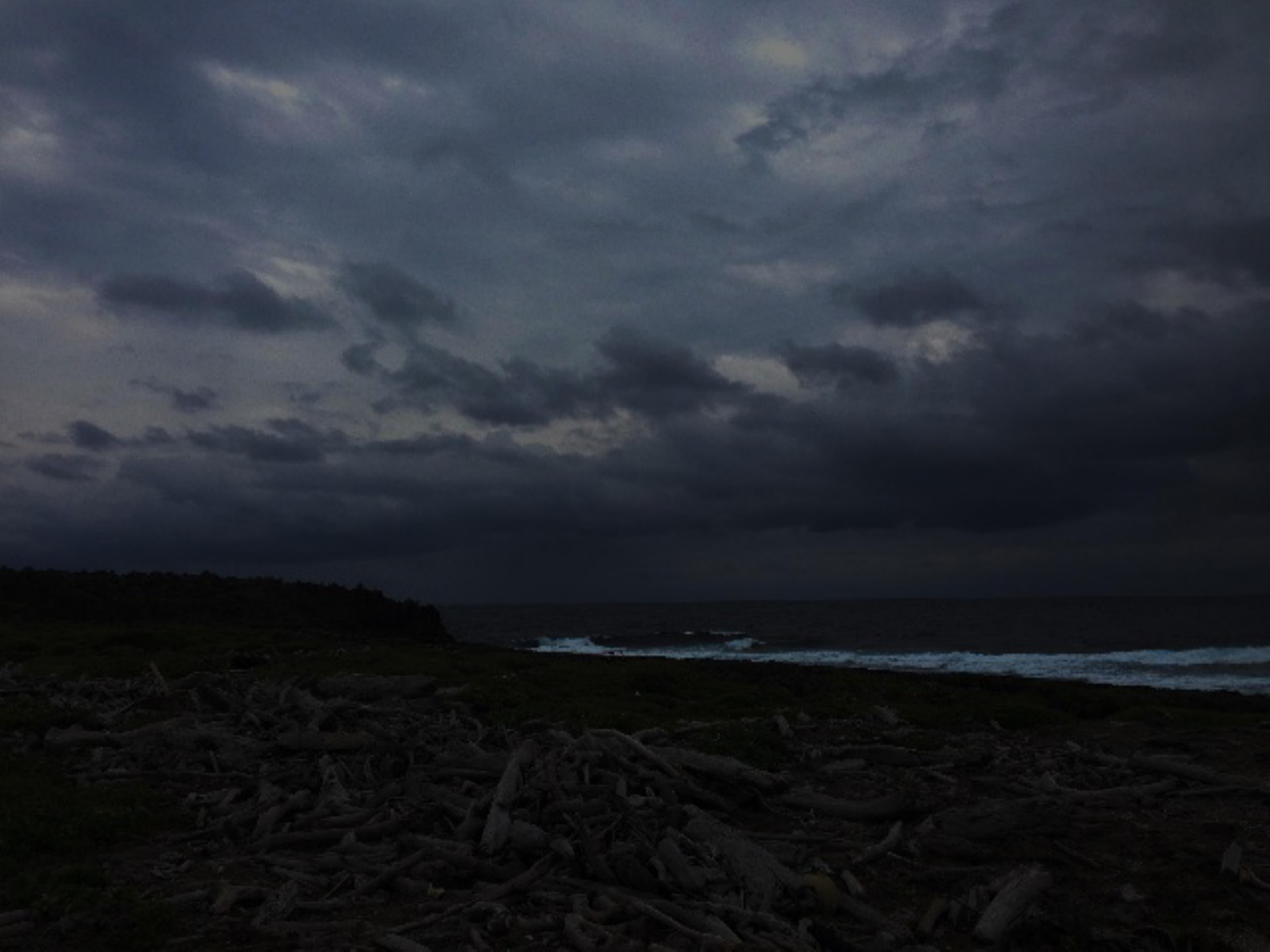}}
	  \quad
	  \subfloat[Zero-DCE \cite{guo2020zero}]{\label{Light:1_116_zerodce}
        \includegraphics[width=0.3\linewidth]{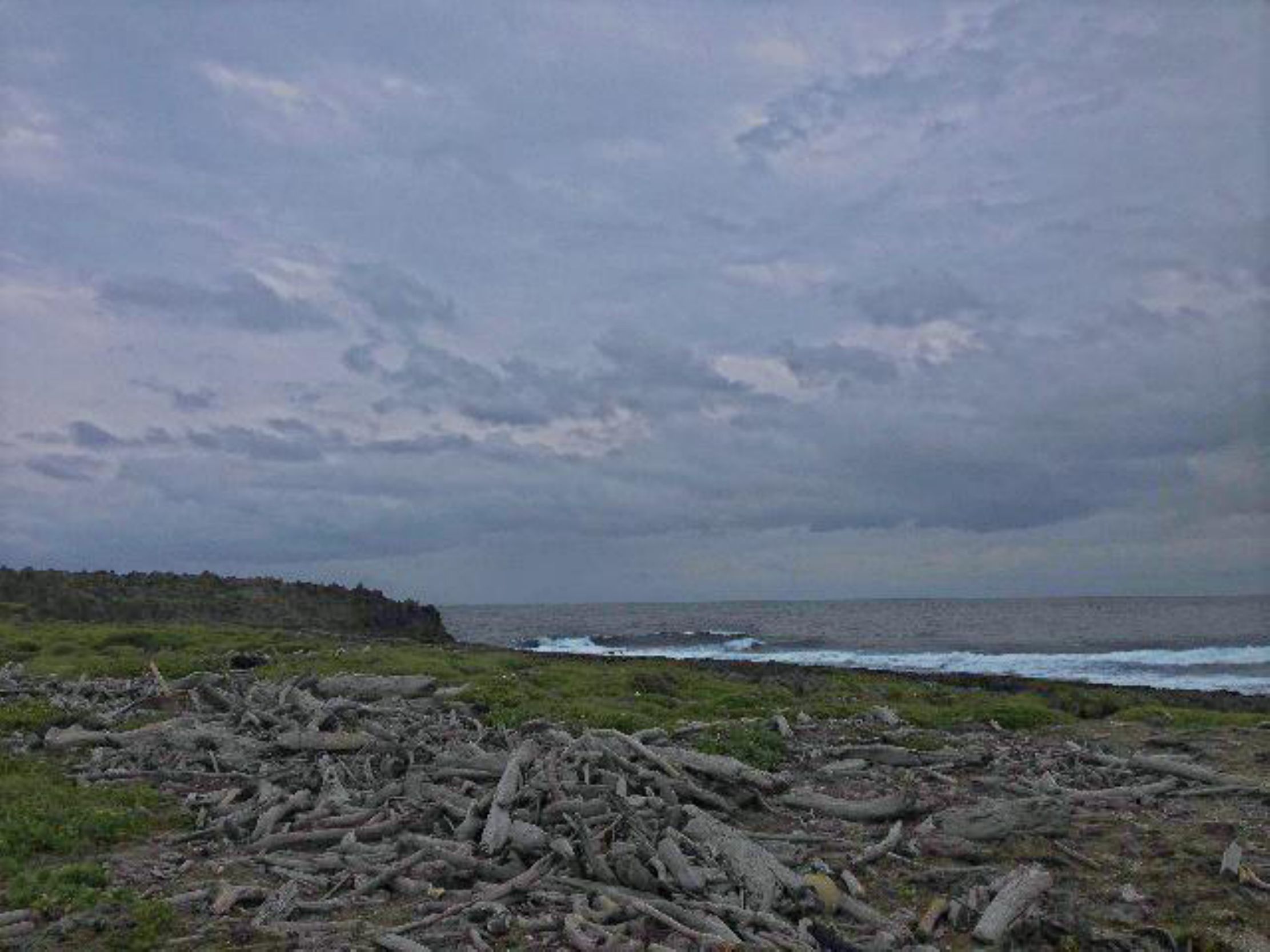}}
	  \quad
	  \subfloat[Ours]{\label{Light:1_116_ours}
        \includegraphics[width=0.3\linewidth]{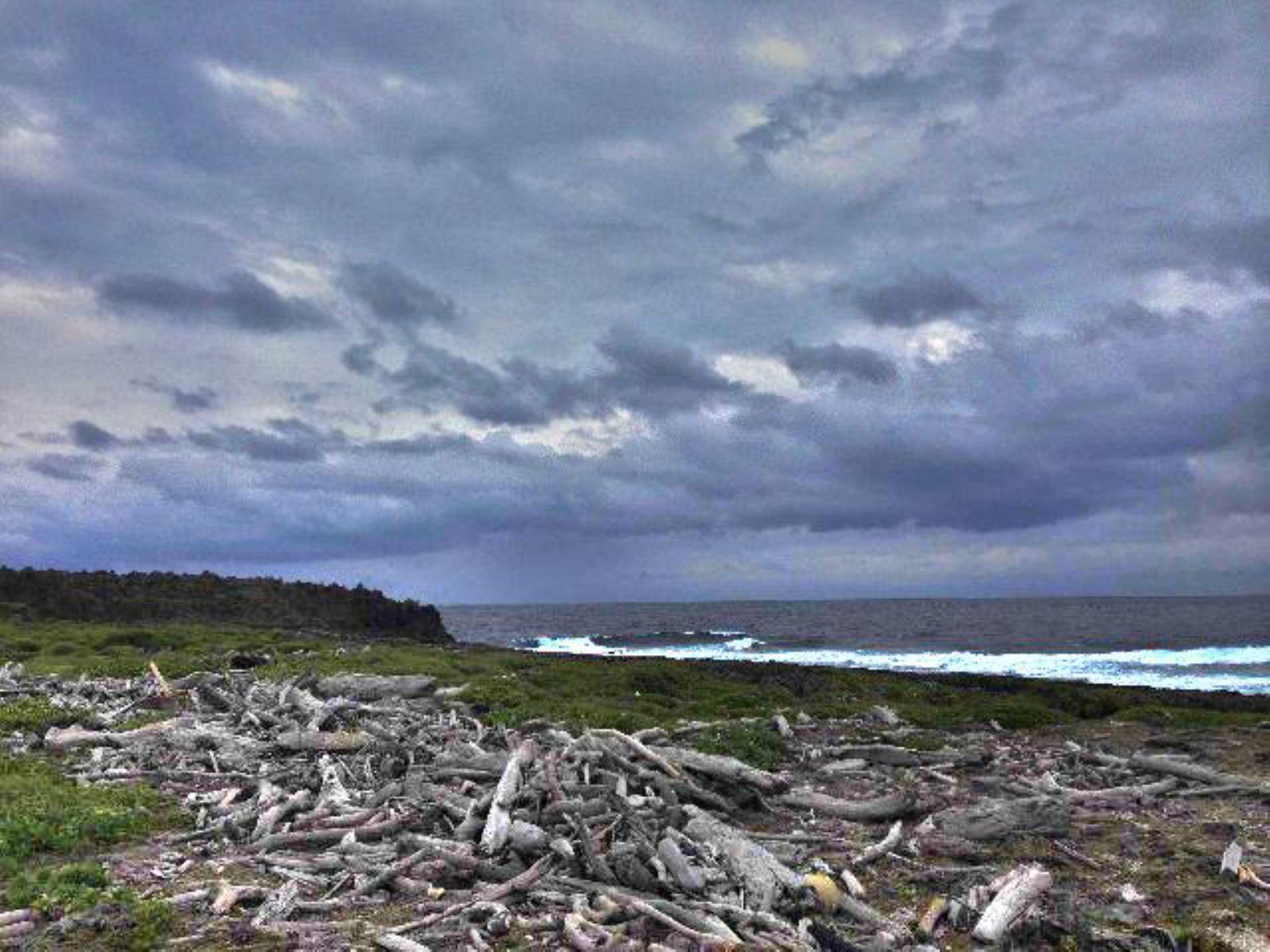}}\\
      \subfloat[Brighter Input]{ \label{Light:3_116_input}
       \includegraphics[width=0.3\linewidth]{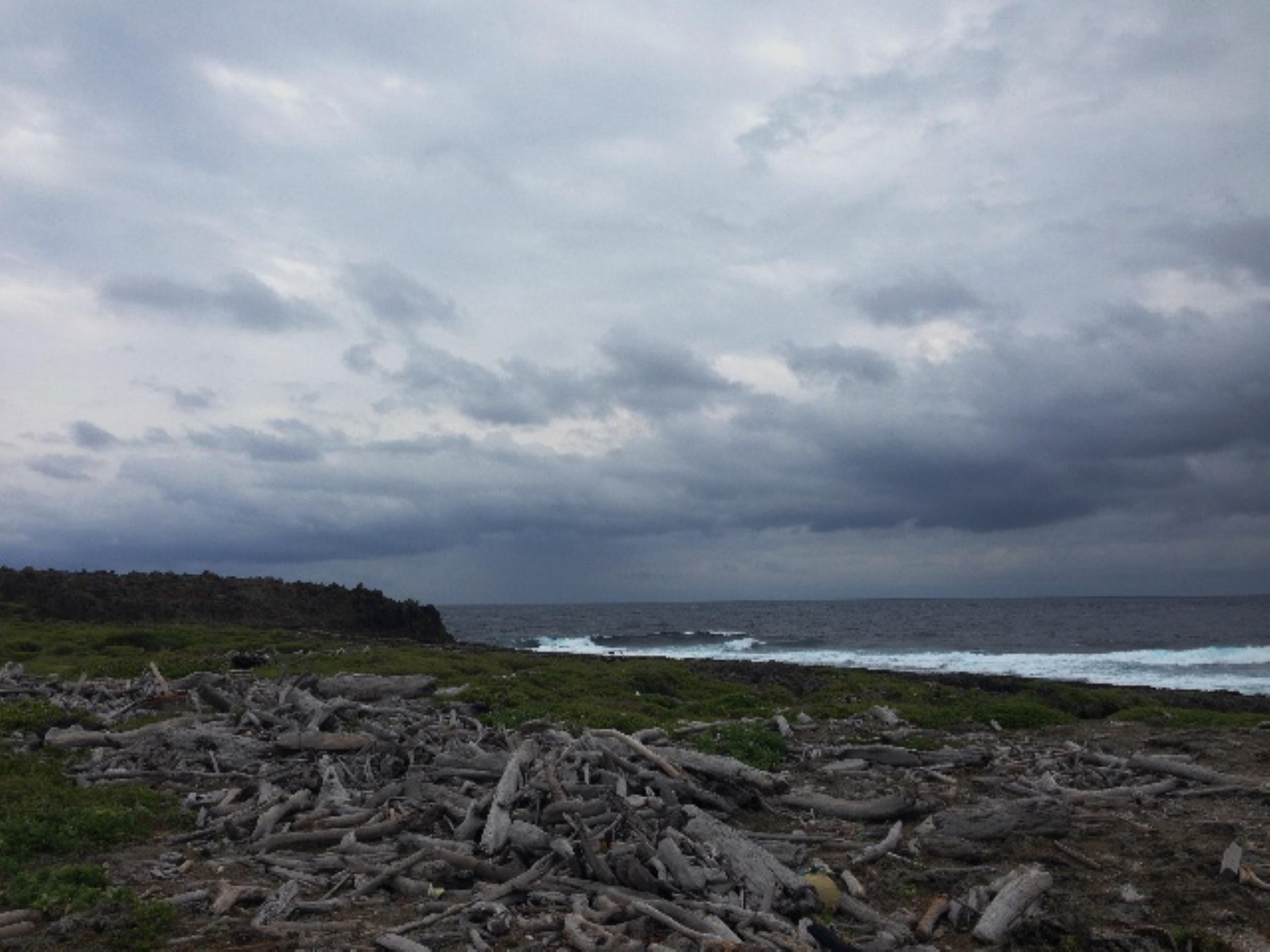}}
	  \quad
	  \subfloat[Zero-DCE \cite{guo2020zero}]{\label{Light:3_116_zerodce}
        \includegraphics[width=0.3\linewidth]{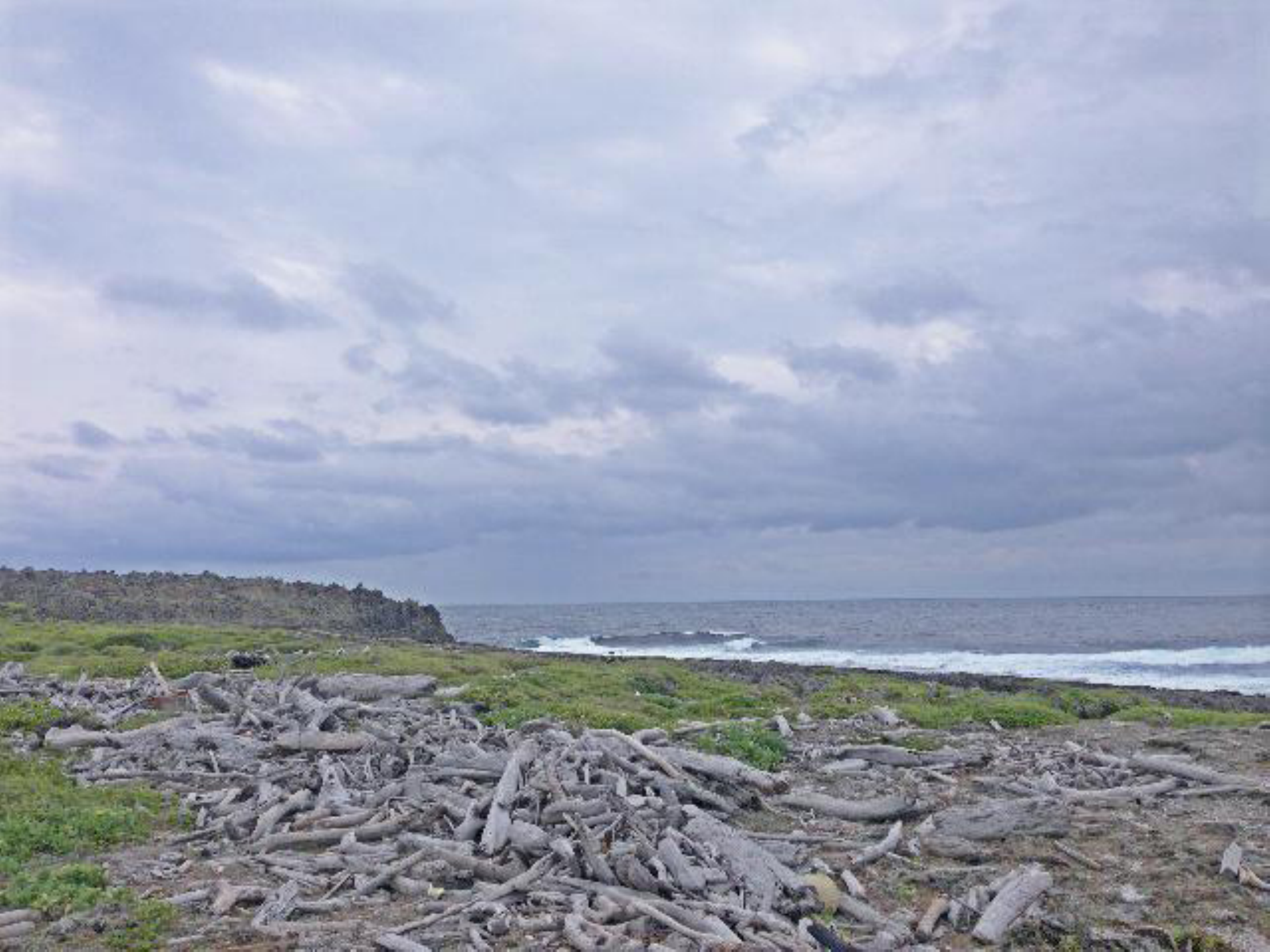}}
	  \quad
	  \subfloat[Ours]{\label{Light:3_116_ours}
        \includegraphics[width=0.3\linewidth]{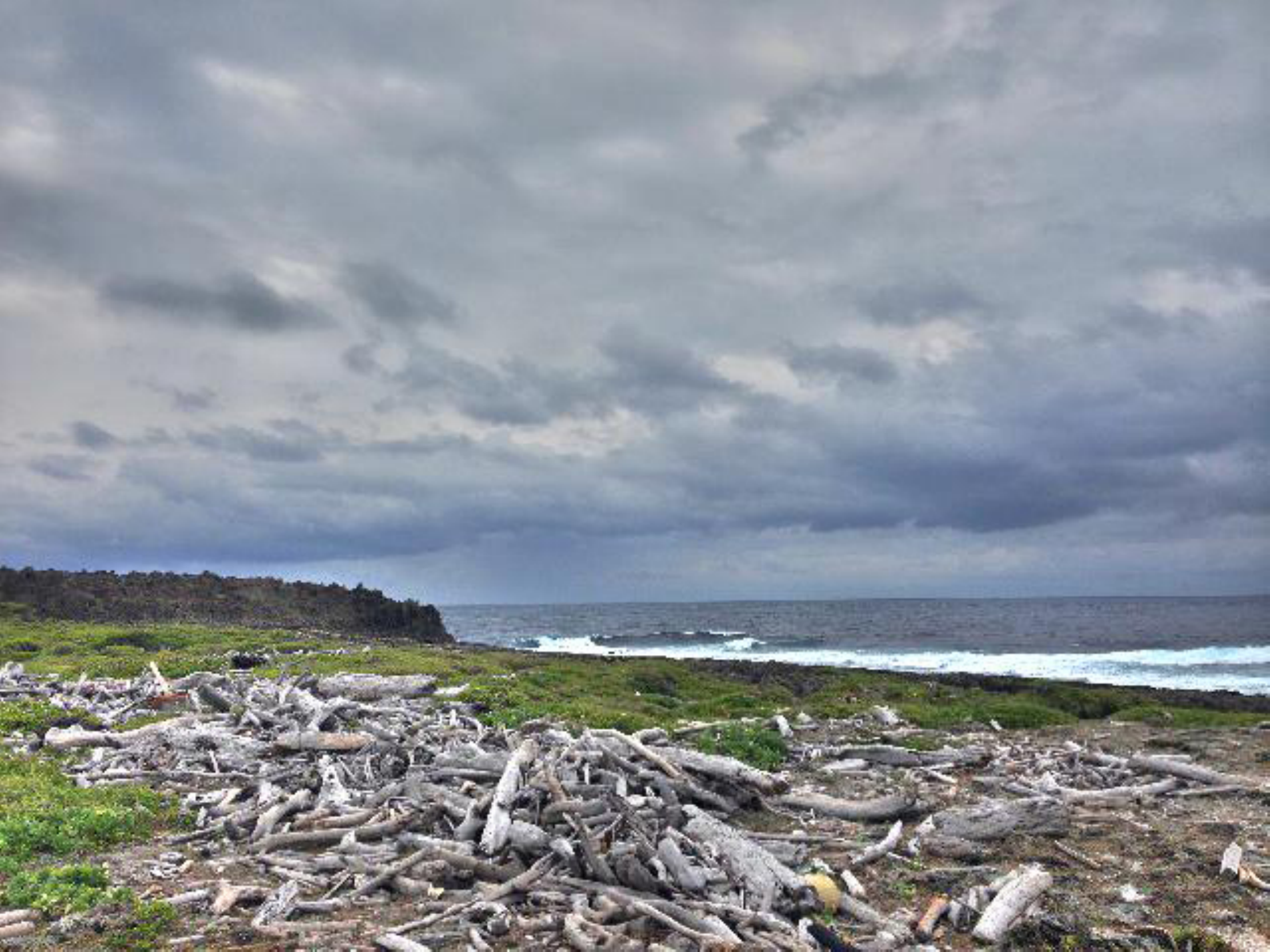}}\\
      \subfloat[Darker Input]{ \label{Light:1_187_input}
       \includegraphics[width=0.3\linewidth]{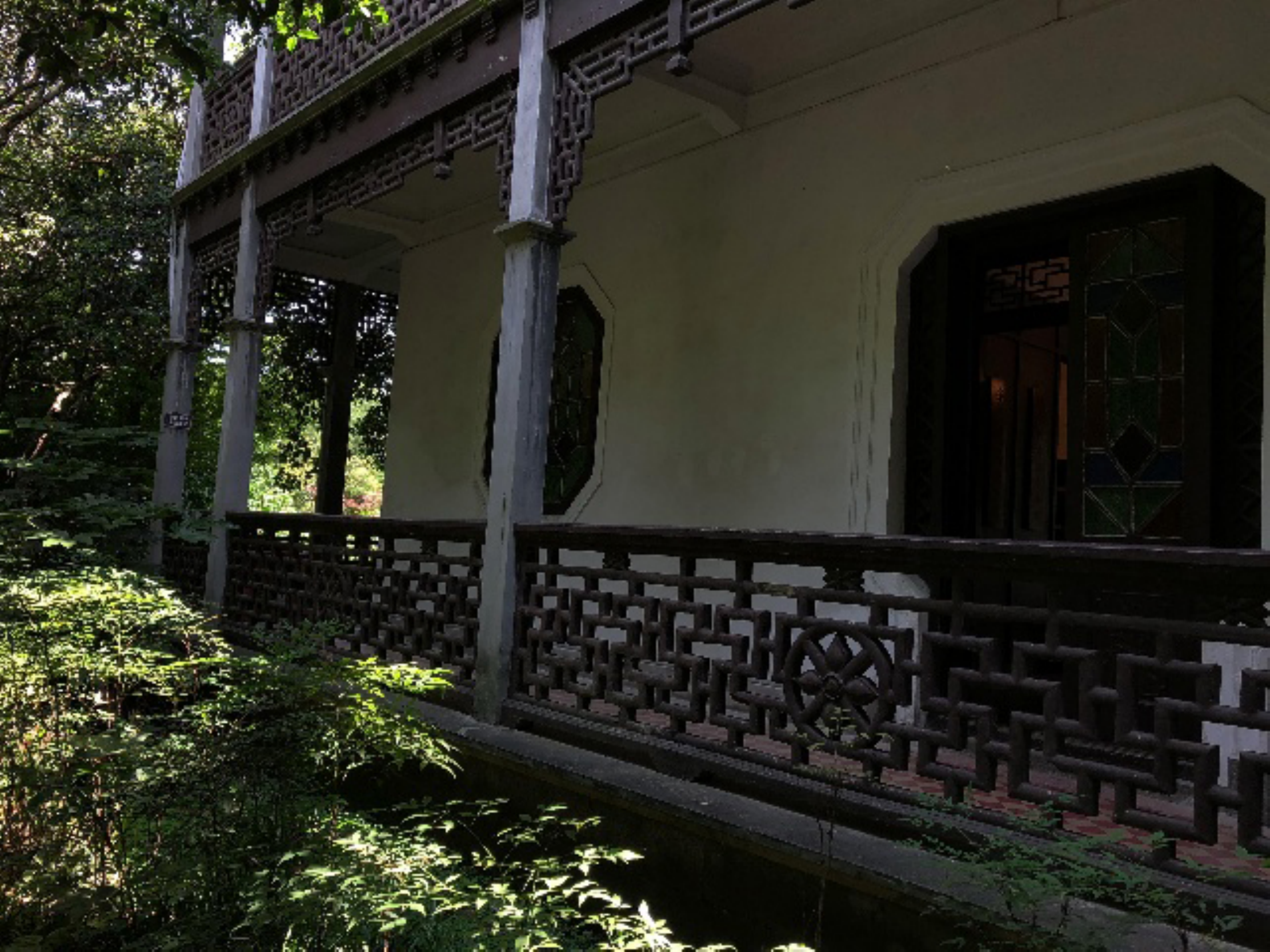}}
	  \quad
	  \subfloat[Zero-DCE \cite{guo2020zero}]{\label{Light:1_187_zerodce}
        \includegraphics[width=0.3\linewidth]{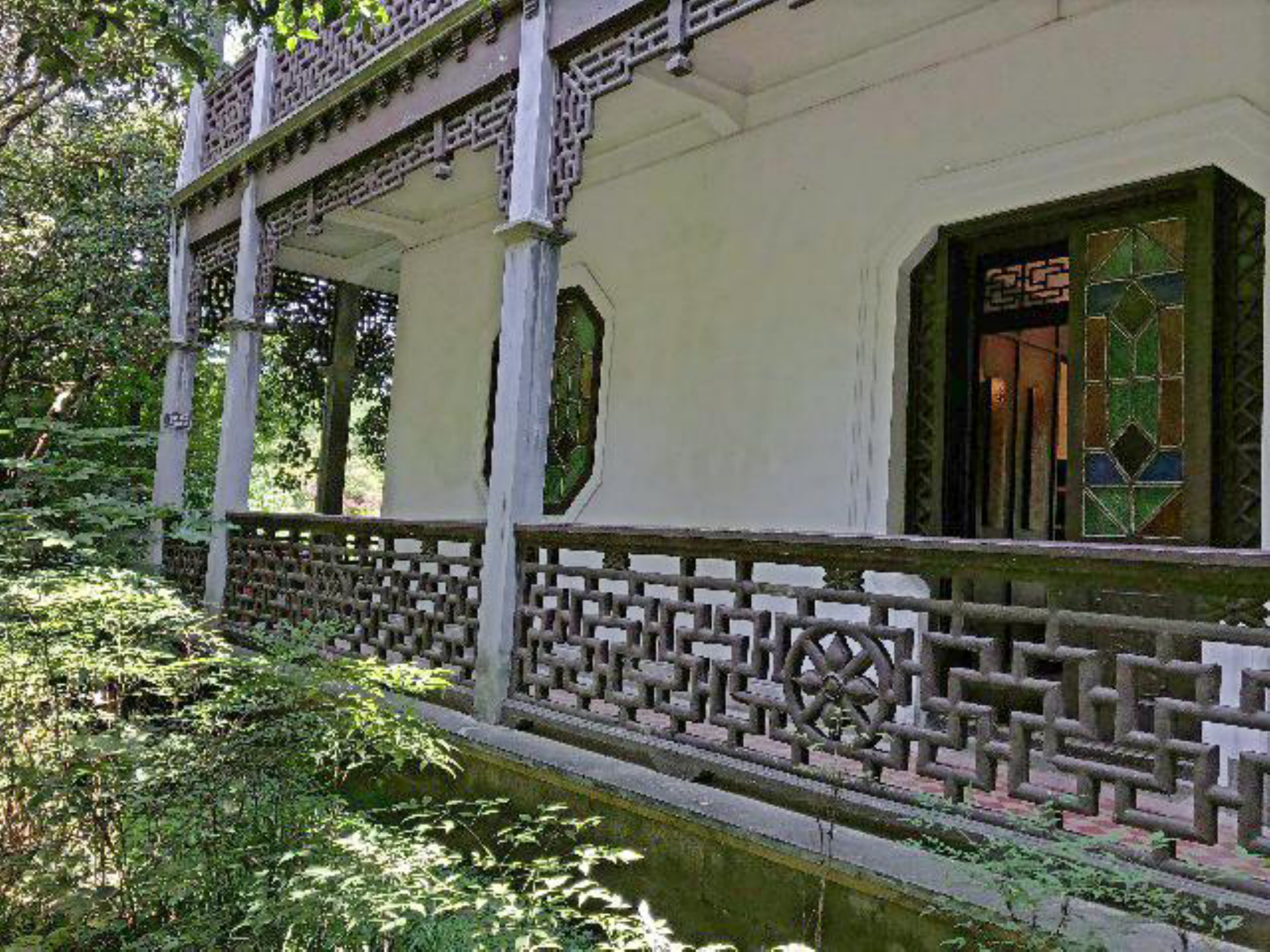}}
	  \quad
	  \subfloat[Ours]{\label{Light:1_187_ours}
        \includegraphics[width=0.3\linewidth]{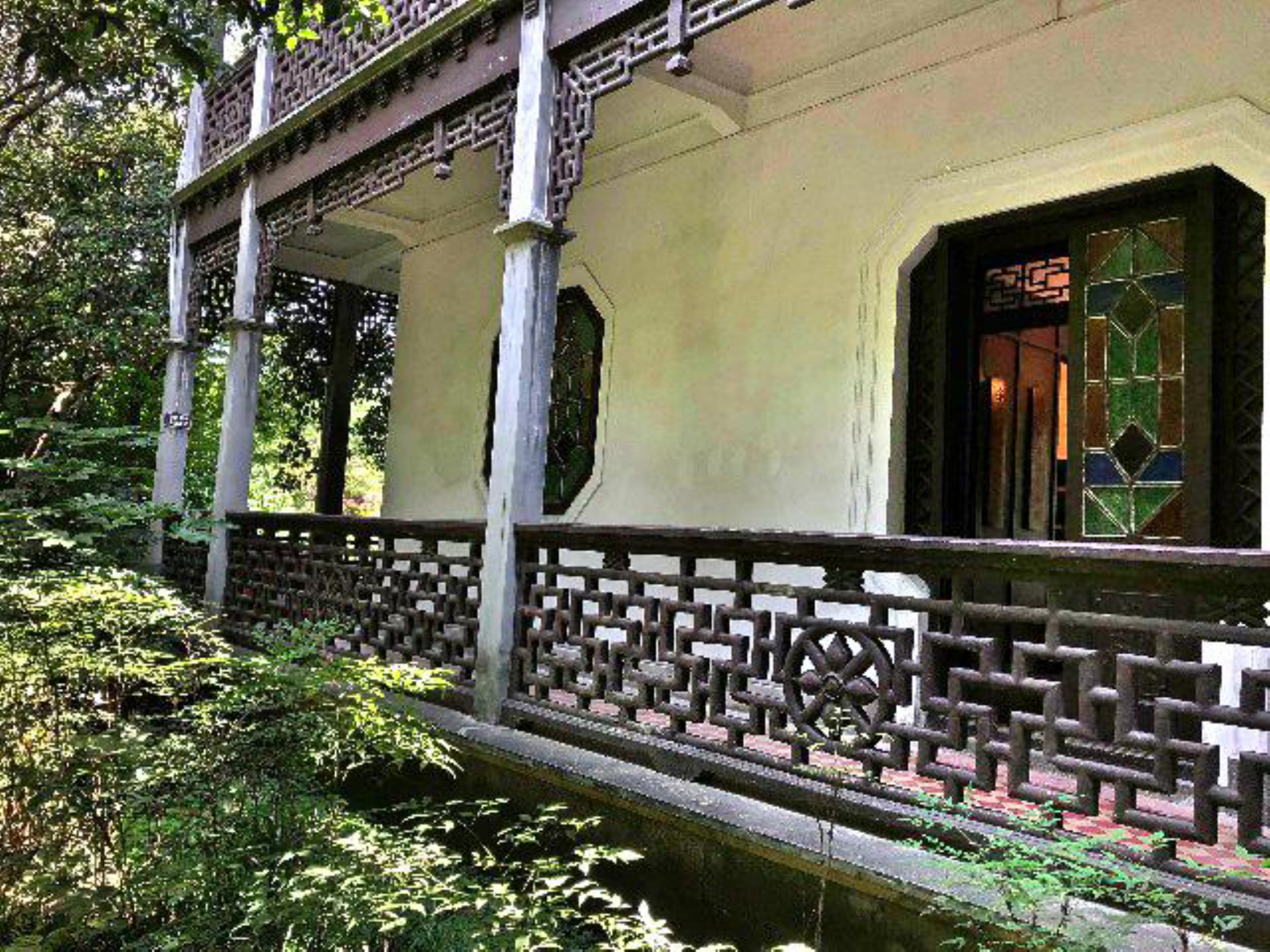}}\\
      \subfloat[brighter Input]{ \label{Light:3_187_input}
       \includegraphics[width=0.3\linewidth]{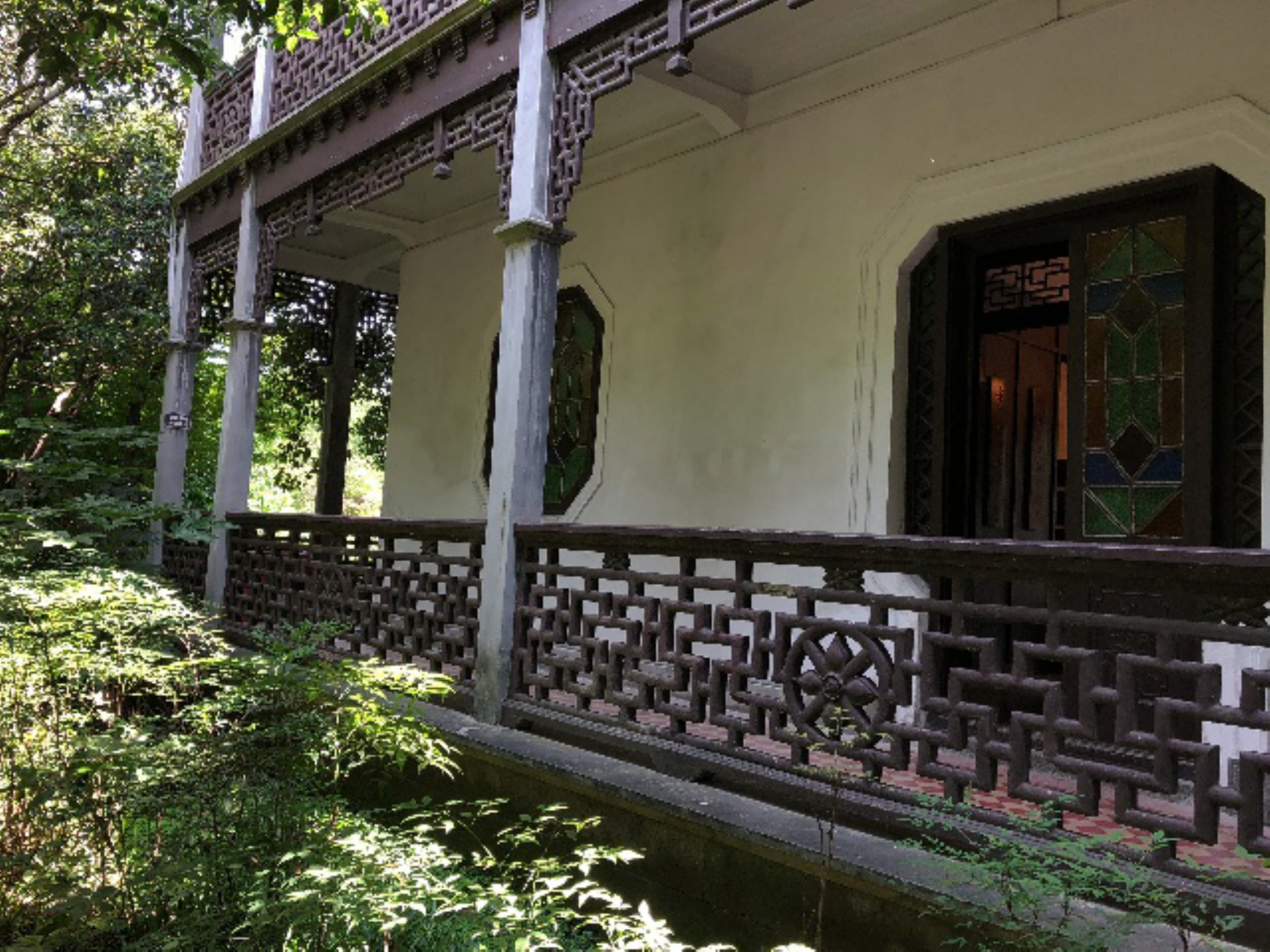}}
	  \quad
	  \subfloat[Zero-DCE \cite{guo2020zero}]{\label{Light:3_187_zerodce}
        \includegraphics[width=0.3\linewidth]{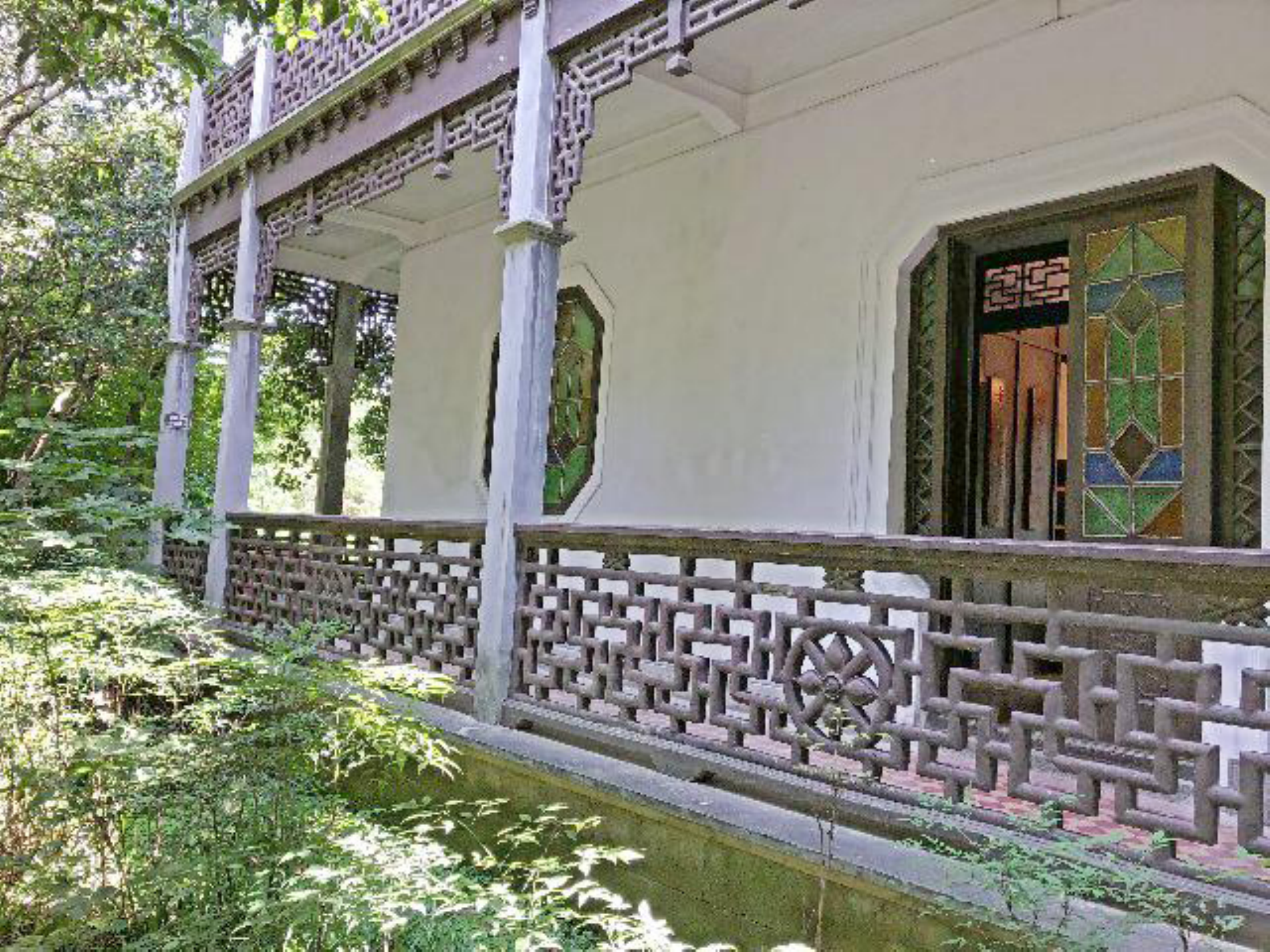}}
	  \quad
	  \subfloat[Ours]{\label{Light:3_187_ours}
        \includegraphics[width=0.3\linewidth]{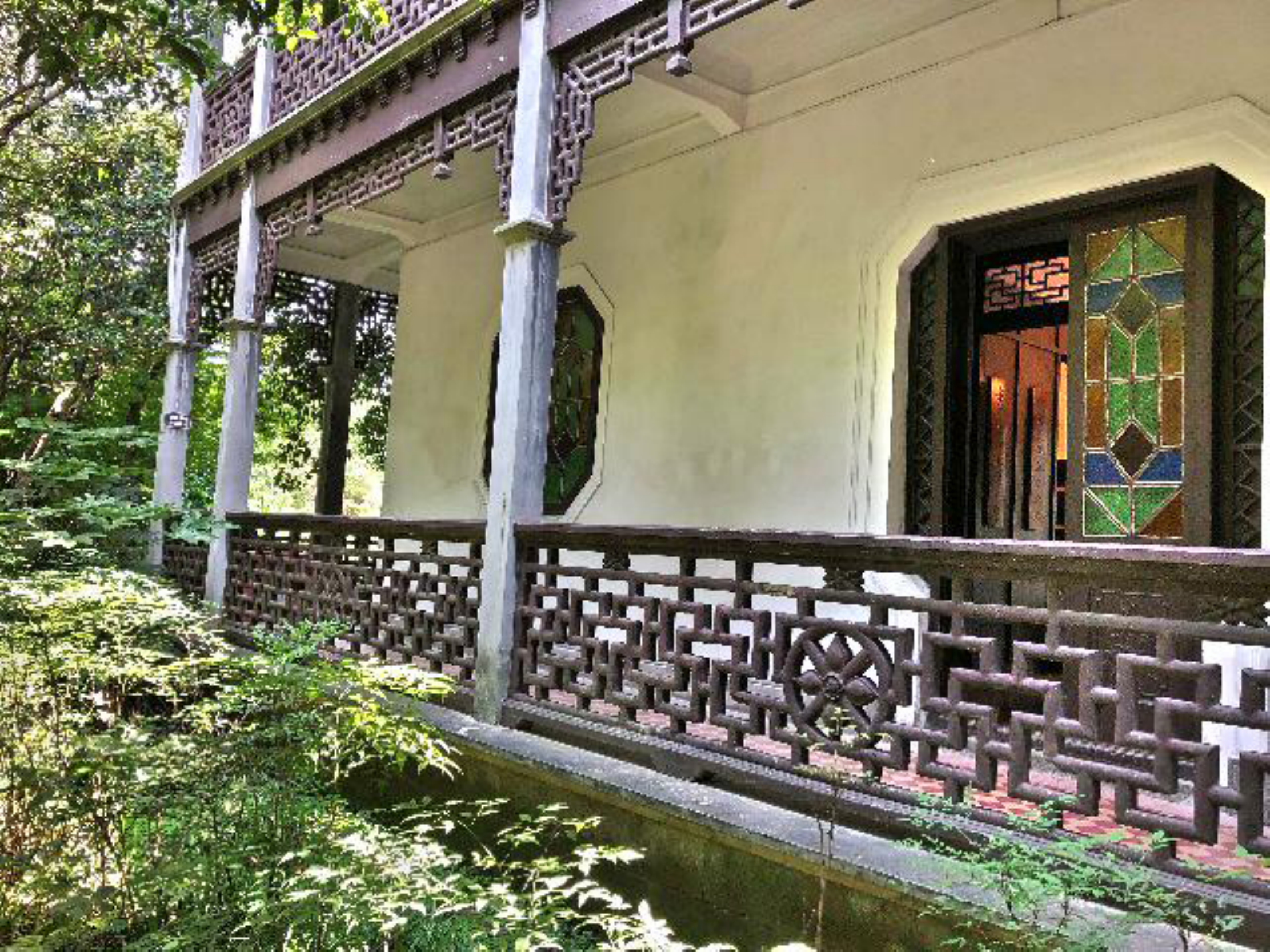}}
	  \caption{The results of input images with different lighting conditions enhanced by Zero-DCE \cite{guo2020zero} and our method.}
	  \label{fig:light} 
\end{figure}

Thanks to treating reflectance as enhanced brightness, our method can be adapted to more lighting conditions. Specifically, we randomly select two sequences from the Part2 of SICE dataset \cite{cai2018learning} as inputs. Each sequence contains a darker input and a brighter input of the same scene. We enhance them by Zero-DCE \cite{guo2020zero} and our method. The results are shown in Fig.\ref{fig:light}.

For the darker inputs, the results of Zero-DCE (Fig.\ref{Light:1_116_zerodce} and Fig.\ref{Light:1_187_zerodce}) are still under-exposed. Besides, for the brighter inputs, Zero-DCE produces more over-exposed regions (\eg, the bush in Fig.\ref{Light:3_187_zerodce}). However, our method can generate more natural brightness under various lighting conditions, reducing under-/over-exposed regions in enhanced images.

\section{Conclusion}
\label{Conclusion}
This paper proposes a novel self-regularized low-light image enhancement method based on Retinex theory, focusing on reducing color deviation and adapting to more lighting conditions. Inspired by HSV color space, this method preserves all colors (Hue, Saturation) and only integrates Retinex theory into brightness (Value). A deep reflectance estimation network is regularized by restricting the consistency of reflectances embedded in both the original and a novel random disturbed form of the brightness. The generated reflectance is treated as enhanced brightness. Extensive experiments demonstrate that our method surpasses multiple state-of-the-art algorithms, qualitatively and quantitatively.

{\small
\bibliographystyle{ieee_fullname}
\bibliography{ours.bib}
}

\end{document}